\documentclass[10pt,twocolumn,letterpaper]{article}

\usepackage{cvpr}
\usepackage{times}
\usepackage{epsfig}
\usepackage{graphicx}
\usepackage{amsmath}
\usepackage{amssymb}
\usepackage{caption}
\usepackage{bm}

\usepackage{booktabs}
\usepackage{algorithm}
\usepackage{algorithmic}
\usepackage{graphicx}
\usepackage{enumitem}
\usepackage{setspace}
\usepackage{multirow}
\usepackage{caption}
\usepackage[normalem]{ulem}
\usepackage{xcolor,colortbl, makecell}
\usepackage{txfonts}
\usepackage{blindtext}
\usepackage[printwatermark]{xwatermark}
\usepackage{mdframed}
\usepackage{tablefootnote} 
\usepackage{array, ragged2e}
\usepackage{adjustbox}

\makeatletter 
\AfterEndEnvironment{mdframed}{%
 \tfn@tablefootnoteprintout% 
 \gdef\tfn@fnt{0}% 
}
\makeatother

\newcommand\customfont[1]{{\usefont{T1}{monoton}{m}{n}#1}}
\newcommand\swag{{\tiny \usefont{T1}{permanentmarker}{m}{n}{SWAG}}}

\newcommand{\datasetname}{{\footnotesize\customfont{VCR}}}
\newcommand{\datasetnamelong}{\textbf{Visual Commonsense Reasoning}}

\newcommand{\evaluationname}{Adversarial Matching}
\newcommand{\modelname}{\textbf{R2C}}
\newcommand{\modelnamelong}{Recognition to Cognition Networks}

\newcommand*\inlineimage[1]{\raisebox{-0.14\baselineskip}{\includegraphics[height=0.95\baselineskip]{#1}}}

\newcommand{\websitelink}{{\tt \href{https://visualcommonsense.com}{visualcommonsense.com}}}
\newcommand{\leaderboardlink}{{\tt \href{https://visualcommonsense.com/leaderboard/}{visualcommonsense.com/leaderboard}}}

\newcommand{\persontag}[3]{\setlength{\fboxsep}{1pt}\colorbox{#1}{\small \usefont{T1}{robotomono}{m}{n}{[#2\hspace{1pt}\inlineimage{#3}\hspace{1pt}]}}}

\definecolor{figtwoperson1}{HTML}{FEFDAF}
\definecolor{figtwoperson2}{HTML}{CBFEFF}
\definecolor{figtwoperson3}{HTML}{B390FE}
\definecolor{figtwoperson4}{HTML}{FFBEFF}
\definecolor{figoneperson1}{HTML}{ECB8E8}
\definecolor{figoneperson2}{HTML}{B5D7B5}

\newcommand{\figtwopersonone}{\persontag{figtwoperson1}{person1}{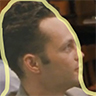}}
\newcommand{\figtwopersontwo}{\persontag{figtwoperson2}{person2}{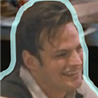}}
\newcommand{\figtwopersonthree}{\persontag{figtwoperson3}{person3}{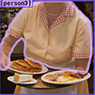}}
\newcommand{\figtwopersonfour}{\persontag{figtwoperson4}{person4}{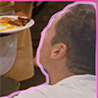}}

\newcommand{\figonepersonone}{\persontag{figoneperson1}{person1}{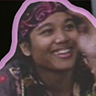}}
\newcommand{\figonepersontwo}{\persontag{figoneperson2}{person2}{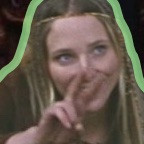}}

\newcommand{\com}[1]{}
\newcommand{\term}[1]{\emph{#1}}  % defining a new term
\usepackage{amsmath}  % required if you use \overset

\DeclareMathOperator*{\softmax}{softmax}

\makeatletter
\renewcommand\paragraph{\@startsection{paragraph}{4}{\z@}%
                                    {.1ex \@plus.2ex \@minus0ex}%
                                    {-1em}%
                                    {\hspace{\parindent}\normalfont\normalsize\bfseries}}

\makeatother

\usepackage[pagebackref=true,breaklinks=true,letterpaper=true,colorlinks,bookmarks=false]{hyperref}

\cvprfinalcopy % *** Uncomment this line for the final submission

\def\cvprPaperID{**} % *** Enter the CVPR Paper ID here

\ifcvprfinal\fi
\begin{document}

% \usetikzlibrary{calc}
\newwatermark*[page=1, angle=0,scale=1,xpos=2.9in,ypos=4.58in]{\includegraphics[height=54pt]{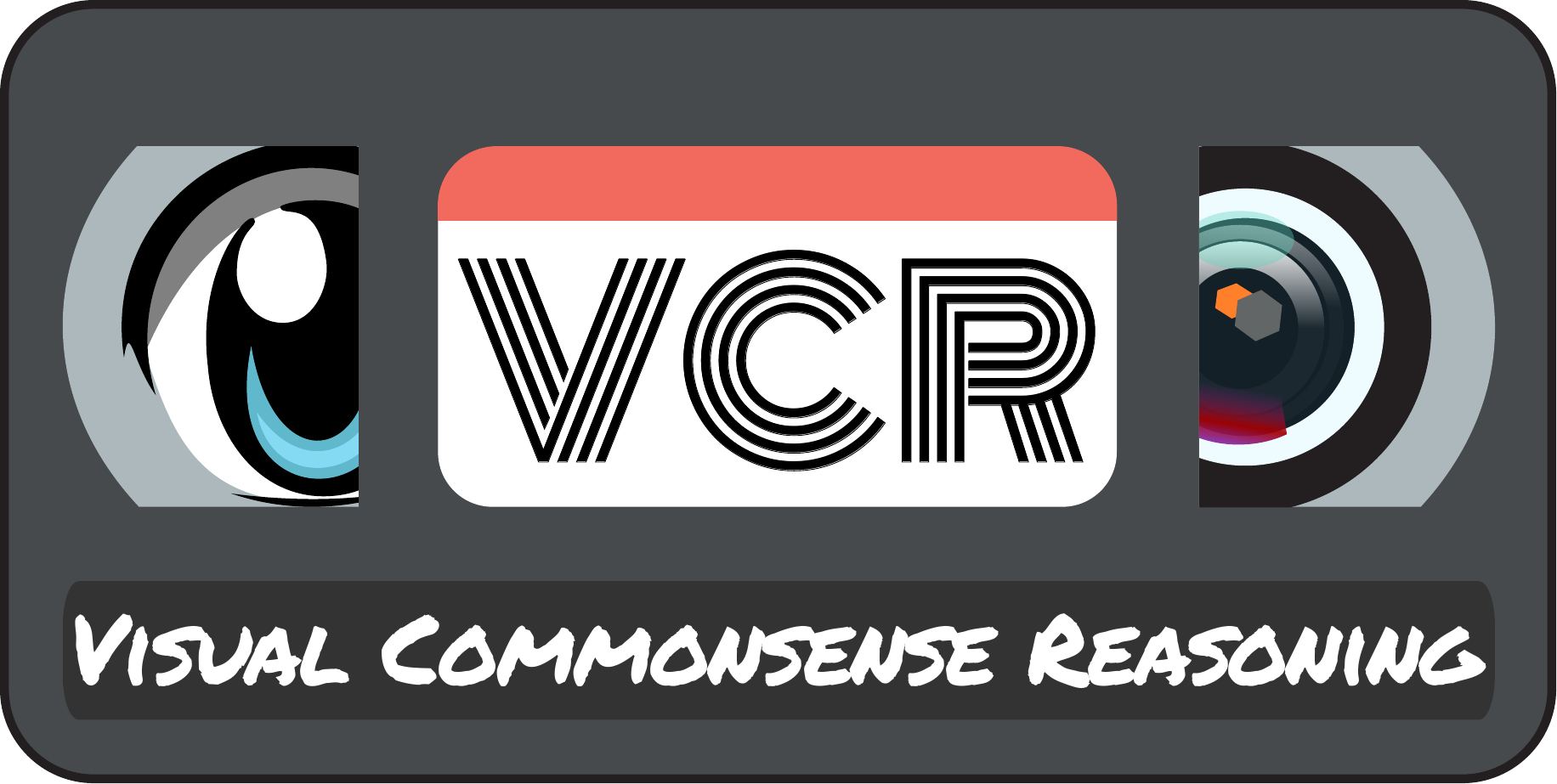}}

%%%%%%%%% TITLE
% \title{zLooking Beyond the Frame: Commonsense Inference about the Visual World}
% \title{Answering and Justifying Visual Commonsense Inferences}
% \title{Commonsense Visual Question Answering and Justification}
% \title{Answering and Justifying commonsense Commonsense Visual Question Answering and Justification}
%\title{Visually Grounded Commonsense Inference}
% \title{Visual Commonsense Reasoning}
\title{From Recognition to Cognition: Visual Commonsense Reasoning}

% Rowan, Yonatan, Ali, Yejin
\author{Rowan Zellers$^\spadesuit$ \: \: 
  Yonatan Bisk$^\spadesuit$ \: \:
  Ali Farhadi$^{\spadesuit\heartsuit}$ \: \:
  Yejin Choi$^{\spadesuit\heartsuit}$\\
  $^\spadesuit$Paul G. Allen School of Computer Science \& Engineering, University of Washington \\
  $^\heartsuit$Allen Institute for Artificial Intelligence \\
    \websitelink\vspace*{-2mm}
  }

% \author{First Author\\
% Institution1\\
% Institution1 address\\
% {\tt\small firstauthor@i1.org}
% % For a paper whose authors are all at the same institution,
% % omit the following lines up until the closing ``}''.
% % Additional authors and addresses can be added with ``\and'',
% % just like the second author.
% % To save space, use either the email address or home page, not both
% \and
% Second Author\\
% Institution2\\
% First line of institution2 address\\
% {\tt\small secondauthor@i2.org}
% }

%%% CALL MAKETITLE IN RENEWED TWOCOLUMN TO ADD FRONTPAGE FIGURE
\twocolumn[{
\renewcommand\twocolumn[1][]{#1}
\maketitle
\vspace*{-0.55cm}
\centering
% \begin{minipage}[t]{\columnwidth}
% 	\centering
% 	\includegraphics[width=\columnwidth]{figures/teaser.pdf}
% \end{minipage}
% \begin{minipage}[t]{\columnwidth}
% 	\centering
% 	\includegraphics[width=\columnwidth]{figures/teaser.pdf}
% \end{minipage}
\includegraphics[width=\linewidth]{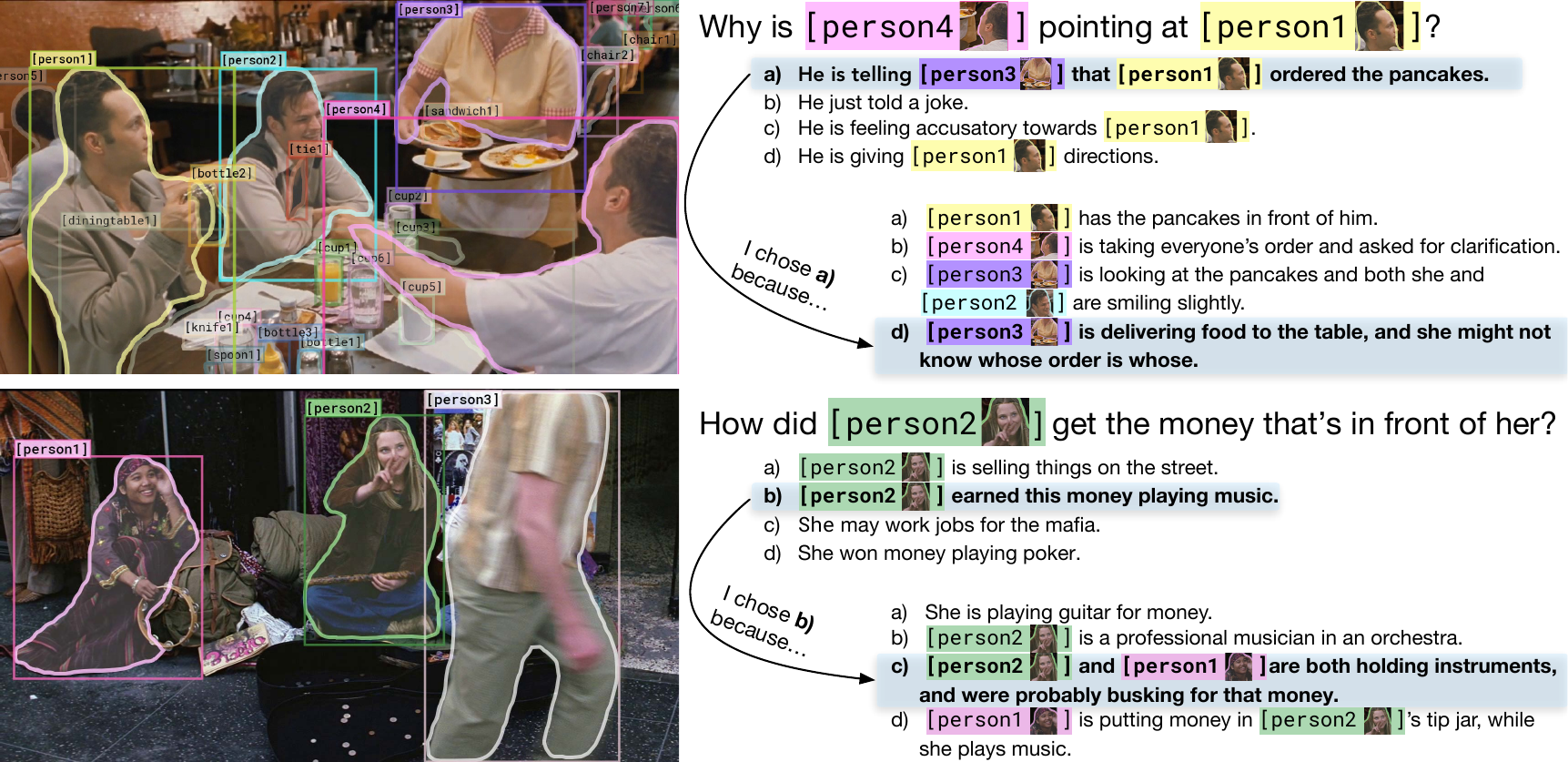}
\vspace{-0.55cm}
\captionof{figure}{\datasetname: Given an image, a list of regions, and a question, a model must answer the question and provide a \emph{rationale} explaining why its answer is right. Our questions challenge computer vision systems to go beyond recognition-level understanding, towards a higher-order cognitive and commonsense understanding of the world depicted by the image.}
%To answer a complex question, like `How did \protect\figonepersontwo get the money that's in front of her,' a model must perform higher-order commonsense reasoning over the objects and people detected, linking their actions to a mental model of the world.
%Our dataset contains 290,000 questions, answers, and rationales like these, annotated over 110,000 images.
\label{fig:teaser}
\vspace*{0.5cm}
}]

\maketitle
\thispagestyle{plain}
\pagestyle{plain}

%%%%%%%%% ABSTRACT
\begin{abstract}
\vspace{-10pt}
Visual understanding goes well beyond object recognition. With one glance at an image, we can effortlessly imagine the world beyond the pixels: for instance, we can infer people's actions, goals, and mental states. While this task is easy for humans, it is tremendously difficult for today's vision systems, requiring higher-order cognition and commonsense reasoning about the world. 
We formalize this task as \datasetnamelong. Given a challenging question about an image, a machine must answer correctly and then provide a rationale justifying its answer.

Next, we introduce a new dataset, \datasetname, consisting of 290k multiple choice QA problems derived from 110k movie scenes. The key recipe for generating non-trivial and high-quality problems at scale is \textbf{\evaluationname}, a new approach to transform rich annotations into multiple choice questions with minimal bias. Experimental results show that while humans find \datasetname~easy (over 90\% accuracy), state-of-the-art vision models struggle ($\sim$45\%).

To move towards cognition-level understanding, we present a new reasoning engine, \modelnamelong~(\modelname), that models the necessary layered inferences for grounding, contextualization, and reasoning. \modelname~helps narrow the gap between humans and machines ($\sim$65\%); still, the challenge is far from solved, and we provide analysis that suggests avenues for future work. 
\end{abstract}

%%%%%%%%% BODY TEXT
\vspace{-16pt}
\section{Introduction}
\vspace{-2pt}
With one glance at an image, we can immediately infer what is happening in the scene beyond what is visually obvious. For example, in the top image of Figure~\ref{fig:teaser}, not only do we see several objects (people, plates, and cups), we can also reason about the entire situation: three people are dining together, they have already ordered their food before the photo has been taken,~\figtwopersonthree~is serving and not eating with them, and what~\figtwopersonone~ordered are the pancakes and bacon (as opposed to the cheesecake), because \figtwopersonfour~is pointing to~\figtwopersonone~while looking at the server,~\figtwopersonthree. 

Visual understanding requires seamless integration between \emph{recognition} and \emph{cognition}: beyond recognition-level perception (e.g., detecting objects and their attributes), one must perform cognition-level reasoning (e.g., inferring the likely intents, goals, and social dynamics of people) \cite{Davis2015CommonsenseRA}.
State-of-the-art vision systems can reliably perform \emph{recognition-level} image understanding, but struggle with complex inferences, like those in Figure~\ref{fig:teaser}.
We argue that as the field has made significant progress on recognition-level building blocks, such as object detection, pose estimation, and segmentation, now is the right time to tackle cognition-level reasoning at scale.

As a critical step toward complete visual understanding, we present the task of \datasetnamelong. 
Given an image, a machine must answer a question that requires a thorough understanding of the visual world evoked by the image.
Moreover, the machine must provide a rationale justifying why that answer is true, referring to the details of the scene, as well as background knowledge about how the world works. These questions, answers, and rationales are expressed using a mixture of rich natural language as well as explicit references to image regions\com{ detected by state-of-the-art object detection systems \cite{Detectron2018, He2017MaskR}}. 
To support clean-cut evaluation, all our tasks are framed as multiple choice QA.

Our new dataset for this task, \datasetname, is the first of its kind and is large-scale --- 290k pairs of questions, answers, and rationales, over 110k unique movie scenes. A crucial challenge in constructing a dataset of this complexity at this scale is how to avoid annotation artifacts. A recurring challenge in most recent QA datasets has been that human-written answers contain unexpected but distinct biases that models can easily exploit. Often these biases are so prominent so that models can select the right answers without even looking at the questions \cite{gururangan2018annotation, poliak_hypothesis_2018, Schwartz:2017}. 

Thus, we present \textbf{\evaluationname}, a novel QA assignment algorithm that allows for robust multiple-choice dataset creation at scale.
The key idea is to recycle each correct answer for a question exactly three times --- as a negative answer for three other questions. Each answer thus has the same probability (25\%) of being correct: this resolves the issue of answer-only biases, and disincentivizes machines from always selecting the most generic answer. We formulate the answer recycling problem as a constrained optimization based on the relevance and entailment scores between each candidate negative answer and the gold answer, as measured by state-of-the-art natural language inference models \cite{chen2017enhanced,peters2018deep,devlin2018bert}. A neat feature of our recycling algorithm is a knob that can control the tradeoff between human and machine difficulty: we want the problems to be hard for machines while easy for humans. 

Narrowing the gap between recognition- and cognition-level image understanding requires grounding the meaning of the natural language passage in the visual data, understanding the answer in the context of the question, and reasoning over the shared and grounded understanding of the question, the answer, the rationale and the image. In this paper we introduce a new model, \textbf{\modelnamelong}~(\modelname). 
Our model performs three inference steps. 
First, it \emph{grounds} the meaning of a natural language passage  with respect to the image regions (objects) that are directly referred to. 
It then \emph{contextualizes} the meaning of an answer with respect to the question that was asked, as well as the global objects not mentioned. Finally, it \emph{reasons} over this shared representation to arrive at an answer. 

Experiments on \datasetname~show that \modelname~greatly outperforms state-of-the-art visual question-answering systems: obtaining 65\% accuracy at question answering, 67\% at answer justification, and 44\% at staged answering and justification. Still, the task and dataset is far from solved: humans score roughly 90\% on each. We provide detailed insights and an ablation study to point to avenues for future research. 

In sum, our major contributions are fourfold: (1) we formalize a new task, Visual Commonsense Reasoning, and (2) present a large-scale multiple-choice QA dataset, \datasetname, (3) that is automatically assigned using \evaluationname, a new algorithm for robust multiple-choice dataset creation. (4) We also propose a new model, \modelname, that aims to mimic the layered inferences from recognition to cognition; this also establishes baseline performance on our new challenge. The dataset is available to download, along with code for our model, at \websitelink.

\section{Task Overview}
\begin{figure}[t!]
    \vspace{-3mm}
    \centering
    \includegraphics[width=\columnwidth]{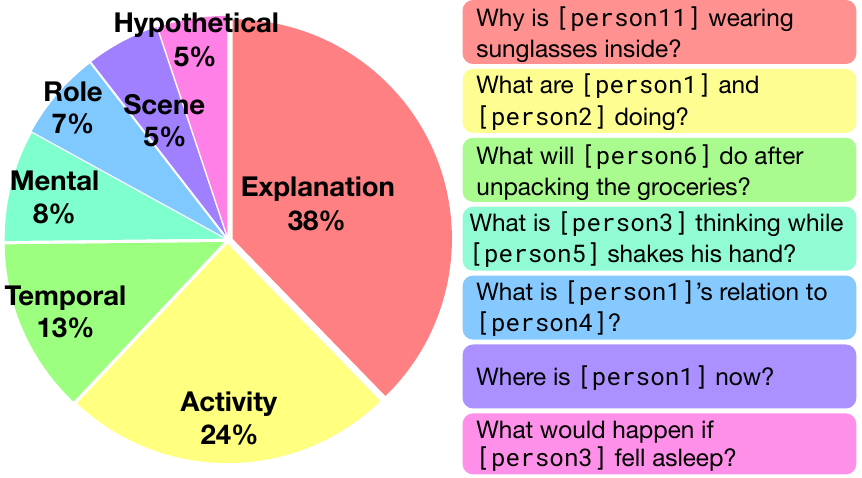}
    \vspace*{-4mm}
    \caption{Overview of the types of inference required by questions in \datasetname. Of note, 38\% of the questions are explanatory `why' or `how' questions, 24\% involve cognition-level activities, and 13\% require temporal reasoning (i.e., what might come next). These categories are not mutually exclusive; an answer might require several hops of different types of inferences (see appendix Sec~\ref{sec:datasetanalysis}).}
    \label{fig:questiondiversity}
\end{figure}

We present \datasetname, a new task that challenges vision systems to holistically and cognitively understand the content of an image. For instance, in Figure~\ref{fig:teaser}, we need to understand the activities (\figtwopersonthree~is delivering food), the {roles} of people (\figtwopersonone~is a customer who previously ordered food), the mental states of people (\figtwopersonone~wants to eat), and the likely events before and after the scene (\figtwopersonthree~will serve the pancakes next). Our task covers these categories and more: a distribution of the inferences required is in Figure~\ref{fig:questiondiversity}.

Visual understanding requires not only answering questions correctly, but doing so \emph{for the right reasons}. We thus require a model to give a \emph{rationale} that explains why its answer is true. Our questions, answers, and rationales are written in a mixture of rich natural language as well as detection tags, like `\figtwopersontwo': this helps to provide an unambiguous link between the textual description of an object (`the man on the left in the white shirt') and the corresponding image region. 

To make evaluation straightforward, we frame our ultimate task -- of staged answering and justification -- in a multiple-choice setting. Given a question along with four answer choices, a model must first select the right answer. If its answer was correct, then it is provided four rationale choices (that could purportedly justify its correct answer), and it must select the correct rationale. We call this $Q{\rightarrow}AR$ as for the model prediction to be correct requires \emph{both the chosen answer and then the chosen rationale} to be correct.

Our task can be decomposed into two multiple-choice sub-tasks, that correspond to answering ($Q{\rightarrow}A$) and justification ($QA{\rightarrow}R$) respectively:

% Evaluation is straightforward, as our dataset is entirely multiple choice.

% % We decompose the task into the following three modes:
% For our task, we consider the following three modes:
% \begin{itemize}[labelwidth=!,itemsep=0pt,topsep=1pt,parsep=1pt]
%     \item $Q{\rightarrow}A$: given a question, select the correct answer.
%     \item $QA{\rightarrow}R$: given a question and the correct answer, select the correct rationale.
%     \item $Q{\rightarrow}AR$: given a question, select the correct answer, then the correct rationale. The model prediction is correct only if both the answer \emph{and the rationale} are correct, helping alleviate concerns that a model answers correctly, but for questionable reasons. %\cite{agrawal2016analyzing,  jabri2016revisiting, balanced_vqa_v2, agrawal2018don, zellers2018scenegraphs}.
% \end{itemize}
% The joint answering+justification setting ($Q{\rightarrow}AR$) decomposes naturally into answering ($Q{\rightarrow}A$) and justification ($QA{\rightarrow}R$). In this paper, we make the observation that both of these subtasks can be approached under a unified framework, which we describe below:

\mdfdefinestyle{MyFrame}{%
    linecolor=gray,
    linewidth=0.25pt,
%   topline=false,
%   rightline=false,
%   bottomline=false,
%   leftline=false,
    innertopmargin=3pt,
    innerbottommargin=3pt,
    innerrightmargin=3pt,
    innerleftmargin=3pt,
    leftmargin = -2pt,
    rightmargin = -2pt,
    backgroundcolor= black!03}
\vspace{-4pt}
\begin{mdframed}[style=MyFrame]
\paragraph{Definition} \emph{\datasetname~subtask.}
A single example of a \datasetname~subtask consists of an image $\boldsymbol{I}$, and: 

\begin{itemize}[labelwidth=!,itemsep=0pt,topsep=1pt,parsep=1pt]
\item A sequence $\boldsymbol{o}$ of object detections. Each object detection $o_{i}$ consists of a \term{bounding box} $\mathbf{b}$, a segmentation mask $\mathbf{m}$\tablefootnote{The task is agnostic to the representation of the mask, but it could be thought of as a list of polygons $\boldsymbol{p}$, with each polygon consisting of a sequence of 2d vertices inside the box $\boldsymbol{p}_j = \{x_t, y_t\}_t$.}, and a class label $\ell_i \in \mathcal{L}$.
\item A \term{query} $\boldsymbol{q}$, posed using a mix of natural language and pointing. Each word $\boldsymbol{q}_i$ in the query is either a word in a vocabulary $\mathcal{V}$, or is a tag referring to an object in $\boldsymbol{o}$. 
\item A set of $N$ \term{responses}, where each response $\boldsymbol{r}^{(i)}$ is written in the same manner as the query: with natural language and pointing. Exactly one response is correct.
\end{itemize}
The model chooses a single (best) response.
% The model predicts the most likely response.
% performs $N$-way classification among responses.
% The model must choose the correct answer; baseline accuracy is thus $1{/}N$.
% The model's goal is to predict which of the $N$ responses is correct., and we thus evaluate in terms of accuracy, where baseline performance is $1{/}N$.
\end{mdframed}
\vspace{-1mm}
In question-answering ($Q{\rightarrow}A$), the query is the question and the responses are answer choices. In answer justification ($QA{\rightarrow}R$), the query is the concatenated question and correct answer, while the responses are rationale choices.%We can then compute performance on $Q{\rightarrow}AR$ by combining the performance of the subtasks.\footnote{Specifically using a bitwise {\tt AND}.}

In this paper, we evaluate models in terms of accuracy and use $N{=}4$ responses. Baseline accuracy on each subtask is then 25\% ($1{/}N$). In the holistic setting ($Q{\rightarrow}AR$), baseline accuracy is 6.25\% ($1{/}N^2$) as there are two subtasks.
% In our paper, we will use $N{=}4$ responses.

\section{Data Collection}
\begin{figure*}[t!]
    \vspace{-3mm}
    \centering
    \includegraphics[width=\linewidth]{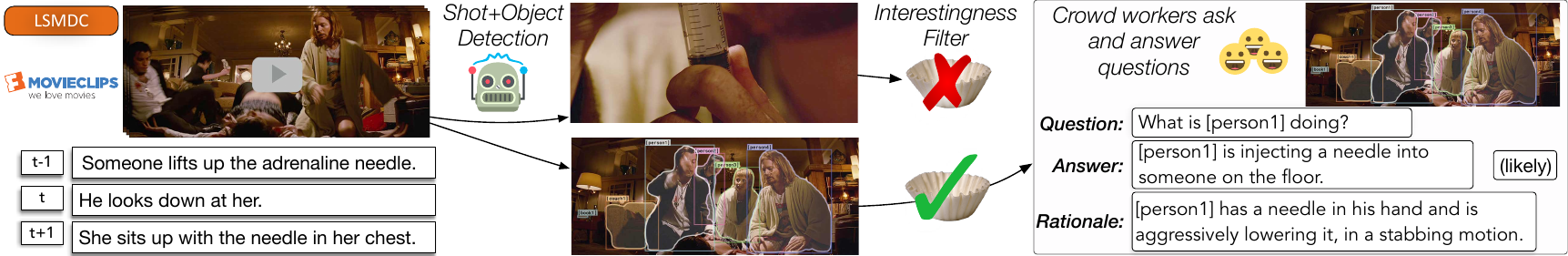}
    \caption{An overview of the construction of \datasetname. Using a state-of-the-art object detector \cite{He2017MaskR,Detectron2018}, we identify the objects in each image. The most interesting images are passed to crowd workers, along with scene-level context in the form of scene descriptions (MovieClips) and video captions (LSMDC, \cite{rohrbach_movie_2017}). The crowd workers use a combination of natural language and detection tags to ask and answer challenging visual questions, also providing a rationale justifying their answer.} 
    \label{fig:ourdataapproach}
 %   \vspace{-0.3cm}
\end{figure*}

In this section, we describe how we collect the questions, \emph{correct answers} and \emph{correct rationales} for \datasetname. Our key insight -- towards collecting commonsense visual reasoning problems at scale -- is to carefully select interesting situations. We thus extract still images from movie clips. The images from these clips describe complex situations that humans can decipher without additional context: for instance, in Figure~\ref{fig:teaser}, we know that \figtwopersonthree~will serve \figtwopersonone~pancakes, whereas a machine might not understand this unless it sees the entire clip.

% The goal is to collect commonsense visual reasoning problems at scale, while ensuring diverse and high-quality inferences. 

\paragraph{Interesting and Diverse Situations} 
To ensure diversity, we make no limiting assumptions about the predefined set of actions. Rather than searching for predefined labels, which can introduce search engine bias \cite{torralba2011unbiased, Devlin2015ExploringNN,Fouhey18}, we collect images from movie scenes. The underlying scenes come from the Large Scale Movie Description Challenge \cite{rohrbach_movie_2017} and YouTube movie clips.\footnote{Namely, Fandango MovieClips: {\tt\href{https://youtube.com/user/movieclips}{youtube.com/user/movieclips}}.} To avoid simple images, we train and apply an `interestingness filter' (e.g. a closeup of a syringe in Figure~\ref{fig:ourdataapproach}).\footnote{We annotated images for `interestingness' and trained a classifier using CNN features and detection statistics, details in the appendix, Sec~\ref{sec:datasetcreationdetails}.} 

We center our task around challenging questions requiring cognition-level reasoning. To make these cognition-level questions simple to ask, and to avoid the clunkiness of referring expressions, \datasetname's language integrates object tags (\figonepersontwo)~and explicitly excludes referring expressions (`the woman on the right.') These object tags are detected from Mask-RCNN \cite{He2017MaskR, Detectron2018}, and the images are filtered so as to have at least three high-confidence tags.
% and given to crowd workers when annotating the data. \footnote{All images have at least three detections that workers can reference.}

% To make cognition-level questions simple to ask, 

% explicitly avoiding referring expressions. To do this, and to avoid the clunkiness of language 

% For example, instead of annotating a low-level action: (the girl in the middle 

% `the guy who is raising his hand.' we instead wish to capture the social commonsense: `he wants to ask a question'. To accomplish this, our interface provides annotators with object detections from Mask-RCNN \cite{He2017MaskR, Detectron2018}.\footnote{All images have at least three detections that workers can reference.} We allow workers to refer to detection tags (like~\figtwopersonthree) when annotating the data. Workers are also provided with context in the form of video captions, helping them ask about and answer what will happen next. 

% Workers can also ask about what happens next, as given by 

% Moreover, we provide winterorkers with extra video-level context,\footnote{For LSMDC, this consists of the previous, current, and next caption of the movie. For MovieClips, this is obtained by extracting the title of the movie clip, as well as its summary.} %to hint at what might happen next in the video. 
% to stimulate their imagination. 

\paragraph{Crowdsourcing Quality Annotations} Workers on Amazon Mechanical Turk were given an image with detections, along with additional context in the form of video captions.\footnote{This additional clip-level context helps workers ask and answer about what will happen next.} They then ask one to three questions about the image; for each question, they provide a reasonable answer and a rationale. To ensure top-tier work, we used a system of quality checks and paid our workers well.\footnote{More details in the appendix, Sec~\ref{sec:datasetcreationdetails}.}

The result is an underlying dataset with high agreement and diversity of reasoning. Our dataset contains a myriad of interesting commonsense phenomena (Figure~\ref{fig:questiondiversity}) and a great diversity in terms of unique examples (Supp Section~\ref{sec:datasetanalysis}); almost every answer and rationale is unique.

\section{Adversarial Matching}
We cast \datasetname~as a four-way multiple choice task, to avoid the evaluation difficulties of language generation or captioning tasks where current metrics often prefer incorrect machine-written text over correct human-written text \cite{lin2014microsoft}. %\footnote{Constructing new metrics is an active line of research \cite{chaganty2018price, cui2018learning}.}
However, it is not obvious how to obtain high-quality incorrect choices, or counterfactuals, at scale. While past work has asked humans to write several counterfactual choices for each correct answer \cite{tapaswi2016movieqa, lei2018tvqa}, this process is expensive. Moreover, it has the potential of introducing annotation artifacts: subtle patterns that are by themselves highly predictive of the `correct' or `incorrect' label \cite{Schwartz:2017,gururangan2018annotation, poliak_hypothesis_2018}. 
% Asking humans to produce these distractors \cite{lei2018tvqa,tapaswi2016movieqa} is expensive and can lead to the aforementioned annotation artifacts that models can exploit to predict the answer without looking at the image \cite{balanced_vqa_v2}.

\begin{figure}[t!]
    \vspace{-3mm}
    \centering
    \includegraphics[width=\columnwidth]{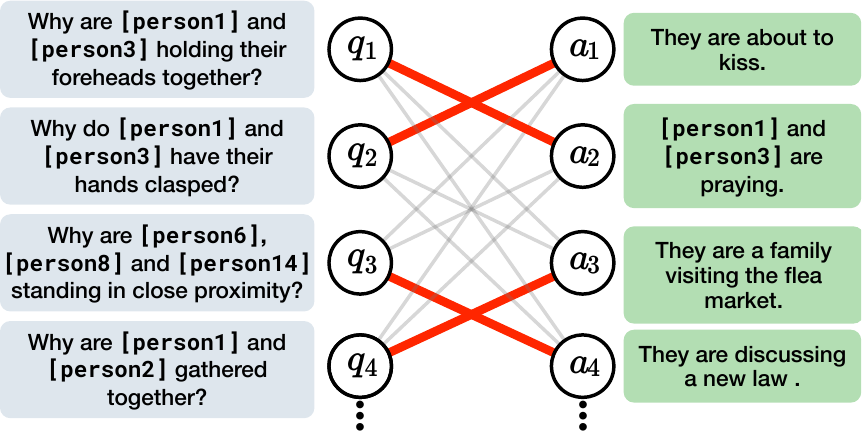}
    \vspace*{-6mm}
    \caption{Overview of \evaluationname. Incorrect choices are obtained via maximum-weight bipartite matching between queries and responses; the weights are scores from state-of-the-art natural language inference models. Assigned responses are highly relevant to the query, while they differ in meaning versus the correct responses.}
    %Models can hardly do well by exploiting biases in the responses, since each appears exactly once in the dataset as the right choice, and three times as a wrong choice.
    % }
    % To obtain the incorrect examples for \datasetname, we perform a maximum-weight bipartite matching between queries $\boldsymbol{q}$ and responses $\boldsymbol{r}$, where the weights come from state-of-the-art natural language inference models that score each pair. The result is a challenging dataset that is robust to ending bias. A text-only model can hardly do well given only responses: each appears exactly once in the dataset (as a positive ending) and three times as negative endings.} 
    \label{fig:matching}
\end{figure}

%To address this issue, we introduce \evaluationname, a new method to obtain counterfactuals. Our key insight is that the problem of obtaining good counterfactuals can 
In this work, we propose \evaluationname: a new method that allows for any `language generation' dataset to be turned into a multiple choice test, while requiring minimal human involvement. An overview is shown in Figure~\ref{fig:matching}. Our key insight is that the problem of obtaining good counterfactuals can be broken up into two subtasks: the counterfactuals must be as \textbf{relevant} as possible to the context (so that they appeal to machines), while they cannot be overly \textbf{similar} to the correct response (so that they don't become correct answers incidentally). We balance between these two objectives to create a dataset that is challenging for machines, yet easy for humans.

Formally, our procedure requires two models: one to compute the relevance between a query and a response, $P_{rel}$, and another to compute the similarity between two response choices, $P_{sim}$. Here, we employ state-of-the-art models for Natural Language Inference: BERT \cite{devlin2018bert} and ESIM+ELMo \cite{chen2017enhanced, peters2018deep}, respectively.\footnote{We finetune $P_{rel}$ (BERT), on the annotated data (taking steps to avoid data leakage), whereas $P_{sim}$ (ESIM+ELMo) is trained on entailment and paraphrase data - details in appendix Sec~\ref{sec:matchingdetails}.} Then, given %$N$
dataset examples $(\boldsymbol{q}_i, \boldsymbol{r}_i)_{1{\le}i{\le}N}$, we obtain a counterfactual for each $\boldsymbol{q}_i$ by performing maximum-weight bipartite matching \cite{munkres1957algorithms, jonker1987shortest} on a weight matrix $\mathbf{W} \in \mathbb{R}^{N \times N}$, given by
{\setlength{\abovedisplayskip}{10pt}
\setlength{\belowdisplayskip}{10pt}
\setlength{\abovedisplayshortskip}{0pt}
\setlength{\belowdisplayshortskip}{0pt}
\begin{equation}
\label{eqn:matching}
\mathbf{W}_{i,j} = \log(P_{rel}(\boldsymbol{q}_{i}, \boldsymbol{r}_{j})) + \lambda\log(1-P_{sim}(\boldsymbol{r}_{i}, \boldsymbol{r}_{j})).
\end{equation}
}
Here, $\lambda{>}0$ controls the tradeoff between similarity and relevance.\footnote{We tuned this hyperparameter by asking crowd workers to answer multiple-choice questions at several thresholds, and chose the value for which human performance is above $90\%$ - details in appendix Sec~\ref{sec:matchingdetails}.} To obtain multiple counterfactuals, we perform several bipartite matchings. To ensure that the negatives are diverse, during each iteration we replace the similarity term with the maximum similarity between a candidate response $\boldsymbol{r}_{j}$ and all responses currently assigned to $\boldsymbol{q}_{i}$.
% To obtain multiple counterfactuals, we perform another bipartite matching,only replacing the similarity term with the maximum similarity across answers already assigned to the question. 
% We iteratively apply maximum-weight bipartite matching on $\mathbf{W}$ \cite{munkres1957algorithms, jonker1987shortest} to obtain a new negative answer. %\footnote{We use the \href{LAP}{https://github.com/gatagat/lap} implementation.} 
% $\lambda{>}0$ is tuned to ensure the similarity penalty only fires for confident entailment predictions.
% To ensure diverse negatives, each iteration replaces the similarity term with the maximum similarity across answers already assigned to the question. 

\begin{figure*}[t]
    \vspace{-3mm}
    \centering
    \includegraphics[width=\linewidth]{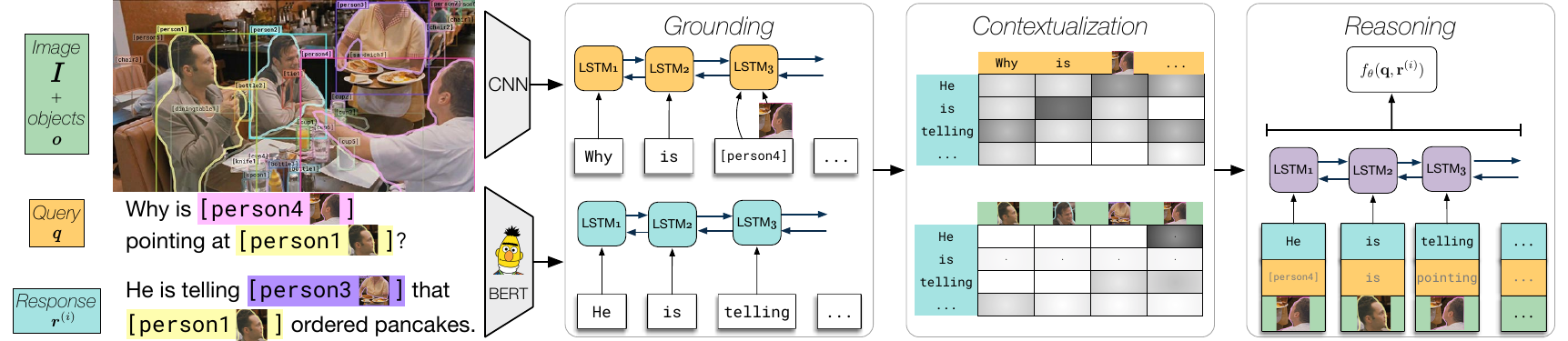}
    \vspace*{-3mm}
    \caption{High-level overview of our model, \modelname. We break the challenge of Visual Commonsense Reasoning into three components: grounding the query and response, contextualizing the response within the context of the query and the entire image, and performing additional reasoning steps on top of this rich representation.} 
    \label{fig:ourmodel}
    \vspace{-1mm}
\end{figure*}

\paragraph{Ensuring dataset integrity} To guarantee that there is no question/answer overlap between the training and test sets, we split our full dataset (by movie) into 11 folds. We match the answers and rationales invidually for each fold. Two folds are pulled aside for validation and testing. 

\section{Recognition to Cognition Networks}
We introduce~\modelnamelong~(\modelname), a new model for visual commonsense reasoning. To perform well on this task requires a deep understanding of language, vision, and the world. For example, in Figure~\ref{fig:ourmodel}, answering `\textit{Why is~\figtwopersonfour~pointing at~\figtwopersonone?}' requires multiple inference steps. First, we \textbf{ground} the meaning of the query and each response, which involves referring to the image for the two people. Second, we \textbf{contextualize} the meaning of the query, response, and image together. This step includes resolving the referent `he,' and why one might be pointing in a diner. Third, we \textbf{reason} about the interplay of relevant image regions, the query, and the response. In this example, the model must determine the social dynamics between~\figtwopersonone~and~\figtwopersonfour. We formulate our model as three high-level stages: grounding, contextualization, and reasoning, and use standard neural building blocks to implement each component.
% to know that the right answer to the query \texttt{What did [person3] order?} is that \texttt{[person3] ordered pancakes and bacon}, we must take multiple inference steps. First, we must \emph{ground} the meaning of each sentence, which requires referring to the image for \texttt{[person3]}. Second, we must \emph{contextualize} the meaning of the response, with respect to the question - at least to verify that \texttt{[person3]} refers to the same person as mentioned by the question. Third, we might have to perform additional inference - in this example, this might involve looking at \texttt{[person2]}, who is pointing at \texttt{[person3]}. In turn, we divide our model into three high-level pieces, for each of grounding, contextualization, and reasoning. Our modules are built using standard neural architecture - mainly, CNNs, LSTMs and attention. Moreover, we also use state-of-the-art BERT representations for language \cite{devlin2018bert}.

In more detail, recall that a model is given an image, a set of objects $\boldsymbol{o}$, a query $\boldsymbol{q}$, and a set of responses $\boldsymbol{r}^{(i)}$ (of which exactly one is correct). The query $\boldsymbol{q}$ and response choices $\boldsymbol{r}^{(i)}$ are all expressed in terms of a mixture of natural language and pointing to image regions: notation-wise, we will represent the object tagged by a word $w$ as $o_w$. If $w$ isn't a detection tag, $o_w$ refers to the entire image boundary. Our model will then consider each response $\boldsymbol{r}$ separately, using the following three components:

\paragraph{Grounding}
The grounding module will learn a joint image-language representation for each token in a sequence. Because both the query and the response contain a mixture of tags and natural language words, we apply the same grounding module for each (allowing it to share parameters). At the core of our grounding module is a bidirectional LSTM \cite{Hochreiter:1997:LSM:1246443.1246450} which at each position is passed as input a word representation for $w_i$, as well as visual features for $o_{w_i}$. We use a CNN to learn object-level features: the visual representation for each region $o$ is Roi-Aligned from its bounding region \cite{ren2015faster,He2017MaskR}. To additionally encode information about the object's class label $\ell_o$, we project an embedding of $\ell_o$ (along with the object's visual features) into a shared hidden representation. Let the output of the LSTM over all positions be $\mathbf{r}$, for the response and $\mathbf{q}$ for the query.
% We use a CNN to learn object features: the visual representations for each region $o_i$ are Roi-Aligned from the corresponding bounding region. To additionally encode information about the object's class label $\ell_i$, we project an embedding of $\ell_i$ along with the visual features, into a shared hidden representation.
%  Visual features are vectors from a shared hidden space learned by
% jointly embedding object class labels and object image features from Faster-RCNN \cite{ren2015faster,He2017MaskR}.  As each word is passed to the Grounding LSTM, if the current word is a detection tag, we pass the merged object CNN featured, otherwise global features for the entire image.  The LSTM is run separately for the query and each response.

\paragraph{Contextualization}
Given a grounded representation of the query and response, we use attention mechanisms to contextualize these sentences with respect to each other and the image context. For each position $i$ in the response, we will define the attended query representation as $\hat{\mathbf{q}}_{i}$ using the following equation:
{\setlength{\abovedisplayskip}{5pt}
\setlength{\belowdisplayskip}{5pt}
\setlength{\abovedisplayshortskip}{0pt}
\setlength{\belowdisplayshortskip}{0pt}
\begin{equation}
\label{eqn:context}
    \alpha_{i,j} = \softmax_j(\mathbf{r}_i\mathbf{W}\mathbf{q}_j) \qquad \hat{\mathbf{q}}_{i} = \sum_{j} \alpha_{i,j}\mathbf{q}_j.
\end{equation}
}
To contextualize an answer with the image, including implicitly relevant objects that have not been picked up from the grounding stage, we perform another bilinear attention between the response $\mathbf{r}$ and each object $o$'s image features. Let the result of the object attention be $\hat{\mathbf{o}}_i$.

\paragraph{Reasoning}
Last, we allow the model to \emph{reason} over the response, attended query and objects. We accomplish this using a bidirectional LSTM that is given as context $\hat{\mathbf{q}}_i$, $\mathbf{r}_i$, and $\hat{\mathbf{o}}_i$ for each position $i$. For better gradient flow through the network, we concatenate the output of the reasoning LSTM along with the question and answer representations for each timestep: the resulting sequence is max-pooled and passed through a multilayer perceptron, which predicts a logit for the query-response compatibility.

\paragraph{Neural architecture and training details}
For our image features, we use ResNet50 \cite{he2016deep}. To obtain strong representations for language, we used BERT representations \cite{devlin2018bert}. BERT is applied over the entire question and answer choice, and we extract a feature vector from the second-to-last layer for each word. %\footnote{Since BERT was pretrained on text quite different from our domain, we finetune on \datasetname~for one epoch to account for the domain shift.}
We train \modelname~by minimizing the multi-class cross entropy between the prediction
%$\textrm{softmax}(\mathbf{s})$ 
for each response $\boldsymbol{r}^{(i)}$, and the gold label. See the appendix (Sec~\ref{sec:modeldetails}) for detailed training information and hyperparameters.\footnote{Our code is also available online at \href{visualcommonsense.com}{visualcommonsense.com}.} %We train the model end-to-end using Adam %(learning rate of $2\cdot 10^{-4}$) and a batch size of 96. More details are in the supplemental material and the code will be made available online.\footnote{\url{anonymous.link}}

\section{Results}
\newcommand{\nodata}[1]{\multicolumn{#1}{|c|}{\cellcolor{gray!10}}}
\newcommand{\nodataright}[1]{\multicolumn{#1}{|c}{\cellcolor{gray!10}}}
\newcommand{\nd}{\cellcolor{gray!10}}

\newcommand{\tinyrule}{\cline{2-3} \\[-1.0em]}

\newcommand{\resultswidth}{1.35cm}
\begin{table}[t!]
\vspace{-3mm}
\centering
\begin{small}
\setlength{\tabcolsep}{4pt}
\begin{tabular}{@{} c@{\hspace{0.4em}} l @{\hspace{0.7em}}|
l@{\hspace{0.7em}}l@{\hspace{0.5em}} |l@{\hspace{0.7em}}l  |l@{\hspace{0.7em}}l@{}}
&\multicolumn{1}{c}{} & \multicolumn{2}{c}{$Q \rightarrow A$} & \multicolumn{2}{c}{$QA \rightarrow R$} & \multicolumn{2}{c}{$Q \rightarrow AR$}\\ 
\multicolumn{2}{c|}{Model} & Val & Test & Val & Test & Val & Test \\ 
\toprule
& Chance & 25.0 & 25.0 & 25.0 & 25.0 & \phantom{0}6.2 &  \phantom{0}6.2 \\ %\spacedhline
\midrule
\multirow{4}{*}{\rotatebox[origin=c]{90}{Text Only}}& BERT & 53.8 & 53.9 & 64.1 & 64.5 & 34.8 & 35.0 \\ 
& BERT (response only) & 27.6 & 27.7 & 26.3 & 26.2 & \phantom{0}7.6 & \phantom{0}7.3 \\ 
& ESIM+ELMo & 45.8 & 45.9 & 55.0 & 55.1 & 25.3 & 25.6 \\ 
& LSTM+ELMo & 28.1 & 28.3 &  28.7 & 28.5 & \phantom{0}8.3 & \phantom{0}8.4 \\ \midrule
\multirow{4}{*}{\rotatebox[origin=c]{90}{VQA}} 
% & BottomUp+BERT & \multicolumn{2}{c}{blah} & 63.0 & 62.9 & \multicolumn{2}{c}{blah} \\ 
& RevisitedVQA \cite{jabri2016revisiting} & 39.4 & 40.5 & 34.0 & 33.7 & 13.5 & 13.8 \\ 
& BottomUpTopDown\cite{Anderson2017updown} & 42.8 & 44.1 & 25.1 & 25.1 & 10.7 & 11.0 \\  
& MLB \cite{Kim2017} & 45.5 & 46.2 & 36.1 & 36.8 & 17.0 & 17.2 \\ 
& MUTAN \cite{Ben-younes_2017_ICCV} & 44.4 & 45.5 & 32.0 & 32.2 & 14.6 & 14.6 \\ \midrule
& \modelname &\bf{63.8} & \bf{65.1} & \bf{67.2} & \bf{67.3} & \bf{43.1} & \bf{44.0} \\ \midrule
% & One turker & & 86.7 & & 87.6 & & 76.0 \\ 
% & Three turkers & & 90.4 & & 91.2 & & 82.9 \\ 
& Human & & 91.0 & & 93.0 & & 85.0 \\ 
\bottomrule
\end{tabular}
\end{small}
\vspace*{-1mm}\caption{Experimental results on \datasetname. VQA models struggle on both question-answering ($Q \rightarrow A$) as well as answer justification ($Q \rightarrow AR$), possibly due to the complex language and diversity of examples in the dataset. While language-only models perform well, our model~\modelname~obtains a significant performance boost. Still, all models underperform human accuracy at this task. For more up-to-date results, see the leaderboard at \leaderboardlink.}\vspace*{-1mm}
\label{tab:results}
\end{table}
In this section, we evaluate the performance of various models on \datasetname. Recall that our main evaluation mode is the staged setting ($Q{\rightarrow}AR$). Here, a model must choose the right answer for a question (given four answer choices), and then choose the right rationale for that question and answer (given four rationale choices). If it gets either the answer or the rationale wrong, the entire prediction will be wrong. This holistic task decomposes into two sub-tasks wherein we can train individual models: question answering ($Q{\rightarrow}A$) as well as answer justification ($QA{\rightarrow}R$). Thus, in addition to reporting combined $Q{\rightarrow}AR$ performance, we will also report $Q{\rightarrow}A$ and $QA{\rightarrow}R$.
%%%%%%%%%% TODO

% Our goal is for models to not only make accurate commonsense inferences about the world, but to be able to justify their inferences in natural language. To achieve this requires a model to perform well on both modes in \datasetname: question answering ($Q \to A$) as well as answer justification ($QA \rightarrow R$). Thus, we also evaluate models' joint accuracy $Q\rightarrow AR$. In this mode, a model must choose the right answer for a question (given four answer choices), and then choose the right rationale for that question and answer (given four rationale choices). If it gets either the answer or the rationale wrong, it receives no points.

\paragraph{Task setup} A model is presented with a query $\boldsymbol{q}$, and four response choices $\boldsymbol{r}^{(i)}$. Like our model, we train the baselines using multi-class cross entropy between the set of responses and the label. Each model is trained separately for question answering and answer justification.\footnote{We follow the standard train, val and test splits.}
% Each model $f_\theta$ is trained to choose the response that matches the ground truth:
% \begin{equation}
%     \hat{y} = \argmax_i f_\theta(\boldsymbol{q}, \boldsymbol{r}^{(i)})
% \end{equation}
%This objective allows us to train our models, including baselines, using multi-class cross-entropy. 
% For simplicity, we train separate models for question answering and answer justification.\footnote{We follow the standard train, val and test splits.}
% We tune on the validation set and only run models on the test set once to report final results. Train, val, and test splits will be released along with the dataset.}

\subsection{Baselines}
We compare our \modelname~to several strong language and vision baselines.
\paragraph{Text-only baselines}
We evaluate the level of visual reasoning needed for the dataset by also evaluating purely text-only models. For each model, we represent $\boldsymbol{q}$ and $\boldsymbol{r}^{(i)}$ as streams of tokens, with the detection tags replaced by the object name (e.g. \texttt{chair5} $\rightarrow$ \texttt{chair}). To minimize the discrepancy between our task and pretrained models, we replace person detection tags with gender-neutral names.
\begin{enumerate}[wide, labelwidth=!,labelindent=0pt,noitemsep,topsep=0pt,label=\textbf{\alph*}.]
\item {\bf BERT} \cite{devlin2018bert}: BERT is a recently released NLP model that achieves state-of-the-art performance on many NLP tasks. %BERT is fine-tuned on \datasetname~for three epochs.
\item {\bf BERT (response only)} We use the same BERT model, however, during fine-tuning and testing the model is only given the response choices $\boldsymbol{r}^{(i)}$. 
\item {\bf ESIM+ELMo} \cite{chen2017enhanced}: ESIM is another high performing model for sentence-pair classification tasks, particularly when used with ELMo embeddings \cite{peters2018deep}. 
\item {\bf LSTM+ELMo}: Here an LSTM with ELMo embeddings is used to score responses $\boldsymbol{r}^{(i)}$. 
\end{enumerate}

\paragraph{VQA Baselines}
Additionally we compare our approach to models developed on the VQA dataset \cite{antol2015vqa}. All models use the same visual backbone as \modelname~(ResNet 50) as well as text representations (GloVe; \cite{pennington2014glove}) that match the original implementations.
\begin{enumerate}[wide, labelwidth=!,labelindent=0pt,noitemsep,topsep=0pt,label=\textbf{\alph*}.]
\setcounter{enumi}{4}
\item {\bf RevisitedVQA} \cite{jabri2016revisiting}: This model takes as input a query, response, and image features for the entire image, and passes the result through a multilayer perceptron, which has to classify `yes' or `no'.\footnote{For VQA, the model is trained by sampling positive or negative answers for a given question; for our dataset, we simply use the result of the perceptron (for response $\boldsymbol{r}^{(i)}$) as the $i$-th logit.} %Averaged GloVe embeddings as the question and answer representations. 
\item {\bf Bottom-up and Top-down attention} (BottomUpTopDown) \cite{Anderson2017updown}: This model attends over region proposals given by an object detector. To adapt to \datasetname, we pass this model object regions referenced by the query and response.
\item {\bf Multimodal Low-rank Bilinear Attention} (MLB) \cite{Kim2017}: This model uses Hadamard products to merge the vision and language representations given by a query and each region in the image.
\item {\bf Multimodal Tucker Fusion} (MUTAN) \cite{Ben-younes_2017_ICCV}: This model expresses joint vision-language context in terms of a tensor decomposition, allowing for more expressivity.
\end{enumerate}

We note that BottomUpTopDown, MLB, and MUTAN all treat VQA as a multilabel classification over the top 1000 answers \cite{Anderson2017updown, lin2016leveraging}. Because \datasetname~is highly diverse (Supp~\ref{sec:datasetanalysis}), for these models we represent each response $\boldsymbol{r}^{(i)}$ using a GRU \cite{cho2014learning}.\footnote{To match the other GRUs used in \cite{Anderson2017updown, Kim2017, Ben-younes_2017_ICCV} which encode $\boldsymbol{q}$.} The output logit for response $i$ is given by the dot product between the final hidden state of the GRU encoding $\boldsymbol{r}^{(i)}$, and the final representation from the model.

\begin{table}[t!]
\vspace{-3mm}
\centering
% \begin{normal}
\setlength{\tabcolsep}{4pt}
\begin{tabular}{@{} c@{\hspace{0.4em}} l @{\hspace{0.1em}}|
l@{\hspace{0.5em}} |l@{\hspace{0.5em}}  |l @{}}
& Model & $Q\rightarrow A$ & $QA\rightarrow R$ & $Q\rightarrow AR$\\ \toprule
& \modelname & \textbf{63.8} & \textbf{67.2} & \textbf{43.1} \\ \toprule
& No query & 48.3 & 43.5 & 21.5 \\ %\spacedhline
& No reasoning module & 63.6 & 65.7 & 42.2 \\ %\spacedhline
& No vision representation & 53.1 & 63.2 & 33.8 \\ %\spacedhline
& GloVe representations & 46.4 & 38.3 & 18.3 \\ %\spacedhline
\bottomrule
\end{tabular}
% \end{small}
\vspace*{-2mm}\caption{Ablations for \modelname, over the validation set. `No query' tests the importance of integrating the query during contextualization; removing this reduces $Q{\rightarrow}AR$ performance by 20\%. In `no reasoning', the LSTM in the reasoning stage is removed; this hurts performance by roughly 1\%. Removing the visual features during grounding, or using GloVe embeddings rather than BERT, lowers performance significantly, by 10\% and 25\% respectively.}\vspace*{-2mm}
\label{tab:ablations}
\end{table}

\newcolumntype{P}[1]{>{\RaggedRight\hspace{0pt}}p{#1}}
\definecolor{correctrow}{HTML}{C8E7FF}
\definecolor{incrow}{HTML}{C89A9D}

% pred answer, correct answer, this answer ID, the answer
\newcommand{\maybecolor}[4]{%
  \ifthenelse{\equal{#2}{#3}}
    {\ifthenelse{\equal{#1}{#3}}{\cellcolor{correctrow!50}\textbf{#4}}{\textbf{#4}}}%
    {\ifthenelse{\equal{#1}{#3}}{\cellcolor{incrow!50}#4}{#4}}%
}

%q, 4as, pred, gt
\newcommand{\aquestionqa}[7]{\raisebox{1mm}{{\footnotesize\renewcommand{\arraystretch}{1.5}\setlength{\tabcolsep}{0.1em} \begin{tabular}[t]{@{}l@{\hspace{0.005\textwidth}}@{}P{0.34\textwidth}@{}}
\multicolumn{2}{l}{\hspace{-.005\textwidth}\cellcolor{gray!01}\parbox{0.33\textwidth}{{\small \RaggedRight #1}}} \\ \midrule
&\maybecolor{#6}{#7}{1}{#2}\\
&\maybecolor{#6}{#7}{2}{#3}\\
&\maybecolor{#6}{#7}{3}{#4}\\ 
&\maybecolor{#6}{#7}{4}{#5}\\
\end{tabular}
}}}

\newcommand{\aquestionqar}[7]{\raisebox{3mm}{{\footnotesize\renewcommand{\arraystretch}{1.3}\setlength{\tabcolsep}{0.1em}\begin{tabular}[t]{@{}l@{\hspace{0.005\textwidth}}@{}P{0.34\textwidth}@{}}
\multicolumn{2}{l}{\hspace{-.005\textwidth}\cellcolor{gray!01}\parbox{0.315\textwidth}{{\small \RaggedRight #1}}} \vspace*{-1mm}\\ \midrule
&\maybecolor{#6}{#7}{1}{#2}\\
&\maybecolor{#6}{#7}{2}{#3}\\
&\maybecolor{#6}{#7}{3}{#4}\\ 
&\maybecolor{#6}{#7}{4}{#5}\\
\end{tabular}
}}}

\newcommand{\qualpersontag}[3]{\setlength{\fboxsep}{1pt}\colorbox{#1}{\scriptsize \usefont{T1}{robotomono}{m}{n}{[#2\hspace{1pt}\inlineimage{#3}\hspace{1pt}]}}}

\definecolor{ex1person1}{RGB}{255,151,227}
\definecolor{ex1person2}{RGB}{151, 212, 118}
\definecolor{ex1person3}{RGB}{123, 156, 192}
\newcommand{\exonepersonone}{\qualpersontag{ex1person1!80}{person1}{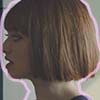}}
\newcommand{\exonepersontwo}{\qualpersontag{ex1person2!80}{person2}{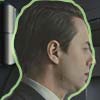}}
\newcommand{\exonepersonthree}{\qualpersontag{ex1person3!80}{person3}{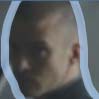}}

\definecolor{ex2person1}{RGB}{174,139,136}
\definecolor{ex2person2}{RGB}{184, 255, 153}
\definecolor{ex2person3}{RGB}{149, 152, 255}
\definecolor{ex2book1}{RGB}{255, 254, 255}

\newcommand{\extwopersonone}{\qualpersontag{ex2person1!80}{person1}{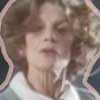}}
\newcommand{\extwopersontwo}{\qualpersontag{ex2person2!80}{person2}{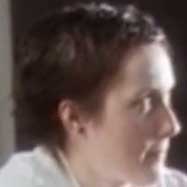}}
\newcommand{\extwopersonthree}{\qualpersontag{ex2person3!80}{person3}{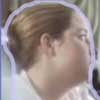}}
\newcommand{\extwobookone}{\qualpersontag{ex2book1!80}{book1}{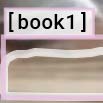}}

\definecolor{ex3person1}{RGB}{255,151,227}
\definecolor{ex3person2}{RGB}{151, 212, 118}
\definecolor{ex3person3}{RGB}{123, 156, 192}
\definecolor{ex3person4}{RGB}{248, 162, 107}
\definecolor{ex3person5}{RGB}{223, 255, 255}

\newcommand{\exthreepersontwo}{\qualpersontag{ex3person2!80}{person2}{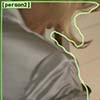}}
\newcommand{\exthreepersonthree}{\qualpersontag{ex3person3!80}{person3}{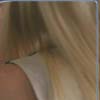}}
\newcommand{\exthreepersonfour}{\qualpersontag{ex3person4!80}{person4}{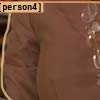}}
\newcommand{\exthreepersonfive}{\qualpersontag{ex3person5!80}{person5}{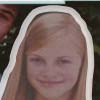}}

\newcommand{\exthreebigpersontwo}{\persontag{ex3person2!80}{person2}{qualex/ex3_person2.jpg}}
\newcommand{\exthreebigpersonthree}{\persontag{ex3person3!80}{person3}{qualex/ex3_person3.jpg}}
\newcommand{\exthreebigpersonfour}{\persontag{ex3person4!80}{person4}{qualex/ex3_person4.jpg}}
\newcommand{\exthreebigpersonfive}{\persontag{ex3person5!80}{person5}{qualex/ex3_person5.jpg}}

\definecolor{ex4couch1}{RGB}{255,224,255}
\definecolor{ex4dog1}{RGB}{250,255,124}
\definecolor{ex4person1}{RGB}{168,121,133}
\definecolor{ex4person2}{RGB}{123,255,178}
\definecolor{ex4person3}{RGB}{255,112,184}
\definecolor{ex4pottedplant6}{RGB}{203,193,194}

\newcommand{\exfourcouch}{\qualpersontag{ex4couch1!80}{couch1}{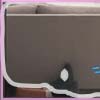}}
\newcommand{\exfourdog}{\qualpersontag{ex4dog1!80}{dog1}{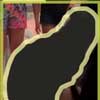}}
\newcommand{\exfourplant}{\qualpersontag{ex4pottedplant6!80}{pottedplant6}{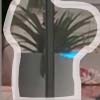}}
\newcommand{\exfourpersonone}{\qualpersontag{ex4person1!80}{person1}{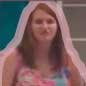}}
\newcommand{\exfourpersontwo}{\qualpersontag{ex4person2!80}{person2}{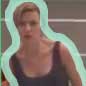}}
\newcommand{\exfourpersonthree}{\qualpersontag{ex4person3!80}{person3}{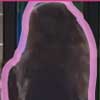}}

\newcommand{\qualimg}[1]{\raisebox{5mm}{\adjincludegraphics[padding=0ex 0ex 0ex 0ex,margin=0ex 0ex 0ex 0ex,valign=t,width=0.28\textwidth]{#1}}}

\begin{figure*}[t!]
    \vspace{-4mm}
    \centering\setlength\tabcolsep{1pt}
    \begin{tabular}[t]{@{}p{0.29\textwidth}@{}p{0.35\textwidth}@{\hspace{0.01\textwidth}}p{0.35\textwidth}@{}}%
\qualimg{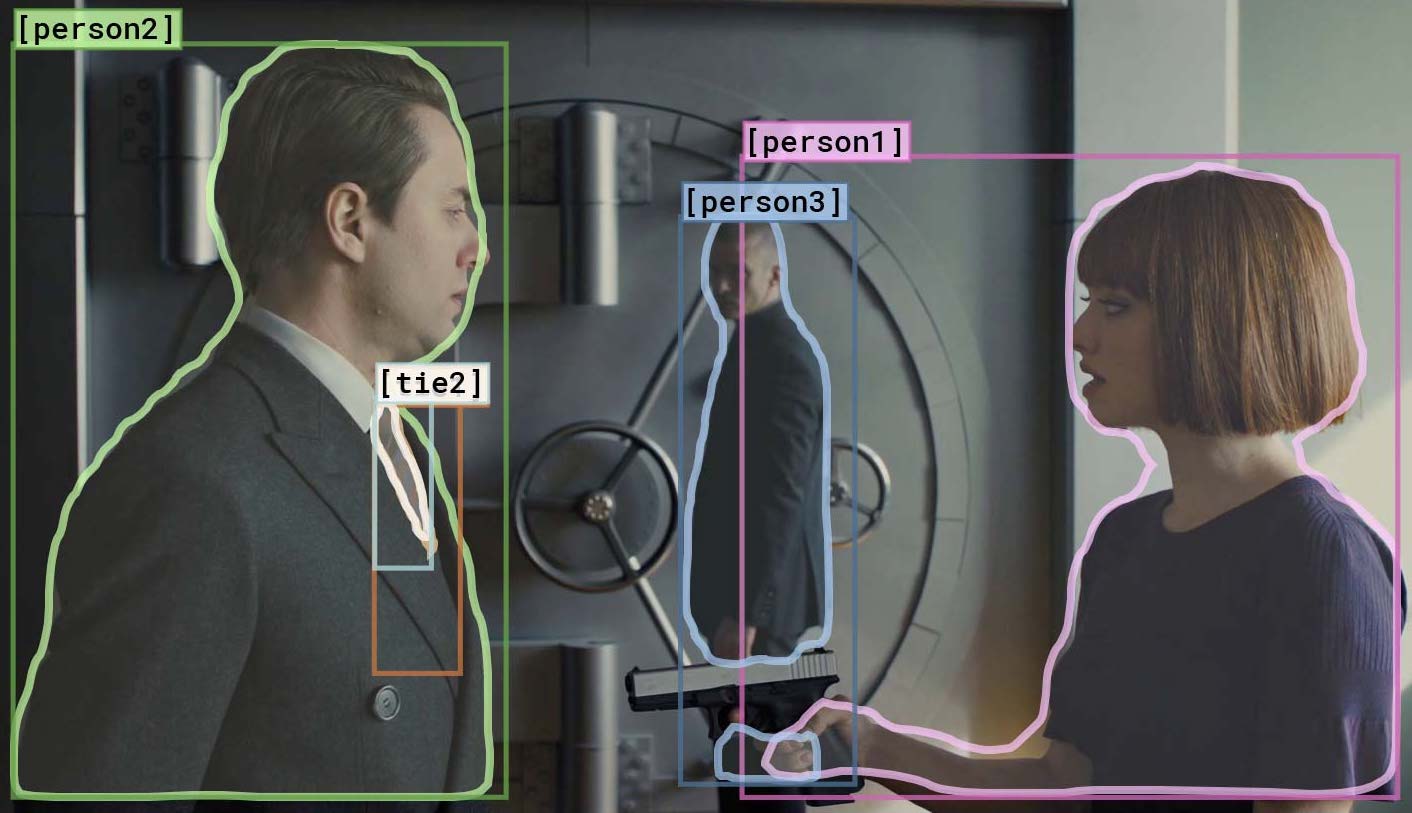}&\aquestionqa{Why is \exonepersonone~pointing a gun at \exonepersontwo?}{a)\exonepersonone~wants to kill \exonepersontwo.(1\%)}{b) \exonepersonone~and \exonepersonthree~are robbing the bank and \exonepersontwo~is the bank manager. (71\%)}{c) \exonepersontwo has done something to upset \exonepersonone. (18\%)}{d) Because \exonepersontwo~is \exonepersonone's daughter. \exonepersonone~wants to protect \exonepersontwo. (8\%)}{2}{2}&\aquestionqar{\emph{b) is right because...}}{a) \exonepersonone~is chasing \exonepersonone~and \exonepersonthree~ because they just robbed a bank. (33\%)}{b) Robbers will sometimes hold their gun in the air to get everyone's attention. (5\%)}{c) The vault in the background is similar to a bank vault. \exonepersonthree~is waiting by the vault for someone to open it. (49\%)}{d) A room with barred windows and a counter usually resembles a bank. (11\%)}{3}{3}\\ \midrule%
\qualimg{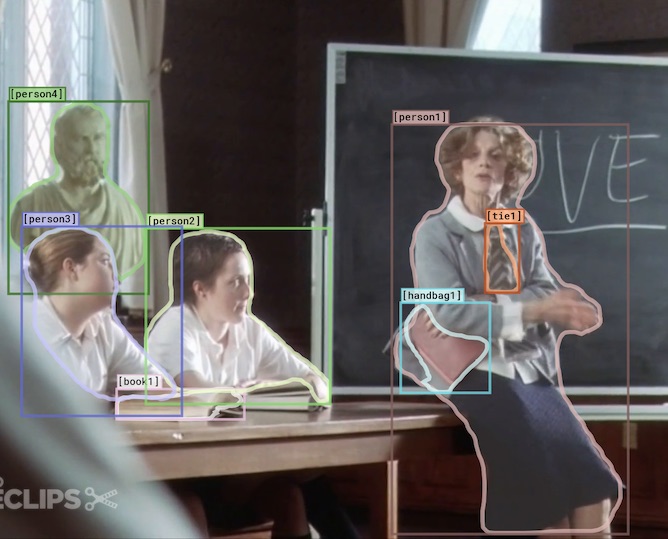}&\aquestionqa{What would \extwopersonone~do if she caught \extwopersontwo~and\extwopersonthree~whispering?}{a) \extwopersonone~would look to her left. (7\%)}{b) She would play with \extwobookone. (7\%)}{c) She would look concerned and ask what was funny. (39\%)}{d) She would switch their seats. (45\%)}{4}{4}&\aquestionqar{\emph{d) is right because...}}{a) When students are talking in class they're supposed to be listening - the teacher separates them. (64\%)}{b) Plane seats are very cramped and narrow, and it requires cooperation from your seat mates to help get through. (15\%)}{c) It's not unusual for people to want to get the closest seats to a stage. (14\%)}{d) That's one of the only visible seats I can see that's still open, the plane is mostly full. (6\%)}{1}{1} \\ \midrule%
\qualimg{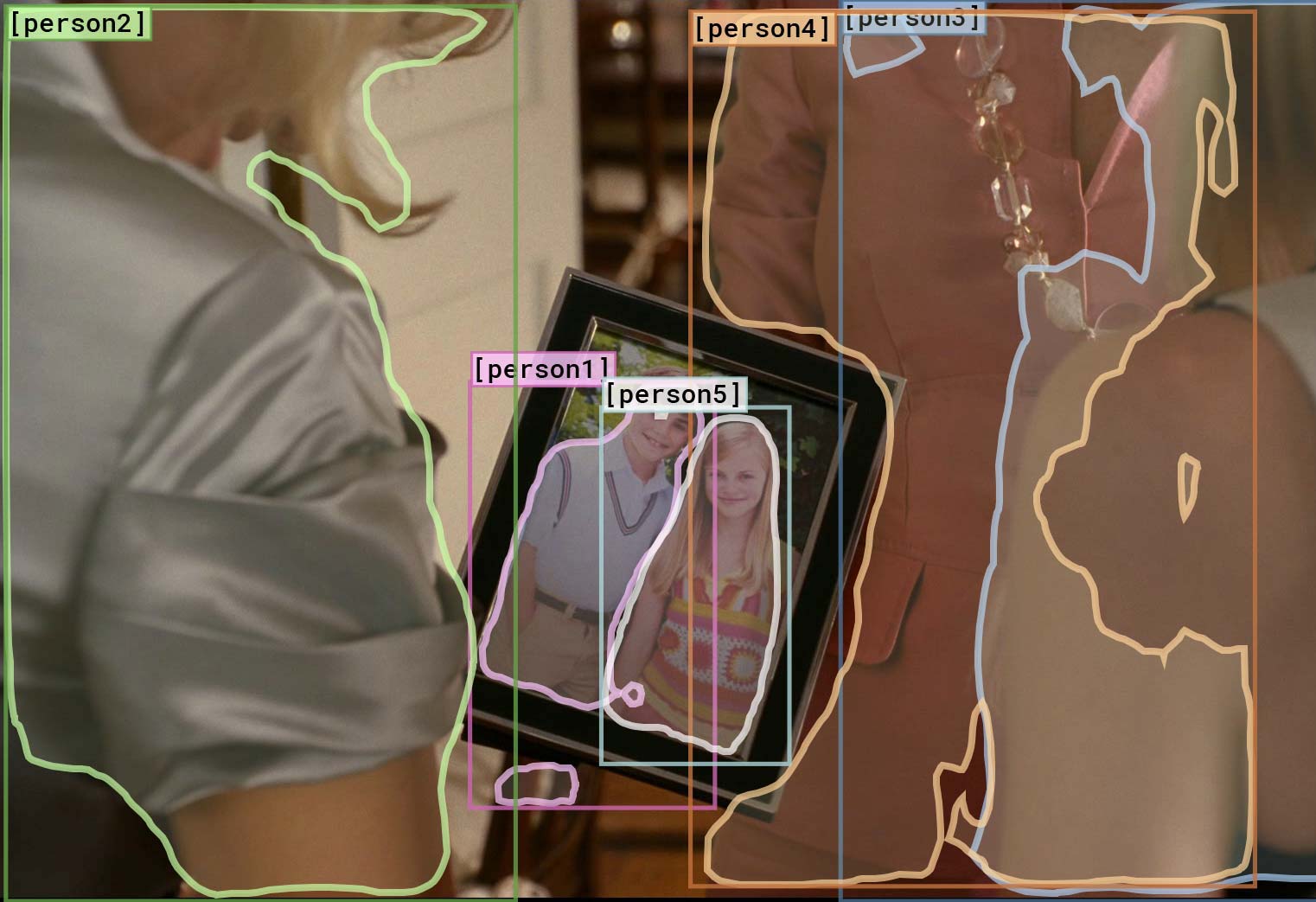}&\raisebox{2mm}{\aquestionqa{What's going to happen next?}{a) \exthreepersontwo~is going to walk up and punch \exthreepersonfour~in the face. (10\%)}{b) Someone is going to read \exthreepersonfour~a bedtime story.  (15\%)}{c) \exthreepersontwo~is going to fall down. (5\%)}{d) \exthreepersontwo~is going to say how cute \exthreepersonfour's children are. (68\%)}{4}{4}}&\aquestionqar{\emph{d) is right because...}}{a) They are the right age to be father and son and \exthreepersonfive~is hugging \exthreepersonthree~like they are his son. (1\%)}{b) It looks like \exthreepersonfour~is showing the photo to \exthreepersontwo, and \exthreepersontwo~will want to be polite. (31\%)}{c) \exthreepersontwo~is smirking and looking down at \exthreepersonfour. (6\%)}{d) You can see \exthreepersonfour~smiling and facing the crib and decor in the room (60\%)}{4}{2} \\ \midrule%
\qualimg{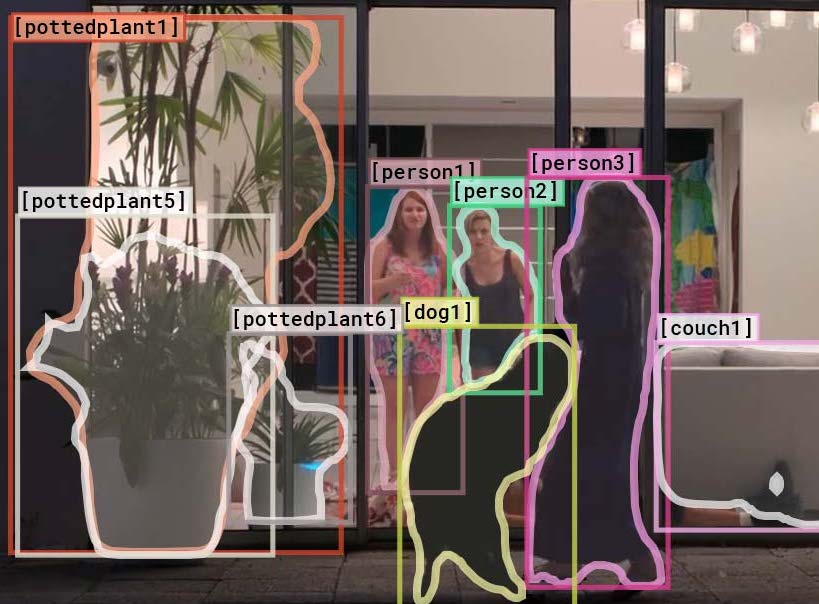}&\aquestionqa{Why can't \exfourpersonthree~go in the house with \exfourpersonone~and \exfourpersontwo?}{a) She does not want to be there. (12\%)}{b) \exfourpersonthree~has \exfourdog~with her. (14\%)}{c) She needs the light. (45\%)}{d) She is too freaked out (26\%)}{3}{2}&\aquestionqar{\emph{b) is right because...}}{a) \exfourpersonone~is going away by himself. (60\%)}{b) \exfourdog~is small enough to carry. \exfourpersonthree~appears to own him. (33\%)}{c) If \exfourdog~was in the house, he would likely knock over \exfourplant~and likely scratch \exfourcouch. (4\%)}{d) \exfourpersonone~looks like he may have lead \exfourpersontwo into the room to see\exfourdog.(1\%)}{1}{3}  \\ %\bottomrule
    \end{tabular}
    \vspace*{-3mm}
    \caption{Qualitative examples from \modelname. Correct predictions are \colorbox{correctrow}{\textbf{highlighted in blue}}. Incorrect predictions are \colorbox{incrow}{in red} with the correct choices \textbf{bolded}.\com{ In the first two rows, the model correctly chooses the right answer and rationale. In the third, it chooses the wrong rationale (hallucinating a crib), in the fourth, it also chooses the wrong answer.} For more predictions, see see {\tt \href{https://visualcommonsense.com/explore}{visualcommonsense.com/explore}}.} 
    \label{fig:qualexamples}
    \vspace{-3mm}
\end{figure*}
\paragraph{Human performance} We asked five different workers on Amazon Mechanical Turk to answer 200 dataset questions from the test set. A different set of five workers were asked to choose rationales for those questions and answers. Predictions were combined using a majority vote.
% \paragraph{Experimental setup} To ensure an apples to apples comparison, we do the same thing for each model... blah

\subsection{\vspace*{-.5mm}Results and Ablations}\vspace*{-1mm}
We present our results in Table~\ref{tab:results}. Of note, standard VQA models struggle on our task. The best model, in terms of $Q{\rightarrow}AR$ accuracy, is MLB, with 17.2\% accuracy. Deep text-only models perform much better: most notably, BERT \cite{devlin2018bert} obtains 35.0\% accuracy. One possible justification for this gap in performance is a bottlenecking effect: whereas VQA models are often built around multilabel classification of the top 1000 answers,  \datasetname~requires reasoning over two (often long) text spans. Our model, \modelname~obtains an additional boost over BERT by 9\% accuracy, reaching a final performance of 44\%. Still, this figure is nowhere near human performance: 85\% on the staged task, so there is significant headroom remaining.

\paragraph{Ablations}
We evaluated our model under several ablations to determine which components are most important. Removing the query representation (and query-response contextualization entirely) results  in a drop of 21.6\% accuracy points in terms of $Q\rightarrow AR$ performance. Interestingly, this setting allows it to leverage its image representation more heavily: the text based response-only models (BERT response only, and LSTM+ELMo) perform barely better than chance. Taking the reasoning module lowers performance by 1.9\%, which suggests that it is beneficial, but not critical for performance.  The model suffers most when using GloVe representations instead of BERT: a loss of 24\%. This suggests that strong textual representations are crucial to \datasetname~performance.

\paragraph{Qualitative results}
Last, we present qualitative examples in Figure~\ref{fig:qualexamples}. \modelname~works well for many images: for instance, in the first row, it correctly infers that a bank robbery is happening. Moreover, it picks the right rationale: even though all of the options have something to do with `banks' and `robbery,' only \textbf{c)} makes sense. Similarly, analyzing the examples for which \modelname~chooses the right answer but the wrong rationale allows us to gain more insight into its understanding of the world. In the third row, the model incorrectly believes there is a crib while assigning less probability mass on the correct rationale - that \exthreebigpersontwo~is being shown a photo of \exthreebigpersonfour's children, which is why \exthreebigpersontwo~might say how cute they are.

\section{Related Work}
\paragraph{Question Answering}
Visual Question Answering \cite{antol2015vqa} was one of the first large-scale datasets that framed visual understanding as a QA task, with questions about COCO images \cite{lin2014microsoft} typically answered with a short phrase. This line of work also includes `pointing' questions \cite{visualgenome, zhu2016cvpr} and templated questions with open ended answers \cite{yu_visual_2015}. Recent datasets also focus on knowledge-base style content \cite{Wang2017FVQAFV,Wu2016AskMA}. On the other hand, the answers in \datasetname~are entire sentences, and the knowledge required by our dataset is largely background knowledge about how the world works.

Recent work also includes movie or TV-clip based QA \cite{tapaswi2016movieqa, maharaj2017dataset, lei2018tvqa}. In these settings, a model is given a video clip, often alongside additional language context such as subtitles, a movie script, or a plot summary.\footnote{As we find in Appendix~\ref{sec:langpriors}, including additional language context tends to boost model performance.} In contrast, \datasetname~features no extra language context besides the question. Moreover, the use of explicit detection tags means that there is no need to perform person identification \cite{RohrbachCVPR2017a} or linkage with subtitles.

An orthogonal line of work has been on referring expressions: asking to what image region a natural language sentence refers to \cite{plummer2015flickr30k, mao2016generation, rohrbach2016grounding,yu2016modeling, yu2017joint, plummer2017phrase, hu2017modeling, hendricks17iccv}. We explicitly avoid referring expression-style questions by using indexed detection tags (like~\figonepersonone).

Last, some work focuses on commonsense phenomena, such as `what if' and `why' questions \cite{wagner18eccvw, pirsiavash2014inferring}. However, the space of commonsense inferences is often limited by the underlying dataset chosen (synthetic \cite{wagner18eccvw} or COCO \cite{pirsiavash2014inferring} scenes). In our work, we ask commonsense questions in the context of rich images from movies.

\paragraph{Explainability}
AI models are often right, but for questionable or vague reasons \cite{biran2017explanation}. This has motivated work in having models provide explanations for their behavior, in the form of a natural language sentence 
\cite{hendricks2016generating, chandrasekaran2018explanations, kim2018textual} or an attention map \cite{hendricks2018grounding, hu2018explainable, Park_2018_CVPR}. Our rationales combine the best of both of these approaches, as they involve both natural language text as well as references to image regions. Additionally, while it is hard to evaluate the quality of generated model explanations, choosing the right rationale in \datasetname~is a multiple choice task, making evaluation straightforward.

\paragraph{Commonsense Reasoning}
Our task unifies work involving reasoning about commonsense phenomena, such as physics \cite{mottaghi2016happens, ye2018interpretable}, social interactions \cite{alahi2016social, moviegraphs, ActProperly2018, Gupta_2018_CVPR}, procedure understanding \cite{Zhou2018TowardsAL, alayrac2016unsupervised} and predicting what might happen next in a video \cite{singh2016krishnacam, Ehsani_2018_CVPR, zhou2015temporal, vondrick2016anticipating, felsen2017will, rhinehart2017first, yoshikawa2018stair}.

\paragraph{Adversarial Datasets}
Past work has proposed the idea of creating adversarial datasets, whether by balancing the dataset with respect to priors \cite{goyal2017making, gururangan2018annotation, ramakrishnan2018overcoming} or switching them at test time \cite{agrawal2018don}. Most relevant to our dataset construction methodology is the idea of Adversarial Filtering \cite{zellers2018swagaf}.\footnote{This was used to create the \swag~dataset, a multiple choice NLP dataset for natural language inference.} Correct answers are human-written, while wrong answers are chosen from a pool of machine-generated text that is further validated by humans. However, the correct and wrong answers come from fundamentally different sources, which raises the concern that models can cheat by performing authorship identification rather than reasoning over the image. In contrast, in \evaluationname, the wrong choices come from the exact same distribution as the right choices, and no human validation is needed.

\section{Conclusion}
In this paper, we introduced Visual Commonsense Reasoning, along with a large dataset \datasetname~for the task that was built using \evaluationname. We presented \modelname, a model for this task, but the challenge -- of cognition-level visual undertanding -- is far from solved.

\vspace*{-3mm}
\section*{Acknowledgements}
\vspace*{-2mm}
{\footnotesize\setstretch{0.4}
We thank the Mechanical Turk workers for doing such an outstanding job with dataset creation - this dataset and paper would not exist without them. Thanks also to Michael Schmitz for helping with the dataset split and Jen Dumas for legal advice. This work was supported by the National Science Foundation through a Graduate Research Fellowship (DGE-1256082) and NSF grants (IIS-1524371, 1637479, 165205, 1703166), the DARPA CwC program through ARO (W911NF-15-1-0543), the IARPA DIVA program through D17PC00343, the Sloan Research Foundation through a Sloan Fellowship, the Allen Institute for Artificial Intelligence, the NVIDIA Artificial Intelligence Lab, and gifts by Google and Facebook. The views and conclusions contained herein are those of the authors and should not be interpreted as representing endorsements of IARPA, DOI/IBC, or the U.S. Government. 

}

% \section{Conclusion}
% \input{sections/08_conclusion.tex}

\section*{Appendix}
\appendix
\begin{abstract}
In our work we presented the new task of Visual Commonsense Reasoning and introduced a large-scale dataset for the task, \datasetname, along with \evaluationname, the machinery that made the dataset construction possible. We also presented \modelname, a new model for the task. In the supplemental material, we provide the following items that shed further insight on these contributions:

\begin{itemize}[labelwidth=!,itemsep=0pt,topsep=1pt,parsep=1pt]
    \item Additional dataset analysis (Section~\ref{sec:datasetanalysis})
    \item More information about dataset creation (Section~\ref{sec:datasetcreationdetails}) and \evaluationname~(Section \ref{sec:matchingdetails})
    \item An extended discussion on language priors (Section~\ref{sec:langpriors})
    \item Model hyperparameters used (Section~\ref{sec:modeldetails})
    \item Additional VQA Baseline Results, with BERT embeddings (Section~\ref{sec:vqabert})
    \item A datasheet for \datasetname~(Section~\ref{sec:datasheet})
    \item A visualization of \modelname's predictions (Section \ref{sec:qualresults2})
\end{itemize}
For more examples, and to obtain the dataset and code, check out {\small\tt\href{https://visualcommonsense.com}{visualcommonsense.com}}.
\end{abstract}

\section{Dataset Analysis}\label{sec:datasetanalysis}
In this section, we continue our high-level analysis of \datasetname. 

\subsection{Language complexity and diversity} How challenging is the language in \datasetname? We show several statistics in Table~\ref{tab:datastats}. Of note, unlike many question-answering datasets wherein the answer is a single word, our answers average to more than 7.5 words. The rationales are even longer, averaging at more than 16 words.

\begin{figure}[t!]
    \vspace{-3mm}
    \centering
    \includegraphics[width=\columnwidth]{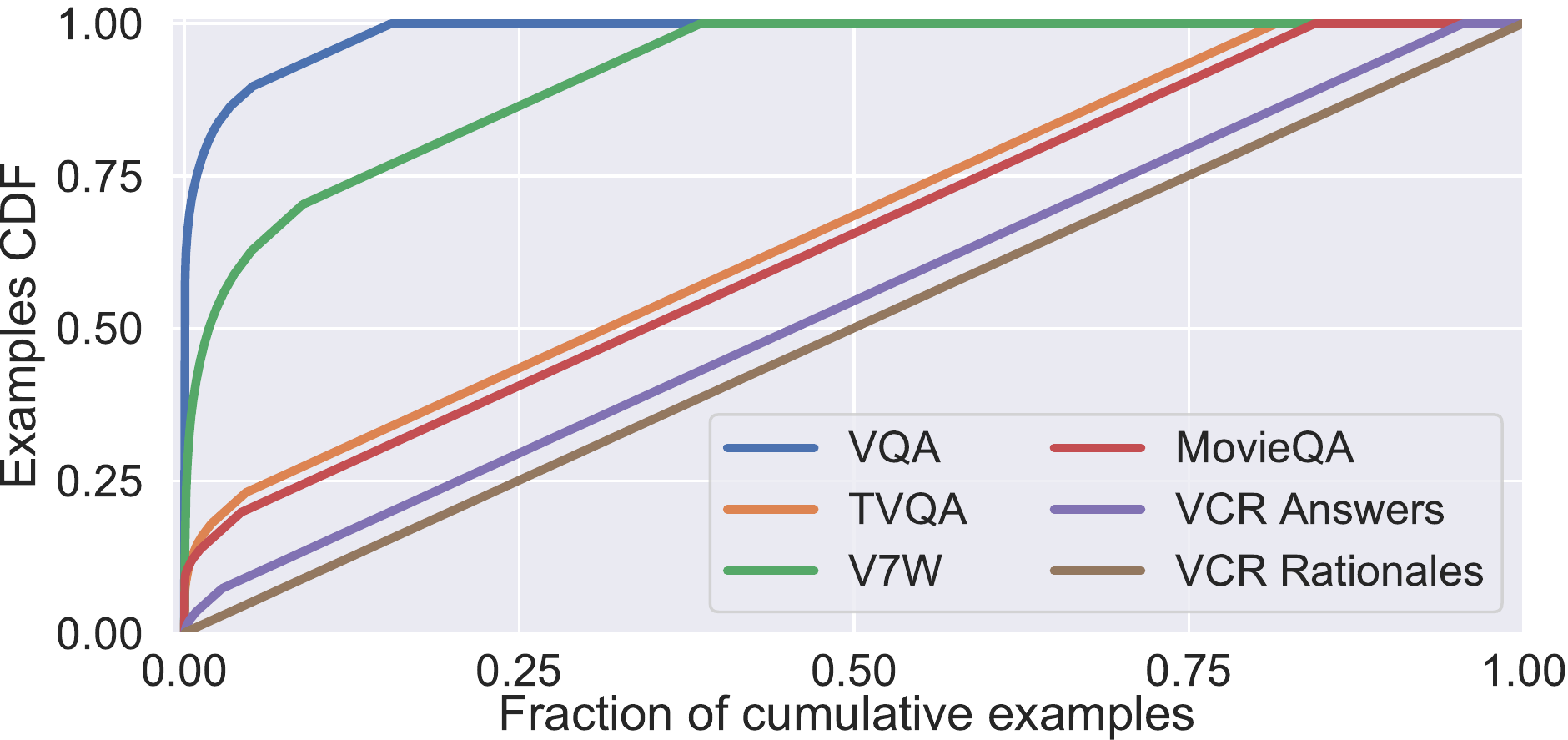}
    \caption{CDF of dataset examples ordered by frequency in question-answering datasets \cite{antol2015vqa, zhu2016cvpr, tapaswi2016movieqa, lei2018tvqa}. To obtain this plot, we sampled 10,000 answers from each dataset (or rationales, for `\datasetname~rationales'). We consider two examples to be the same if they exactly match, after tokenization, lemmatization, and removal of stopwords. Where many datasets in this space are light-tailed, our dataset shows great diversity (e.g. almost every rationale is unique.)}
    \label{fig:cdfofexamples}
\end{figure}

An additional informative statistic is the counts of unique answers and rationales in the dataset, which we plot in Figure~\ref{fig:cdfofexamples}. As shown, almost every answer and rationale is unique. 

\subsection{Objects covered} 
\begin{figure}[h!]
    \vspace{-3mm}
    \centering
    \includegraphics[width=\columnwidth]{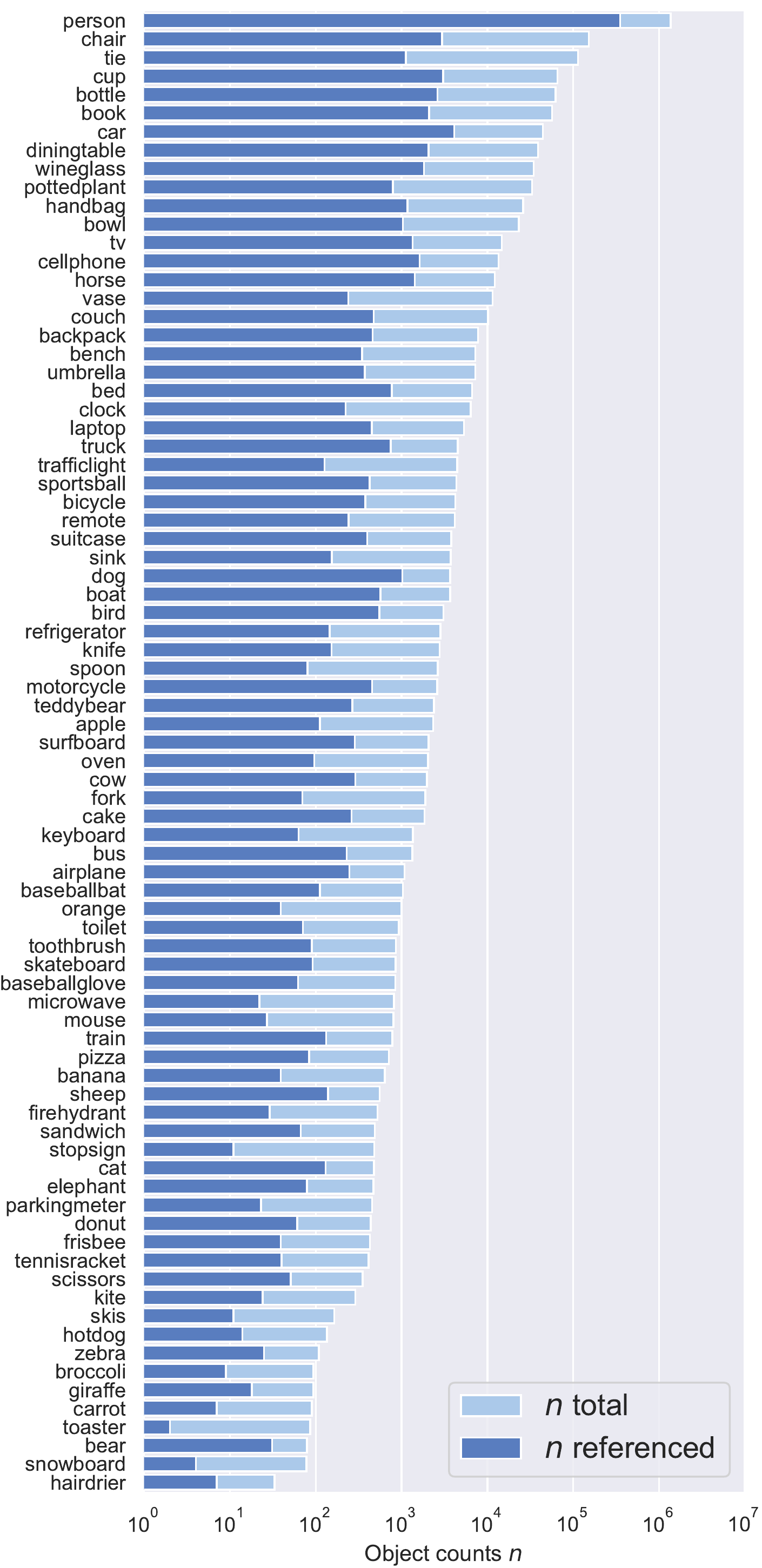}
    \vspace*{-7mm}
    \caption{Distribution of the referenced COCO \cite{lin2014microsoft} objects in \datasetname. We count an object as being `referenced' if, for a given question, answer, and rationale, that object is mentioned explicitly. Note that we do not double-count objects here - if \texttt{person5} is mentioned in the question and the answer, we count it once. This chart suggests that our dataset is mostly human-centric, with some categories being referenced more than others (cars are mentioned more than ties, even though cars appear less often).} \vspace{-3mm}
    \label{fig:objdetected}
\end{figure}

On average, there are roughly two objects mentioned over a question, answer, and rationale. Most of these objects are people (Figure~\ref{fig:objdetected}), though other types of COCO objects are common too \cite{lin2014microsoft}. Objects such as `chair,' `tie,' and `cup' are often detected, however, these objects vary in terms of scene importance: even though more ties exist in the data than cars, workers refer to cars more in their questions, answers, and rationales. Some objects, such as hair driers and snowboards, are rarely detected.

\begin{table}[t!]
    \small\centering
    \begin{tabular}{@{}l@{\hspace{0.4em}} |l@{\hspace{0.4em}} l@{\hspace{0.4em}} l@{\hspace{0.4em}}l@{}}
        \toprule
        & Train & Val & Test \\
        \midrule
        Number of questions & 212,923 & 26,534 & 25,263 \\
        Number of answers per question & 4 & 4 & 4 \\
        Number of rationales per question & 4 & 4 & 4 \\ \midrule
        Number of images & 80,418 & 9,929 & 9,557 \\
        Number of movies covered & 1,945 & 244 & 189 \\ \midrule
        Average question length & 6.61 & 6.63 & 6.58 \\
        Average answer length & 7.54 & 7.65 & 7.55 \\ 
        Average rationale length & 16.16 & 16.19 & 16.07 \\ 
        Average \# of objects mentioned & 1.84 & 1.85 & 1.82 \\ 
        \bottomrule
    \end{tabular}
    \caption{High level dataset statistics, split by fold (train, validation, and test). Note that we held out one fold in the dataset for blind evaluation at a later date; this fold is blind to us to preserve the integrity of the held-out data. Accordingly, the statistics of that fold are not represented here. }
    \label{tab:datastats}
\end{table}

\subsection{Movies covered} 
\begin{figure}[t!]
    \vspace{-3mm}
    \centering
    \includegraphics[width=\columnwidth]{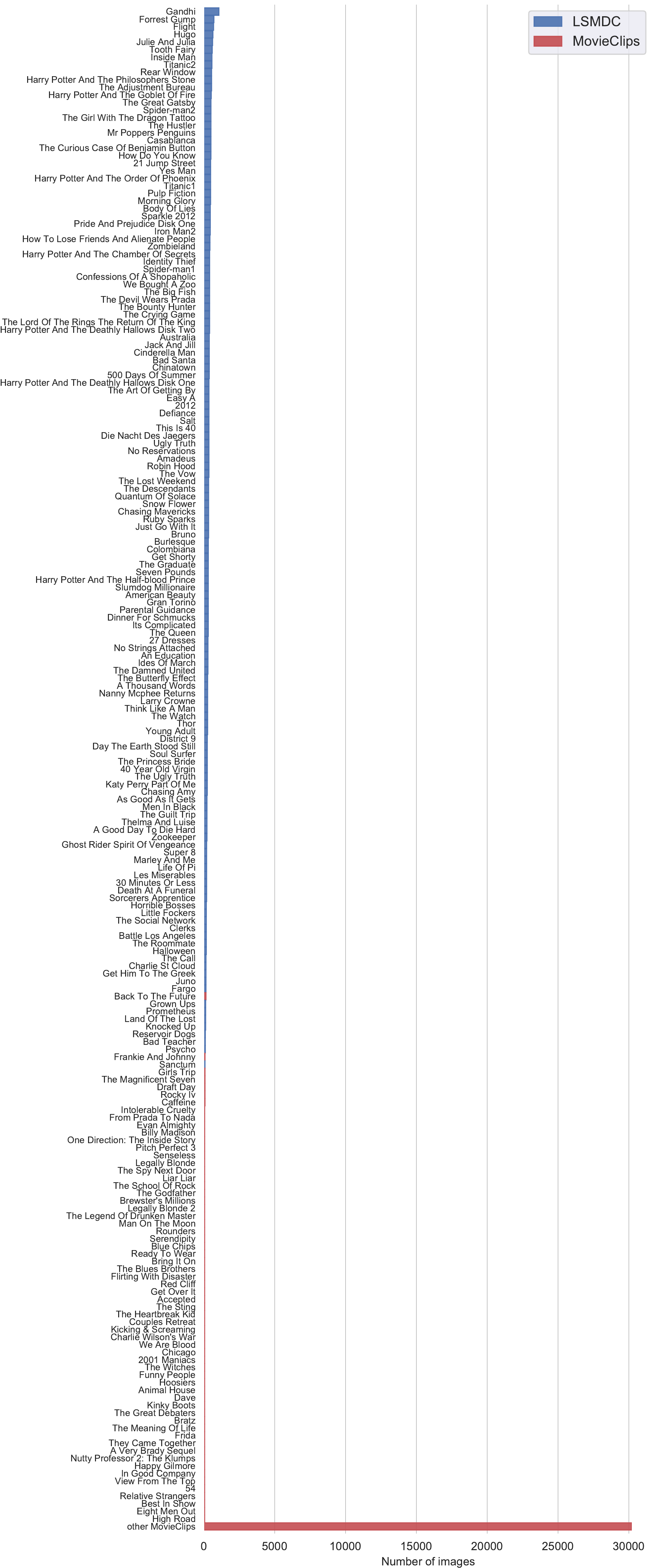}
    \vspace*{-5mm}
    \caption{Distribution of movies in the \datasetname~training set by number of images. Blue bars are movies from LSMDC (46k images); red are MovieClips (33k images). The MovieClips images are spread over a wider range of movies: due to space restrictions, most are under `other MovieClips.'}\vspace*{-3mm}
    \label{fig:moviedistribution}
\end{figure}

\begin{figure}[h!]
    \vspace{-3mm}
    \centering
    \includegraphics[width=\columnwidth]{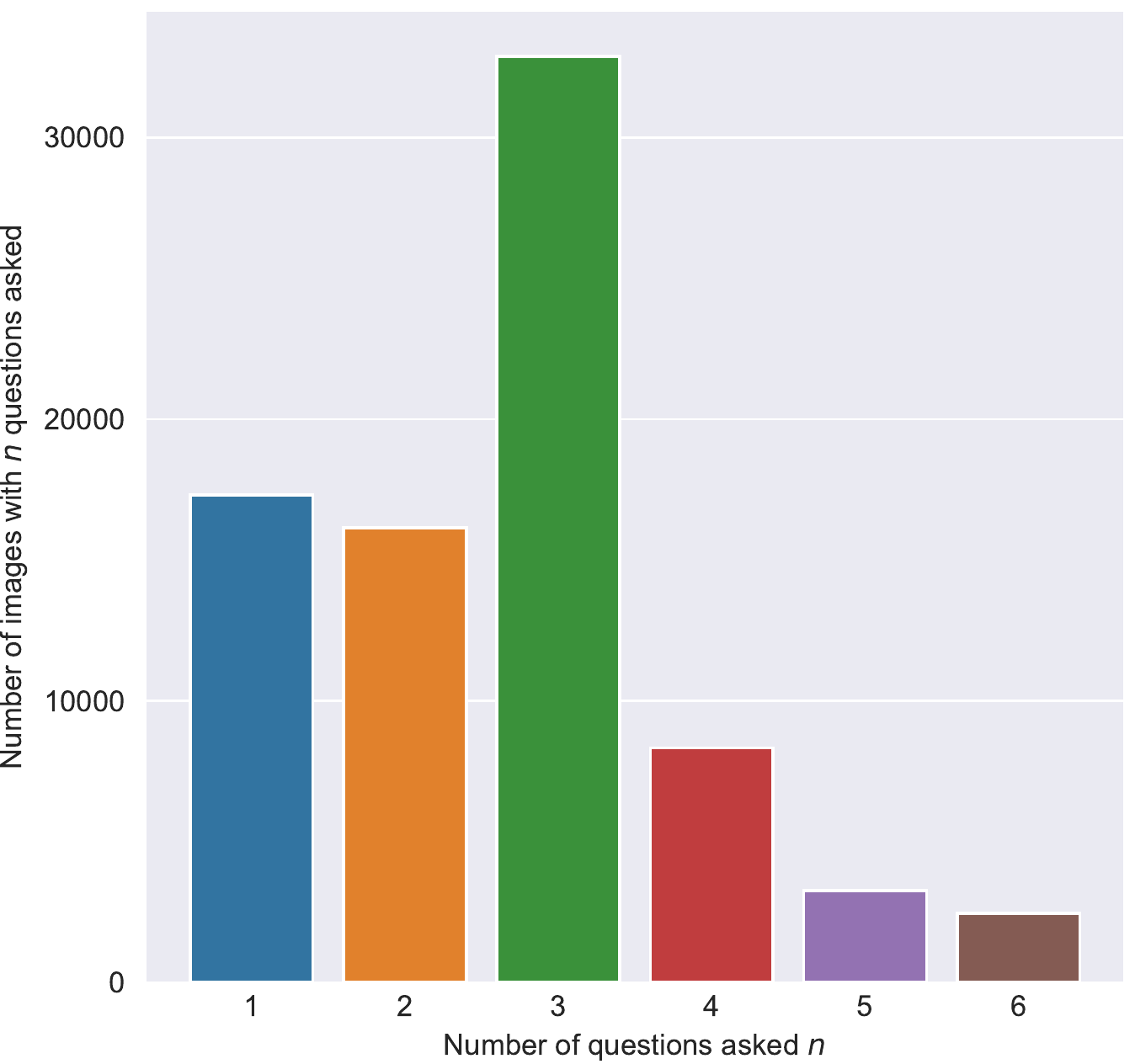}
    \caption{Number of questions asked per image on the \datasetname~training set.
    The average number of questions asked per image is 2.645. Note that while workers could ask anywhere between one to three questions per image, images that were flagged as especially interesting by workers got re-annotated with additional annotations.}
    \label{fig:qpi}
\end{figure}

Our dataset also covers a broad range of movies - over 2000 in all, mostly via MovieClips (Figure~\ref{fig:moviedistribution}). We note that since we split the dataset by movie, the validation and test sets cover a completely disjoint set of movies, which forces a model to generalize. For each movie image, workers ask 2.6 questions on average (Figure~\ref{fig:qpi}), though the exact number varies - by design, workers ask more questions for more interesting images.

\subsection{Inference types} It is challenging to accurately categorize commonsense and cognition-level phenomena in the dataset. One approach that we presented in Figure~\ref{fig:questiondiversity} is to categorize questions by type: to estimate this over the entire training set, we used a several patterns, which we show in Table~\ref{tab:rules}. Still, we note that automatic categorization of the inference types required for this task is hard. This is in part because a single question might require multiple types of reasoning: for example, `Why does \texttt{person1} feel embarrassed?' requires reasoning about \texttt{person1}'s mental state, as well as requiring an explanation. For this reason, we argue that this breakdown underestimates the task difficulty.


\begin{table}[t!]
    \small\centering
    \begin{tabular}{@{}c@{\hspace{0.4em}}c@{\hspace{0.4em}}l@{}}
        \toprule
        Type & Freq. & Patterns \\
        \midrule
        \cmidrule{1-3}
         Explanation & 38\% & why, how come, how does \\
         Activity & 24\% & doing, looking, event, playing, preparing \\
         Temporal & 13\% & happened, before, after, earlier, later, next \\
         Mental & 8\% & feeling, thinking, saying, love, upset, angry \\
         Role & 7\% & relation, occupation, strangers, married \\
         Scene & 5\% & where, time, near \\
         Hypothetical & 5\% & if, would, could, chance, might, may \\
         \bottomrule
    \end{tabular}
    \caption{Some of the rules we used to determine the type of each question. Any question containing a word from one of the above groups (such as `why') was determined to be of that type (`explanation').}
    \label{tab:rules}
\end{table}

%%%%%%%%%%%%%%%%%%%%%%%%%%%%%%%%%%%%%%%%%%%%%%%%%%%%%%%%%%%%%%%%%%%%%%%%
\section{Dataset Creation Details}
\label{sec:datasetcreationdetails}
In this section, we elaborate more on how we collected \datasetname, and about our crowdsourcing process.
\subsection{Shot detection pipeline}
The images in \datasetname~are extracted from video clips from LSMDC \cite{rohrbach_movie_2017} and MovieClips. These clips vary in length from a few seconds (LSMDC) to several minutes (MovieClips). Thus, to obtain more still images from these clips, we performed shot detection. Our pipeline is as follows: 
\begin{itemize}[labelwidth=!,itemsep=0pt,topsep=1pt,parsep=1pt]
    \item We iterate through a video clip at a speed of one frame per second.
    \item During each iteration, we also perform shot detection: if we detect a mean difference of 30 pixels in HSV space, then we register a shot boundary.
    \item After a shot boundary is found, we apply Mask-RCNN \cite{He2017MaskR,Detectron2018} on the middle frame for the shot, and save the resulting image and detection information.
    \end{itemize}
We used a threshold of 0.7 for Mask-RCNN, and the best detection/segmentation model available for us at the time: X-101-64x4d-FPN\footnote{Available via \href{https://github.com/facebookresearch/Detectron/blob/master/MODEL_ZOO.md}{the Detectron Model Zoo}.}, which obtains 42.4 box mAP on COCO, and 37.5 mask mAP.

\subsection{Interestingness Filter}
\begin{figure}[t!]
    \vspace{-3mm}
    \centering
    \includegraphics[width=\columnwidth]{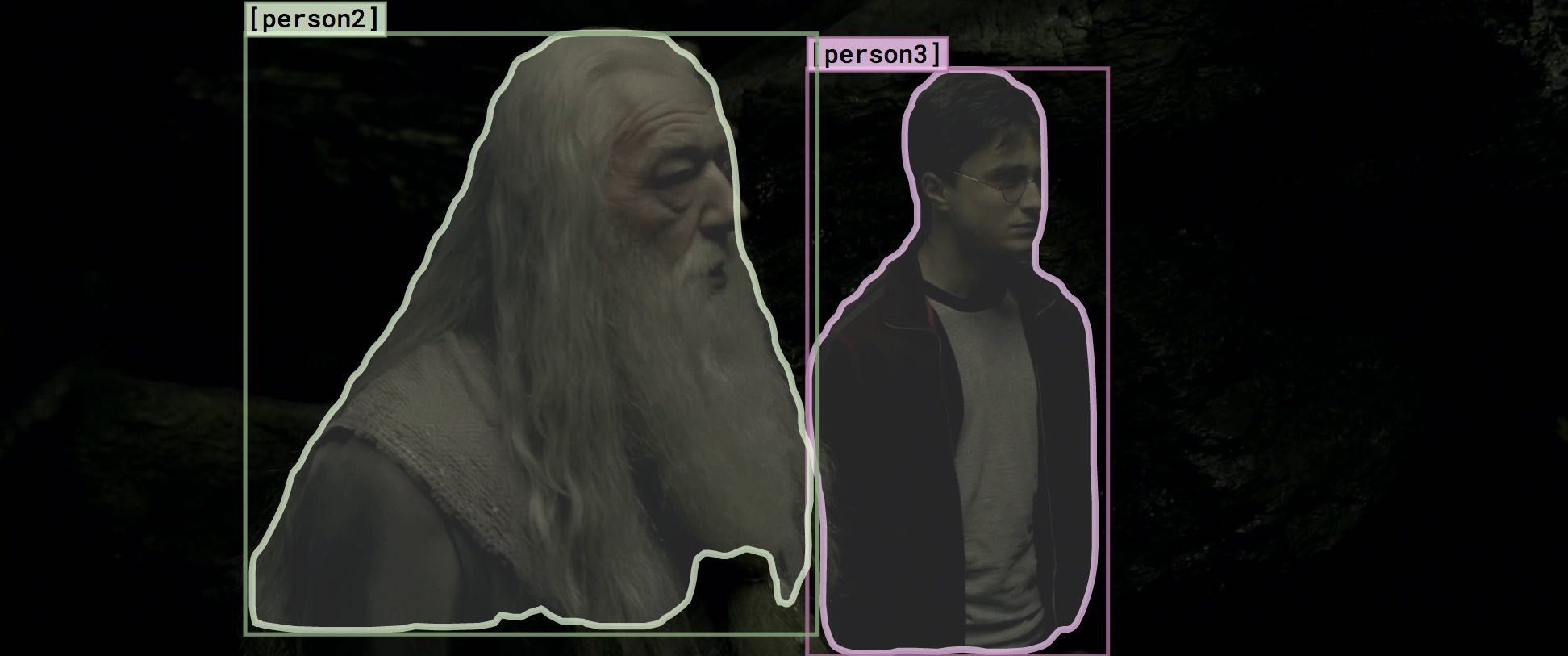}
    \textbf{a)} Boring image.
    \includegraphics[width=\columnwidth]{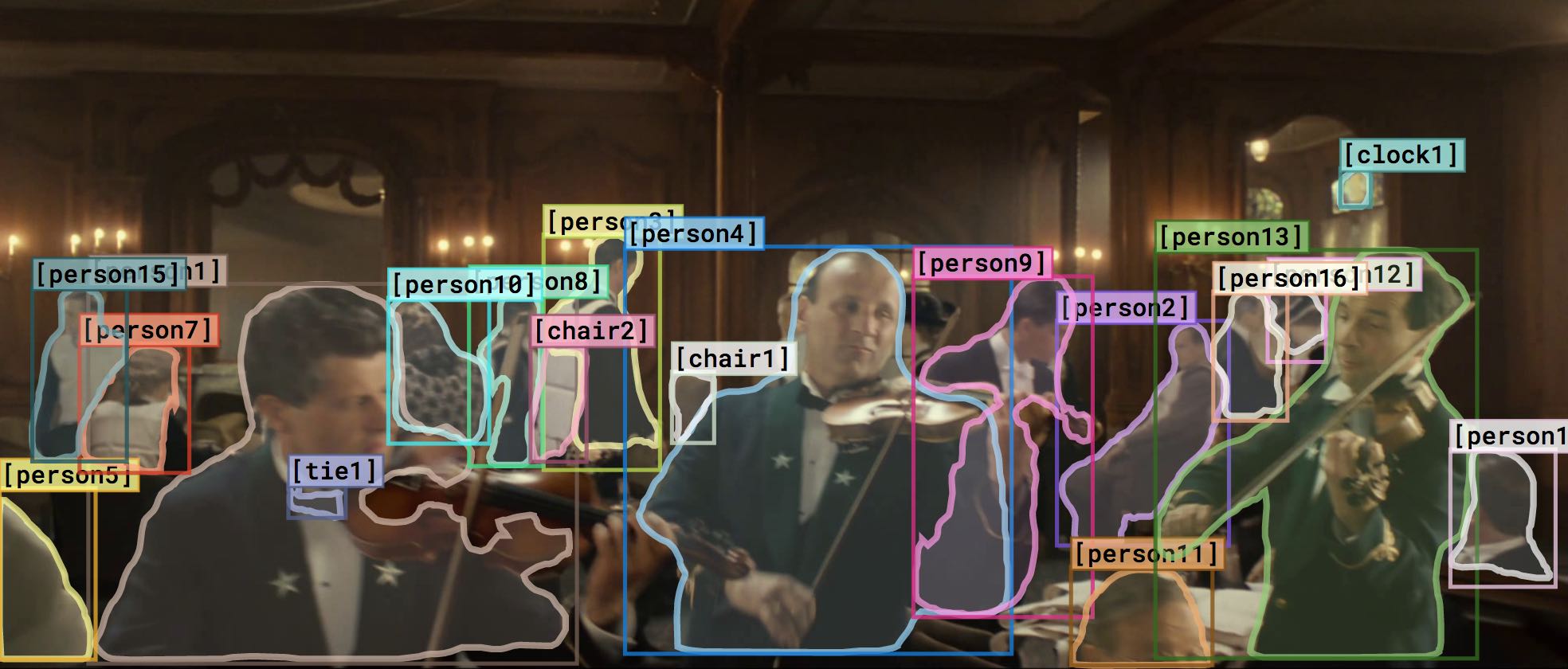}
    \textbf{b)} Interesting image.
    \caption{Two example images that come from the raw video pipeline. Image a) is flagged by our initial filter as `boring', because there are only two people without any additional objects, whereas image b) is flagged as being interesting due to the number of people and objects detected.}
    \label{fig:boring}
\end{figure}

Recall that we use an `interestingness filter' to ensure that the images in our dataset are high quality. First, every image had to have at least two people in it, as detected by Mask RCNN. However, we also found that many images with two or more people were still not very interesting. The two main failure cases here are when there are one or two people detected, but they aren't doing anything interesting (Figure~\ref{fig:boring}a), or when the image is especially grainy and blurry. Thus, we opted to learn an additional classifier for determining which images were interesting.

Our filtering process evolved as we collected data for the task. The first author of this paper first manually annotated 2000 images from LSMDC \cite{rohrbach_movie_2017} as being `interesting' or `not interesting' and trained a logistic regression model to predict said label. The model is given as input the number of people detected by Mask RCNN \cite{He2017MaskR,Detectron2018}, along with the number of objects (that are not people) detected. We used this model to identify interesting images in LSMDC, using a threshold that corresponded to 70\% precision. This resulted in 72k images selected; these images were annotated first.

During the crowdsourcing process, we obtained data that allowed us to build an even better interestingness filter later on. Workers were asked, along with each image, whether they thought that the image was especially interesting (and thus should go to more workers), just okay, or especially boring (and hard to ask even one good question for). We used this to train a deeper model for this task. The model uses a ResNet 50 backbone over the entire image \cite{he2016deep} as well as a multilayer perceptron over the object counts. The entire model is trained end-to-end: 2048 dimensional features from Resnet are concatenated with a 512 dimensional projetion of the object counts, and used to predict the labels.\footnote{In addition to predicting interestingness, the model also predicts the number of questions a worker asks, but we never ended up using these predictions.} We used this model to select the most interesting 40k images from Movieclips, which finished off the annotation process.

\subsection{Crowdsourcing quality data}
\begin{figure}[t!]
    \vspace{-3mm}
    \centering
    \includegraphics[width=\columnwidth]{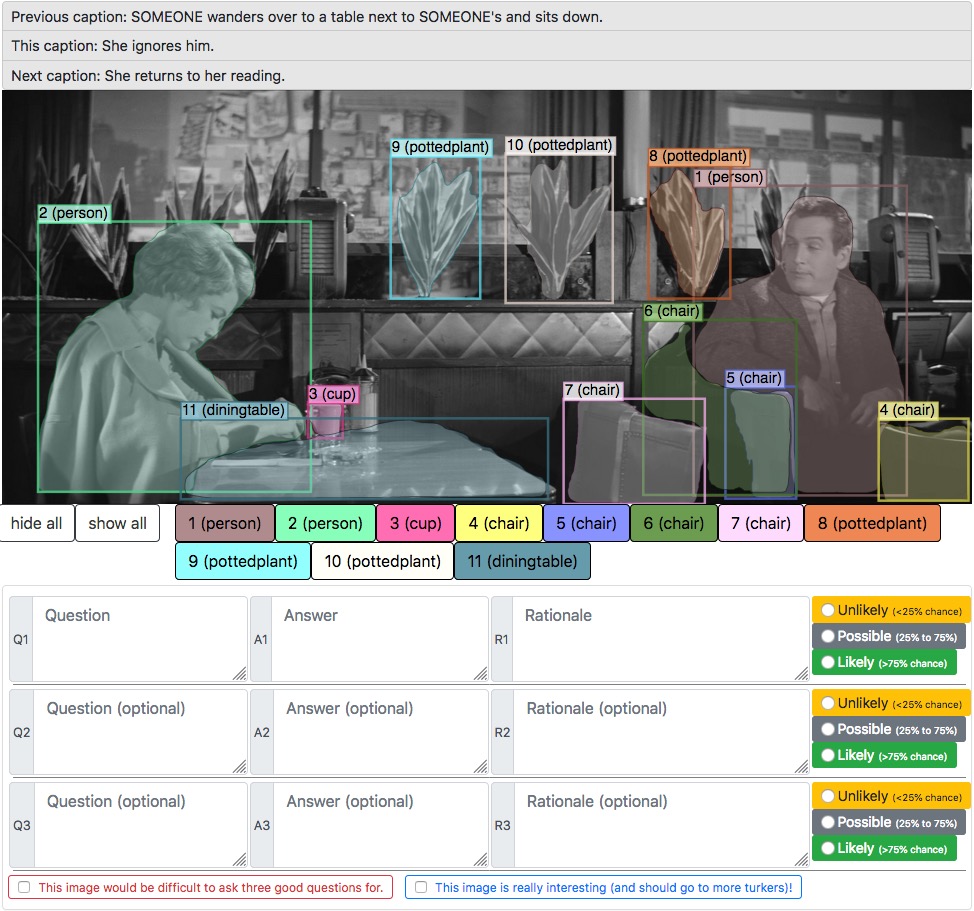}
    \caption{Screenshot of our annotation interface. Workers are given an image, as well as context from the video (here, captions from LSMDC \cite{rohrbach_movie_2017}), and are asked to write one to three questions, answers, and rationales. For each answer, they must mark it as likely, possible, or unlikely. Workers also select whether the image was especially interesting or boring, as this allows us to train a deep model for predicting image interestingness. }
    \label{fig:annotationui}
\end{figure}

As mentioned in the paper, crowdsourcing data at the quality and scale of \datasetname~is challenging. We used several best practices for crowdsourcing, which we elaborate on in this section.

We used Amazon Mechanical Turk for our crowdsourcing. A screenshot of our interface is given in Figure~\ref{fig:annotationui}. Given an image, workers asked questions, answered them, and provided a rationale explaining why their answer might be correct. These are all written in a mixture of natural language text, as well as referring to detection regions. In our annotation UI, workers refer to the regions by writing the tag number.\footnote{Note that this differs a bit from the format in the paper: we originally had workers write out the full tag, like \texttt{[person5]}, but this is often long and the workers would sometimes forget the brackets. Thus, the tag format here is just a single number, like \texttt{5}.}

Workers could ask anywhere between one to three questions per HIT. We paid the workers proportionally at \$0.22 per triplet. According to workers, this resulted in \$8--25/hr. This proved necessary as workers reported feeling ``drained'' by the high quality required.

\paragraph{Automated quality checks}
We added several automated checks to the crowdsourcing UI to ensure high quality. The workers had to write at least four words for the question, three for the answer, and five for the rationale. Additionally, the workers had to explicitly refer to at least one detection on average per question, answer, and rationale triplet. This was automatically detected to ensure that the workers were referring to the detection tags in their submissions.

We also noticed early on was that sometimes workers would write detailed stories that were only loosely connected with the semantic content of the image. To fix this, workers also had to self-report whether their answer was likely (above 75\% probability), possible (25-75\% probability), or unlikely (below 25\% probability). We found that this helped deter workers from coming up with consistently unlikely answers for each image. The likelihood ratings were never used for the task, since we found they weren't necessary to obtain high human agreement.

\paragraph{Instructions}
Like for any crowdsourcing task, we found wording the instructions carefully to be crucial. We encouraged workers to ask about  higher-level actions, versus lower-level ones (such as `What is \texttt{person1} wearing?'), as well as to not ask questions and answers that were overly generic (and thus could apply to many images). Workers were encouraged to answer reasonably in a way that was not overly unlikely or unreasonable. To this end, we provided the workers with high-quality example questions, answers, and rationales.

\paragraph{Qualification exam}
Since we were picky about the types of questions asked, and the format of the answers and rationales, workers had to pass a qualification task to double check that they understood the format. The qualification test included a mix of multiple-choice graded answers as well as a short written section, which was to provide a single question, answer, and rationale for an image. The written answer was checked manually by the first author of this paper.

\paragraph{Work verification}
In addition to the initial qualification exam, we also periodically monitored the annotation quality. Every 48 hours, the first author of this paper would review work and provide aggregate feedback to ensure that workers were asking good questions, answering them well, and structuring the rationales in the right way. Because this took significant time, we then selected several outstanding workers and paid them to do this job for us: through a separate set of HITs, these outstanding workers were paid \$0.40 to provide detailed feedback on a submission that another worker made. Roughly one in fifty HITs were annotated in this way to give extra feedback. Throughout this process, workers whose submission quality dropped were dequalified from the HITs.

%%%%%%%%%%%%%%%%%%%%%%%%%%%%%%%%%%%%%%%%%%%%%%%%%%%
\section{Adversarial Matching Details}\label{sec:matchingdetails}
There are a few more details that we found useful when performing the \evaluationname~to create \datasetname, which we discuss in this section.

\paragraph{Aligning Detections} In practice, most responses in our dataset are not relevant to most queries, due to the diversity of responses in our dataset and the range of detection tags (\texttt{person1}, etc.). 

To fix this, for each query $\boldsymbol{q}_i$ (with associated object list $\boldsymbol{o}_i$ and response $\boldsymbol{r}_i$) we turn each candidate 
$\boldsymbol{r}_j$ into a template, and use a rule based system to probabilistically remap its detection tags to match the objects in $\boldsymbol{o}_i$. With some probability, a tag in $\boldsymbol{r}_j$ is replaced with a tag in $\boldsymbol{q}_i$ and $\boldsymbol{r}_i$. Otherwise, it is replaced with a random tag from $\boldsymbol{o}_i$.

We note that our approach isn't perfect. The remapping system often produces responses that violate predicate/argument structure, such as `\texttt{person1} is kissing \texttt{person1}.' However, \emph{our approach does not need to be perfect}: because the detections for response $\boldsymbol{r}_j$ are remapped uniquely for each query $\boldsymbol{q}_i$, with some probability, there should be at least some remappings of $\boldsymbol{r}_i$ that make sense, and the question relevance model $P_{rel}$ should select them.

\paragraph{Semantic categories} Recall that we use 11 folds for the dataset of around 290k questions, answers, and rationales. Since we must perform \evaluationname~once for the answers, as well as for the rationales, this would naively involve 22 matchings on a fold size of roughly 26k. We found that the major computational bottleneck wasn't the bipartite matching\footnote{We use the \href{LAP}{https://github.com/gatagat/lap} implementation.}, but rather the computation of all-pairs similarity and relevance between $\sim$26k examples.

There is one additional potential problem: we want the dataset examples to require a lot of complex commonsense reasoning, rather than simple attribute identification. However, if the response and the query disagree in terms of gender pronouns, then many of the dataset examples can be reduced to gender identification.

We address both of these problems by dividing each fold into `buckets' of 3k examples for matching. We divide the examples up in terms of the pronouns in the response: if the response contains a female or male pronoun, then we put the example into a `female' or `male' bucket, respectively, otherwise the response goes into the `neutral' bucket. To further divide the dataset examples, we also put different question types in different buckets for the question answering task (e.g. who, what, etc.). For the answer justification task, we cluster the questions and answers using their average GloVe embeddings \cite{pennington2014glove}. 

\paragraph{Relevance model details} Recall that our relevance model $P_{rel}$ is trained to predict the probability that a response $\boldsymbol{r}$ is valid for a query $\boldsymbol{q}$. We used BERT for this task \cite{devlin2018bert}, as it achieves state-of-the-art results across many two-sentence inference tasks. Each input looks like the following, where the query and response are concatenated with a separator in between:

{\small \texttt{[CLS] what is casey doing ? [SEP] casey is getting out of car . [SEP]} }

Note that in the above example, object tags are replaced with the class name (\texttt{car3}$\to$\texttt{car}). Person tags are replaced with gender neutral names (\texttt{person1}$\to$\texttt{casey}) \cite{andrew_flowers_most_2015}.

We fine-tune BERT by treating it as a two-way classification problem. With probability 25\% for a query, BERT is given that query's actual response, otherwise it is given a random response (where the detections were remapped). Then, the model must predict whether it was given the actual response or not. We used a learning rate of $2\cdot 10^{-5}$, the Adam optimizer \cite{Kingma2014AdamAM}, a batch size of 32, and 3 epochs of fine-tuning.\footnote{We note that during the \evaluationname~process, for either Question Answering or Answer Justification, the dataset is broken up into 11 folds. For each fold, BERT is fine-tuned on the other folds, not on the final dataset splits.}

Due to computational limitations, we used BERT-Base as the architecture rather than BERT-Large - the latter is significantly slower.\footnote{Also, BERT-Large requires much more memory, enough so that it's harder to fine-tune due to the smaller feasible batch size.} Already, $P_{rel}$ has an immense computational requirement as it must compute all-pairs similarity for the entire dataset, over buckets of 3000 examples. Thus, we opted to use a larger bucket size rather than a more expensive model.

\paragraph{Similarity model details} While we want the responses to be highly relevant to the query, we also want to avoid cases where two responses might be conflated by humans - particularly when one is the correct response. %We especially want to avoid having the correct response be conflated with an incorrect choice.
This conflation might occur for several reasons: possibly, two responses are \emph{paraphrases} of one another, or one response \emph{entails} another. We lump both under the `similarity' umbrella as mentioned in the paper and introduce a model, $P_{sim}$, to predict the probability of this occurring - broadly speaking, that two responses $\boldsymbol{r}_i$ and $\boldsymbol{r}_j$ have the same meaning.

We used ESIM+ELMo for this task \cite{chen2017enhanced, peters2018deep}, as it still does quite well on two-sentence natural language inference tasks (although not as well as BERT), and can be made much more efficient. At test time, the model makes the similarity prediction when given two token sequences.\footnote{Again, with object tags replaced with the class name, and person tags replaced by gender neutral names.}

We trained this model on freely available NLP corpora. We used the SNLI formalism \cite{bowman2015snli}, in which two sentences are an `entailment' if the first entails the second, `contradiction' if the first is contradicted by the second, and `neutral' otherwise. We combined data from SNLI and MultiNLI \cite{williams17multisnli} as training data. Additionally, we found that even after training on these corpora, the model would struggle with paraphrases, so we also translated SNLI sentences from English to German and back using the Nematus machine translation system \cite{wieting2017learning, sennrich-EtAl:2017:EACLDemo}. These sentences served as extra paraphrase data and were assigned the `entailment' label. We also used randomly sampled sentence pairs from SNLI as additional `neutral' training data. We held out the SNLI validation set to determine when to stop training. We used standard hyperparameters for ESIM+ELMo as given by the AllenNLP library \cite{Gardner2017AllenNLP}.

Given the trained model $P_{nli}$, we defined the similarity model as the maximum entailment probability for either way of ordering the two responses:
\begin{equation}
P_{sim}(\boldsymbol{r}_i, \boldsymbol{r}_j)={\max}\Big\{ P_{nli}(\textrm{ent} | \boldsymbol{r}_i, \boldsymbol{r}_j), P_{nli}(\textrm{ent} | \boldsymbol{r}_j, \boldsymbol{r}_i) \Big\},
\end{equation}
where `$\textrm{ent}$' refers to the `entailment' label. If one response entails the other, we flag them as similar, even if the reverse entailment is not true, because such a response is likely to be a false positive as a distractor.

The benefit of using ESIM+ELMo for this task is that it can be made more efficient for the task of all-pairs sentence similarity. While much of the ESIM architecture involves computing attention between the two text sequences, everything before the first attention can be precomputed. This provides a large speedup, particularly as computing the ELMo representations is expensive. Now, for a fold size of $N$, we only have to compute $2N$ ELMo representations rather than $N^2$.

\begin{figure}[t!]
    \vspace{-3mm}
    \centering
    \includegraphics[width=\columnwidth]{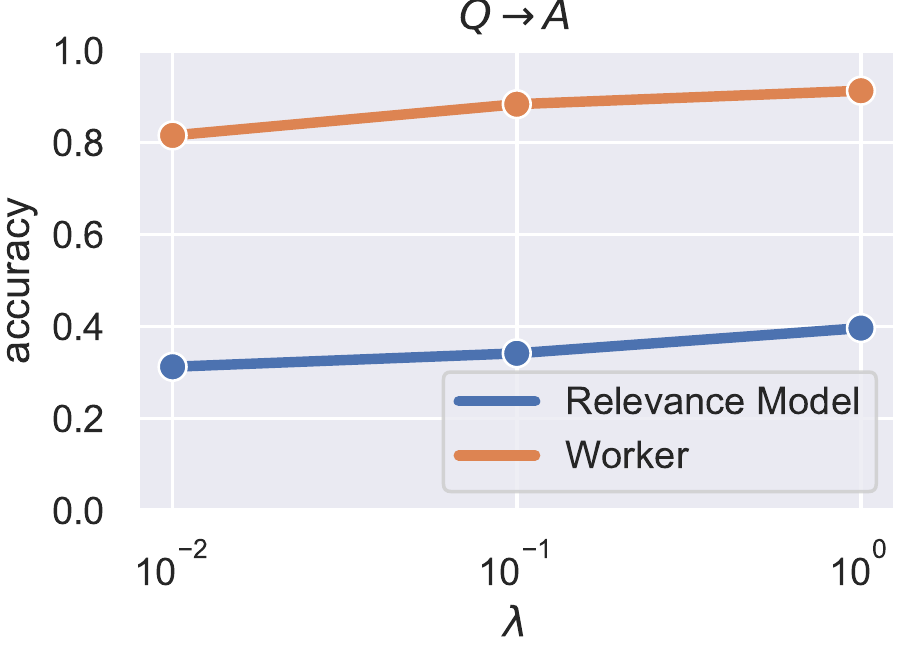}
    % \textbf{a)} Answer validation of $\lambda$
    \vspace*{3mm}
    \includegraphics[width=\columnwidth]{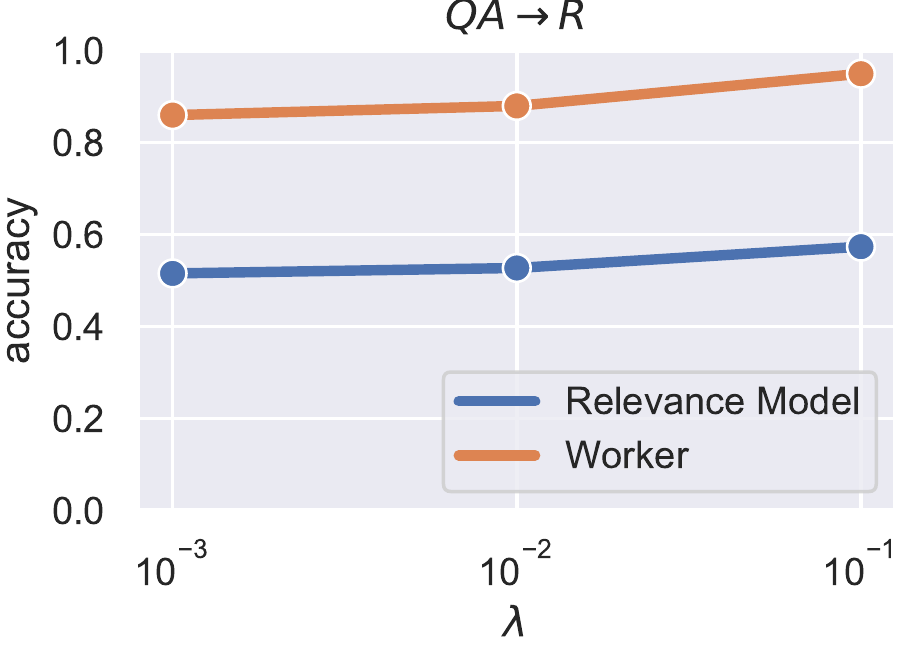}
    % \textbf{b)} Interesting image.
    \caption{Tuning the $\lambda$ hyperparameter. Workers were asked to solve 100 dataset examples from the validation set, as given by \evaluationname~for each considered value of $\lambda$. We used these results to pick reasonable values for the hyperparameter such that the task was difficult for the question relevance model $P_{rel}$, while simple for human workers. We chose $\lambda=0.1$ for $Q \rightarrow A$ and $\lambda=0.01$ for $QA \rightarrow R$.}
    \label{fig:lambdaval}
\end{figure}

\paragraph{Validating the $\lambda$ parameter}
Recall that our hyperparameter $\lambda$ trades off between machine and human difficulty for our final dataset. We shed more insight on how we chose the exact value for $\lambda$ in Figure~\ref{fig:lambdaval}. We tried several different values of $\lambda$ and chose $\lambda=0.1$ for $Q \rightarrow A$ and $\lambda=0.01$ for $QA \rightarrow R$, as at these thresholds human performance was roughly $90\%$. For an easier dataset for both humans and machines, we would increase the hyperparameter.

%%%%%%%%%%%%%%%%%%%%%%%%%%%%%%%%%%%%%

\section{Language Priors and Annotation Artifacts Discussion}\label{sec:langpriors}
There has been much research in the last few years in understanding what `priors' datasets have.\footnote{This line of work is complementary to other notions of dataset bias, like understanding what phenomena datasets cover or don't \cite{torralba2011unbiased}, particularly how that relates to how marginalized groups are represented and portrayed \cite{ schofield2016gender,zhao2017men,sap2017connotation,rudinger2017social}.} Broadly speaking, how well do models do on \datasetname, as well as other visual question answering tasks, without vision?

To be more general, we will consider problems where a model is given a \emph{question} and \emph{answer choices}, and picks exactly one answer. The \emph{answer choices} are the outputs that the model is deciding between (like the responses in \datasetname) and the \emph{question} is the shared input that is common to all \emph{answer choices} (the query, image, and detected objects in \datasetname). With this terminology, we can categorize unwanted dataset priors in the following ways:
\begin{itemize}[labelwidth=!,itemsep=0pt,topsep=0pt,parsep=1pt]
    \item {\bf Answer Priors}: A model can select a correct answer without even looking at the question. Many text-only datasets contain these priors. For instance, the RocStories dataset \cite{mostafazadeh_corpus_2016} (in which a model must classify endings to a story as correct or incorrect), a model can obtain 75\% accuracy by looking at stylistic features (such as word choice and punctuation) in the endings. 
    \item {\bf Non-Visual Priors}: A model can select a correct answer using only non-visual elements of the question. One example is VQA 1.0 \cite{antol2015vqa}: given a question like `What color is the fire hydrant?' a model will classify some answers higher than others (red). This was addressed in VQA 2.0 \cite{balanced_vqa_v2}, however, some answers will still be more likely than others (VQA's answers are open-ended, and an answer to `What color is the fire hydrant?' must be a color).
    % \item {\bf Additional Language Information}: For some visual tasks, a model is given additional language information, which generally makes the task much easier (and reducing the need for the model to look at the image).
\end{itemize}

These priors can either arise from biases in the world (fire hydrants are usually red), or, they can come from annotation artifacts \cite{gururangan2018annotation}: patterns that arise when people write class-conditioned answers. Sometimes these biases are subliminal: when asked to write a correct or incorrect story ending, the correct endings tend to be longer \cite{Schwartz:2017}. Other cases are more obvious: workers often use patterns such as negation to write sentences that contradict a sentence \cite{gururangan2018annotation}.\footnote{For instance, the SNLI dataset contains pairs of sentences with labels such as `entailed' or `contradiction' \cite{bowman2015snli}. For a sentence like `A skateboarder is doing tricks' workers often write `Nobody is doing tricks' which is a contradiction. The result is that the word `nobody' is highly predictive of a word being a contradiction.} 

To what extent do vision datasets suffer from annotation artifacts, versus world priors? We narrow our focus to multiple-choice question answering datasets, in which for humans traditionally write correct \emph{and} incorrect answers to a question (thus, potentially introducing the annotation artifacts). In Table~\ref{tab:priors} we consider several of these datasets: TVQA \cite{lei2018tvqa}, containing video clips from TV shows, along with subtitles; MovieQA \cite{tapaswi2016movieqa}, with videos from movies and questions obtained from higher-level plot summaries; PororoQA \cite{kim2016pororoqa}, with cartoon videos; and TGIFQA \cite{jang2017tgif}, with templated questions from the TGIF dataset \cite{li2016tgif}. We note that these all differ from our proposed~\datasetname~in terms of subject matter, questions asked, number of answers (each of the above has 5 answers possible, while we have 4) and format; our focus here is to investigate how difficult these datasets are for text-only models.\footnote{It should be noted that all of these datasets were released before the existence of strong text-only baselines such as BERT.} Our point of comparison is \datasetname, since our use of \evaluationname~means that humans never write incorrect answers.

\begin{table}[t!]
\vspace{-3mm}
\centering
% \begin{normal}
\setlength{\tabcolsep}{4pt}
\begin{tabular}{@{} l @{\hspace{0.0em}} r @{\hspace{0.1em}} | l@{\hspace{0.5em}} l@{\hspace{0.5em}} l@{\hspace{0.5em}}  c @{}}
Dataset & $\#_{train}$ & Chance  &A& Q+A & S+Q+A\\ \toprule
TVQA \cite{lei2018tvqa} & 122,039 &20.0 & 45.0 & 47.4 & 70.6$^\spadesuit$ \\ %\midrule
MovieQA \cite{tapaswi2016movieqa}& 9,848 & 20.0 &33.8 & 35.4 &  36.3$^\clubsuit$ \\ %\midrule
PororoQA \cite{kim2016pororoqa}$\heartsuit$ & 7,530 &20.0 & 43.1 & 47.4 & \\ %\midrule
TGIFQA \cite{jang2017tgif}$\diamondsuit$ & 73,179 & 20.0 &45.8 & 72.5 & \\ \midrule
\datasetname~$Q{\rightarrow}A$ & \multirow{2}{*}{212,923}& 25.0 &27.6 & 53.8 & \\ 
\datasetname~$QA{\rightarrow}R$ &  & 25.0 & 26.3 & 64.1  & \\ \midrule
\datasetname$^{\textrm{small}}$ $Q{\rightarrow}A$ & \multirow{2}{*}{9,848} & 25.0 & 25.5 & 39.9 & \\ 
\datasetname$^{\textrm{small}}$ $QA \rightarrow R$ & & 25.0 & 25.3 & 50.9 & \\ 
\bottomrule
\end{tabular}
% \end{small}
\vspace*{-2mm}
\caption[]{Text-only results on the validation sets of vision datasets, using BERT-Base. $\#_{train}$ shows the number of training examples. A corresponds to only seeing the answer; in Q+A the model also sees the question; in S+Q+A the model also sees subtitles from the video clip. These results suggest that many multiple choice QA datasets suffer from annotation artifacts, while \evaluationname~helps produce a dataset with minimial biases; moreover, providing extra text-only information (like subtitles) greatly boosts performance. More info:}
{\small \begin{itemize}
    \item[$\spadesuit$:] State of the art.
    \item[$\clubsuit$:] Only 45\% (879/1958) of the questions in the MovieQA validation set have timestamps, which are needed to extract clip-level subtitles, so for the other 55\%, we don't use any subtitle information.
    \item[$\heartsuit$:] No official train/val/test split is available, so we split the data by movie, using 20\% of data for validation and the rest for training.
    \item[$\diamondsuit$:] There seem to be issues with the publicly released train-test split of TGIFQA (namely, a model with high accuracy on a held-out part of the training set doesn't generalize to the provided test set) so we re-split the multiple-choice data ourselves by GIF and hold out 20\% for validation.
\end{itemize}}
\vspace*{-2mm}
\label{tab:priors}
\end{table}

We tackle this problem by running BERT-Base on these models \cite{devlin2018bert}: given only the answer (A), the answer and the question (Q+A), or additional language context in the form of subtitles (S+Q+A), how well does BERT do? Our results in Table~\ref{tab:priors} help support our hypothesis regarding annotation artifacts: whereas accuracy on \datasetname, only given the ending, is 27\% for $Q\rightarrow A$ and 26\% for $Q\rightarrow A$, versus a 25\% random baseline. Other models, where humans write the incorrect answers, have answer-only accuracies from 33.8\% (MovieQA) to 45.8\% (TGIFQA), over a 20\% baseline.

There is also some non-visual bias for all datasets considered: from 35.4\% when given the question and the answers (MovieQA) to 72.5\% (TGIFQA). While these results suggest that MovieQA is incredibly difficult without seeing the video clip, there are two things to consider here. First, MovieQA is roughly 20x smaller than our dataset, with 9.8k examples in training. Thus, we also tried training BERT on `\datasetname$^{\textrm{small}}$': taking 9.8k examples at random from our training set. Performance is roughly 14\% worse, to the point of being roughly comparable to MovieQA.\footnote{Assuming an equal chance of choosing each incorrect ending, the results for BERT on an imaginary 4-answer version of TVQA and MovieQA would be 54.5\% and 42.2\%, respectively.} Second, often times the examples in MovieQA have similar structure, which might help to alleviate stylistic priors, for example:

``Who has followed Boyle to Eamon's apartment?'' Answers:
\begin{enumerate}[labelwidth=!,itemsep=0pt,topsep=0pt,parsep=1pt]
\item Thommo and his IRA squad.
\item Darren and his IRE squad.
\item Gary and his allies.
\item \textbf{Quinn and his IRA squad.}
\item Jimmy and his friends.
\end{enumerate}

On the other hand, our dataset examples tend to be highly diverse in terms of syntax as well as high-level meaning, due to the similarity penalty. We hypothesize that this is why some language priors creep into \datasetname, particularly in the $QA\rightarrow R$ setting: given four very distinct rationales that ostensibly justify why an answer is true, some will likely serve as better justifications than others.

Furthermore, providing additional language information (such as subtitles) to a model tends to boost performance considerably. When given access to subtitles in TVQA,\footnote{We prepend the subtitles that are aligned to the video clip to the beginning of the question, with a special token (\texttt{;}) in between. We trim tokens from the subtitles when the total sequence length is above 128 tokens.} BERT scores 70.6\%, which to the best of our knowledge is a new state-of-the-art on TVQA. 

In conclusion, dataset creation is highly difficult, particularly as there are many ways that unwanted bias can creep in during the dataset creation process. One such bias of this form includes annotation artifacts, which our analysis suggests is prevalent amongst multiple-choice VQA tasks wherein humans write the wrong endings. Our analysis also suggests \evaluationname~can help minimize this effect, even when there are strong natural biases in the underlying textual data.

\section{Model details}\label{sec:modeldetails}

In this section, we discuss implementation details for our model, \modelname.

\paragraph{BERT representations}
As mentioned in the paper, we used BERT to represent text \cite{devlin2018bert}. We wanted to provide a fair comparison between our model and BERT, so we used BERT-Base for each. We tried to make our use of BERT to be as simple as possible, matching our use of it as a baseline. Given a query $\boldsymbol{q}$ and response choice $\boldsymbol{r}^{(i)}$, we merge both into a single sequence to give to BERT. One example might look like the following:

{\small \texttt{[CLS] why is riley riding motorcycle while wearing a hospital gown ? [SEP] she had to leave the hospital in a hurry . [SEP]}}

Note that in the above example, we replaced person tags with gender neutral names \cite{andrew_flowers_most_2015} (\texttt{person3}$\rightarrow$ \texttt{riley}) and replaced object detections by their class name (\texttt{motorcycle1}$\rightarrow$ \texttt{motorcycle}), to minimize domain shift between BERT's pretrained data (Wikipedia and the BookCorpus \cite{moviebook}) and \datasetname.

Each token in the sequence corresponds to a different transformer unit in BERT. We can then use the later layers in BERT to extract contextualized representations for the each token in the query (everything from \texttt{why} to \texttt{?}) and the response (\texttt{she} to \texttt{.}).\footnote{The only slight difference is that, due to the WordPiece encoding scheme, rare words (like \texttt{chortled}) are broken up into subword units (\texttt{cho \#\#rt \#\#led}). In this case, we represent that word as the average of the BERT activations of its subwords.}
 Note that this gives us a different representation for each response choice $i$.

We extract frozen BERT representations from the second-to-last layer of the Transformer.\footnote{Since the domain that BERT was pretrained on (Wikipedia and the BookCorpus \cite{moviebook}) is still quite different from our domain, we fine-tuned BERT on the text of \datasetname~(using the masked language modeling objective, as well as next sentence prediction) for one epoch to account for the domain shift, and then extracted the representations.} Intuitively, this makes sense as the representations that that layer are used for both of BERT's pretraining tasks: next sentence prediction (the unit corresponding to the \texttt{[CLS]} token at the last layer $L$ attends to all units at layer $L-1$), as well as masked language modeling (the unit for a word at layer $L$ looks at its hidden state at the previous layer $L-1$, and uses that to attend to all other units as well). The experiments in \cite{devlin2018bert} suggest that this works well, though not as well as fine-tuning BERT end-to-end or concatenating multiple layers of activations.\footnote{This suggests, however, that if we also fine-tuned BERT along with the rest of the model parameters, the results of \modelname~would be higher.} The tradeoff, however, is that precomputing BERT representations lets us substantially reduce the runtime of \modelname~and allows us to focus on learning more powerful vision representations.

\paragraph{Model Hyperparameters}
A more detailed discussion of the hyperparameters used for \modelname~is as follows. We tried to stick to simple settings (and when possible, used similar configurations for the baselines, particularly with respect to learning rates and hidden state sizes). 

\begin{itemize}[labelwidth=!,itemsep=0pt,topsep=0pt,parsep=1pt]
    \item Our projection of image features maps a 2176 dimensional hidden size (2048 from ResNet50 and 128 dimensional class embeddings) to a 512 dimensional vector.
    \item Our grounding LSTM is a single-layer bidirectional LSTM with a 1280-dimensional input size (768 from BERT and 512 from image features) and uses 256 dimensional hidden states.
    \item Our reasoning LSTM is a two-layer bidirectional LSTM with a 1536-dimensional input size (512 from image features, and 256 for each direction in the attended, grounded query and the grounded answer). It also uses 256-dimensional hidden states.
    \item The representation from the reasoning LSTM, grounded answer, and attended question is maxpooled and projected to a 1024-dimensional vector. That vector is used to predict the $i$th logit.
    \item For all LSTMs, we initialized the hidden-hidden weights using orthogonal initialization \cite{saxe2013exact}, and applied recurrent dropout to the LSTM input with $p_{drop}=0.3$ \cite{gal2016theoretically}. 
    \item The Resnet50 backbone was pretrained on Imagenet \cite{deng2009imagenet,he2016deep}. The parameters in the first three blocks of ResNet were frozen. The final block (after the RoiAlign is applied) is fine-tuned by our model. We were worried, however, that the these representations would drift and so we added an auxiliary loss to the model inspired by \cite{li2017learning}: the 2048-dimensional representation of each object (without class embeddings) had to be predictive of that object's label (via a linear projection to the label space and a softmax).
    \item Often times, there are a lot of objects in the image that are not referred to by the query or response set. We filtered the objects considered by the model to include only the objects mentioned in the query and responses. We also passed in the entire image as an `object' that the model could attend to in the object contextualization layer.
    \item We optimized \modelname~using Adam \cite{Kingma2014AdamAM}, with a learning rate of $2 \cdot 10^{-4}$ and weight decay of $10^{-4}$. Our batch size was 96. We clipped the gradients to have a total $L_2$ norm of at most $1.0$. We lowered the learning rate by a factor of 2 when we noticed a plateau (validation accuracy not increasing for two epochs in a row). Each model was trained for 20 epochs, which took roughly 20 hours over 3 NVIDIA Titan X GPUs.
\end{itemize}

\section{VQA baselines with BERT}\label{sec:vqabert}
We present additional results where baselines for VQA \cite{antol2015vqa} are augmented with BERT embeddings in Table~\ref{tab:baselineswithbert}. We didn't include these results in the main paper, because to the best of our knowledge prior work hasn't used contextualized representations for VQA. (Contextualized representations might be overkill, particularly as VQA questions are short and often simple). From the results, we find that while BERT also helps the baselines, our model \modelname~benefits even more, with a 2.5\% overall boost in the holistic $Q \rightarrow AR$ setting.

\begin{table}[t!]
\vspace{-3mm}
\centering
\begin{small}
\setlength{\tabcolsep}{4pt}
\begin{tabular}{@{}l @{\hspace{0.7em}}|
l@{\hspace{0.7em}}l@{\hspace{0.5em}} |l@{\hspace{0.7em}}l  |l@{\hspace{0.7em}}l@{}}
\multicolumn{1}{c}{} & \multicolumn{2}{c}{$Q \rightarrow A$} & \multicolumn{2}{c}{$QA \rightarrow R$} & \multicolumn{2}{c}{$Q \rightarrow AR$}\\ 
Model & GloVe & BERT & GloVe & BERT &GloVe & BERT\\ \toprule
\textbf{R2C} & \textbf{46.4} & \textbf{63.8} & \textbf{38.3} &\textbf{67.2} & \textbf{18.3} &\textbf{43.1} \vspace{-0.5mm}\\ \midrule
Revisited & 39.4 & 57.5 & 34.0 & 63.5 &13.5  & 36.8  \\ %\spacedhline
BottomUp & 42.8 & 62.3 & 25.1 & 63.0 &10.7 & 39.6 \\ %\spacedhline
MLB & 45.5 &61.8 & 36.1 &65.4 &17.0  &40.6 \\ %\spacedhline
MUTAN & 44.4 &61.0 & 32.0 &64.4 & 14.1 &39.3 \\ %\spacedhline

% & Chance & 25.0 & 25.0 & 25.0 & 25.0 & \phantom{0}6.2 &  \phantom{0}6.2 \\ %\spacedhline
% \midrule
% \multirow{4}{*}{\rotatebox[origin=c]{90}{Text Only}}& BERT & 53.8 & 53.9 & 64.1 & 64.5 & 34.8 & 35.0 \\ 
% & BERT (response only) & 27.6 & 27.7 & 26.3 & 26.2 & \phantom{0}7.6 & \phantom{0}7.3 \\ 
% & ESIM+ELMo & 45.8 & 45.9 & 55.0 & 55.1 & 25.3 & 25.6 \\ 
% & LSTM+ELMo & 28.1 & 28.3 &  28.7 & 28.5 & \phantom{0}8.3 & \phantom{0}8.4 \\ \midrule
% \multirow{4}{*}{\rotatebox[origin=c]{90}{VQA}} 
% % & BottomUp+BERT & \multicolumn{2}{c}{blah} & 63.0 & 62.9 & \multicolumn{2}{c}{blah} \\ 
% & RevisitedVQA \cite{jabri2016revisiting} & 39.4 & 40.5 & 34.0 & 33.7 & 13.5 & 13.8 \\ 
% & BottomUpTopDown\cite{Anderson2017updown} & 42.8 & 44.1 & 25.1 & 25.1 & 10.7 & 11.0 \\  
% & MLB \cite{Kim2017} & 45.5 & 46.2 & 36.1 & 36.8 & 17.0 & 17.2 \\ 
% & MUTAN \cite{Ben-younes_2017_ICCV} & 44.4 & 45.5 & 32.0 & 32.2 & 14.6 & 14.6 \\ \midrule
% & \modelname &\bf{63.8} & \bf{65.1} & \bf{67.2} & \bf{67.3} & \bf{43.1} & \bf{44.0} \\ \midrule
% % & One turker & & 86.7 & & 87.6 & & 76.0 \\ 
% % & Three turkers & & 90.4 & & 91.2 & & 82.9 \\ 
% & Human & & 91.0 & & 93.0 & & 85.0 \\ 
\bottomrule
\end{tabular}
\end{small}
\caption{VQA baselines evaluated with GloVe or BERT, evaluated on the \datasetname~evaluation set with \modelname~as comparison.  While BERT helps the performance of these baselines, our model still performs the best in every setting.}
\label{tab:baselineswithbert}
\end{table}

\section{\datasetname~Datasheet}\label{sec:datasheet}
A datasheet is a list of questions that accompany datasets that are released, in part so that people think hard about the phenomena in their data \cite{gebru2018datasheets}. In this section, we provide a datasheet for \datasetname.

\subsection{Motivation for Dataset Creation}
\paragraph{Why was the dataset created?}
The dataset was created to study the new task of Visual Commonsense Reasoning: essentially, to have models answer challenging cognition-level questions about images and also to choose a rationale justifying each answer.

\paragraph{Has the dataset been used already?}
Yes, at the time of writing, several groups have submitted models to our leaderboard at \leaderboardlink.

\paragraph{Who funded the dataset??}
\datasetname~was funded via a variety of sources; the biggest sponsor was the IARPA DIVA program through D17PC00343.\footnote{However, the views and conclusions contained herein are those of the authors and should not be interpreted as representing endorsements of IARPA, DOI/IBC, or the U.S. Government.}

\subsection{Dataset Composition}
\paragraph{What are the instances?} Each instance contains an image, a sequence of object regions and classes, a query, and a list of response choices. Exactly one response is correct. There are two sub-tasks to the dataset: in Question Answering ($Q{\rightarrow}A$) the query is a question and the response choices are answers. In Answer Justification ($QA {\rightarrow}R$) the query is a question and the correct answer; the responses are rationales that justify why someone would conclude that the answer is true. Both the query and the rationale refer to the objects using detection tags like \texttt{person1}.

\paragraph{How many instances are there?}
There are 212,923 training questions, 26,534 validation questions, and 25,263 questions. Each is associated with a four answer choices, and each question+correct answer is associated with four rationale choices.

\paragraph{What data does each instance consist of?}
The image from each instance comes from a movie, while the object detector was trained to detect objects in the COCO dataset \cite{lin2014microsoft}. Workers ask challenging high-level questions covering a wide variety of cognition-level phenomena. Then, workers provide a rationale: one to several sentences explaining how they came at their decision. The rationale points to details in the image, as well as background knowledge about how the world works. Each instance contains one correct answer and three incorrect counterfactual answers, along with one correct rationale and three incorrect rationales.

\paragraph{Does the data rely on external resources?} No, everything is included.

\paragraph{Are there recommended data splits or evaluation measures?} We release the training and validation sets, as well as the test set without labels. For the test set, researchers can submit their predictions to a public leaderboard. Evaluation is fairly straightforward as our task is multiple choice, but we will also release an evaluation script.

\subsection{Data Collection Process}
\paragraph{How was the data collected?}
We used movie images, with objects detected using Mask RCNN \cite{Detectron2018, He2017MaskR}. We collected the questions, answers, and rationales on Amazon Mechanical Turk.

\paragraph{Who was involved in the collection process and what were their roles?}
We (the authors) did several rounds of pilot studies, and collected data at scale on Amazon Mechanical Turk. In the task, workers on Amazon Mechanical Turk could ask anywhere between one to three questions. For each question, they had to provide an answer, indicate its likelihood on an ordinal scale, and provide a rationale justifying why their answer is true. Workers were paid at 22 cents per question, answer, and rationale.

\paragraph{Over what time frame was the data collected?}
August to October 2018.

\paragraph{Does the dataset contain all possible instances?}
No. Visual Commonsense Inference is very broad, and we focused on a limited set of (interesting) phenomena. Beyond looking at different types of movies, or looking at the world beyond still photographs, there are also different types of inferences that we didn't cover in our work.

\paragraph{If the dataset is a sample, then what is the population?}
The population is that of movie images that were deemed interesting by our interestingness filter (having at least three object detections, of which at least two are people).

\subsection{Data Preprocessing}
\paragraph{What preprocessing was done?}
The line between data preprocessing and dataset collection is blurry for \datasetname. After obtaining crowdsourced questions, answers, and rationales, we applied \evaluationname, turning raw data into a multiple choice task. We also tokenized the text spans.

\paragraph{Was the raw data saved in addition to the cleaned data?}
Yes - the raw data is the correct answers (and as such is a subset of the `cleaned' data).

\paragraph{Does this dataset collection/preprocessing procedure achieve the initial motivation?}
At this point, we think so. Our dataset is challenging for existing VQA systems, but easy for humans.

\subsection{Dataset Distribution}
\paragraph{How is the dataset distributed?}
\datasetname~is freely available for research use at \websitelink.

% \paragraph{What license (if any) is it distributed under?}
% The license info is located at \websitelink.

\subsection{Legal and Ethical Considerations}
\paragraph{Were workers told what the dataset would be used for and did they consent?}
Yes - the instructions said that workers answers would be used in a dataset. We tried to be as upfront as possible to workers. Workers also consented to have their responses used in this way through the Amazon Mechanical Turk Participation Agreement.

\paragraph{If it relates to people, could this dataset expose people to harm or legal action?}
No - the questions, answers, and responses don't contain personal info about the crowd workers.

\paragraph{If it relates to people, does it unfairly advantage or disadvantage a particular social group?}
Unfortunately, movie data is highly biased against women and minorities \cite{schofield2016gender,sap2017connotation}. Our data, deriving from movies as well as from worker elicitations \cite{rudinger2017social}, is no different. For these reasons, we recommend that users do not deploy models trained on \datasetname~in the real world.

\section{Additional qualitative results}\label{sec:qualresults2}

\begin{figure*}[ht]
    \vspace{-3mm}
    \centering
    \includegraphics[width=\linewidth]{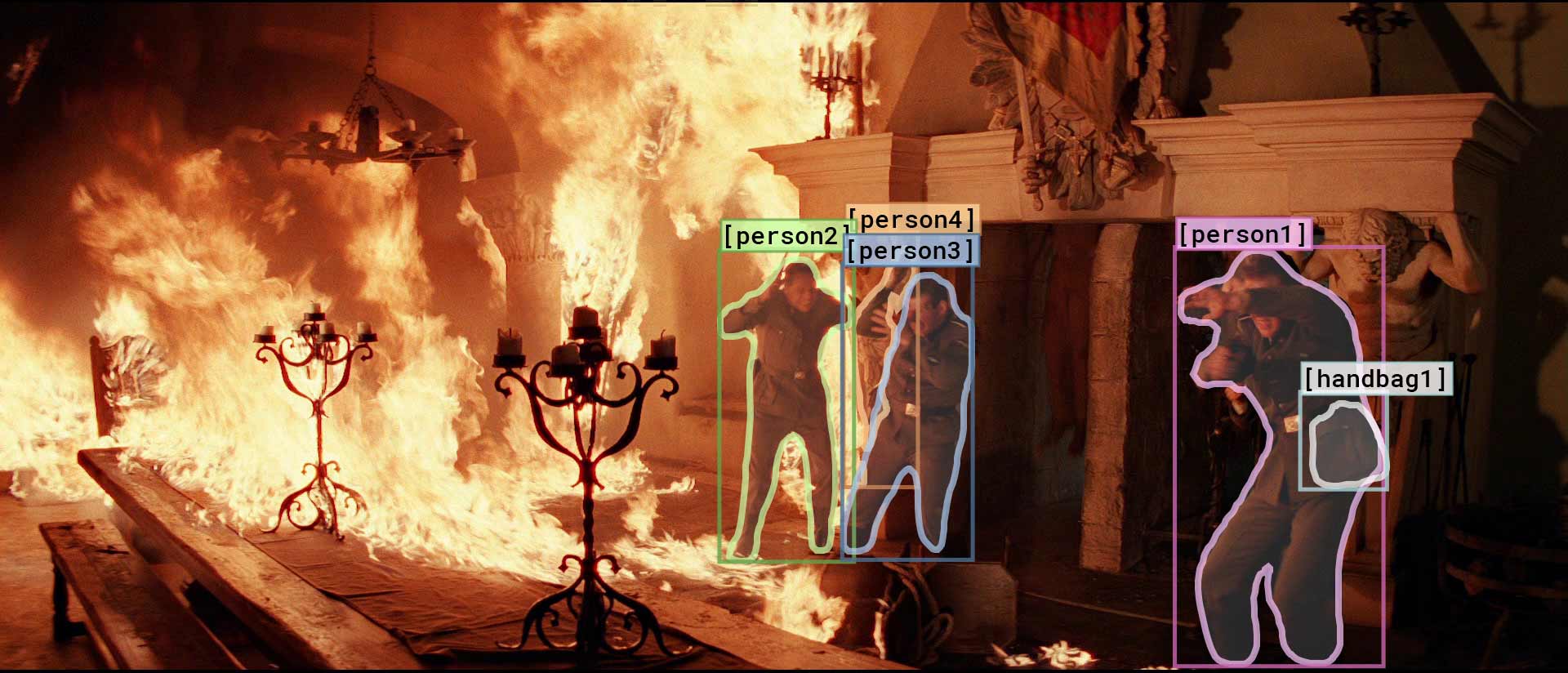}
    
    \vspace*{0.2cm}
    \includegraphics[width=\linewidth]{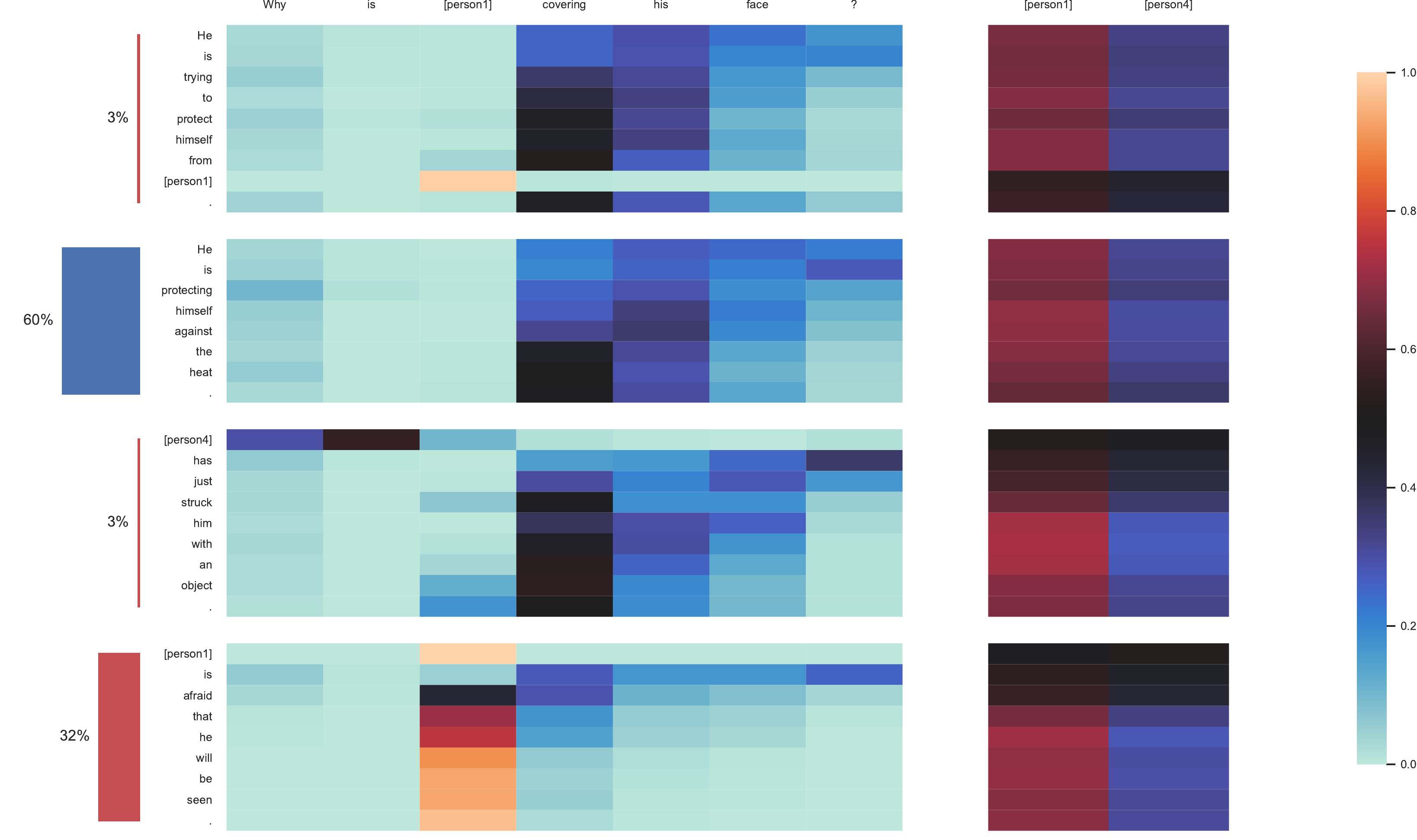}
    \caption{An example from the $Q \rightarrow A$ task. Each super-row is a response choice (four in total). The first super-column is the question: Here, `Why is \texttt{[person1] covering his face?}' and the second super-column represents the relevant objects in the image that \modelname~attends to. Accordingly, each block is a heatmap of the attention between each response choice and the query, as well as each response choice and the objects. The final prediction is given by the bar graph on the left: The model is 60\% confident that the right answer is \textbf{b.}, which is correct.}
    \label{fig:heatmap26qa}
\end{figure*}

In this section, we present additional qualitative results from \modelname. Our use of attention mechanisms allow us to better gain insight into how the model arrives at its decisions. In particular, the model uses the answer to attend over the question, and it uses the answer to attend over relevant objects in the image. Looking at the attention maps help to visualize which items in the question are important (usually, the model focuses on the second half of the question, like `covering his face' in Figure~\ref{fig:heatmap26qa}), as well as which objects are important (usually, the objects referred to by the answer are assigned the most weight).
\clearpage

% \afterpage{%
% \newpage
% }

% \afterpage{%
\begin{figure*}[ht]
    \vspace{-3mm}
    \centering
    \includegraphics[width=\columnwidth]{figures/img26.jpg}
    
    \vspace*{0.2cm}
    \includegraphics[width=\linewidth]{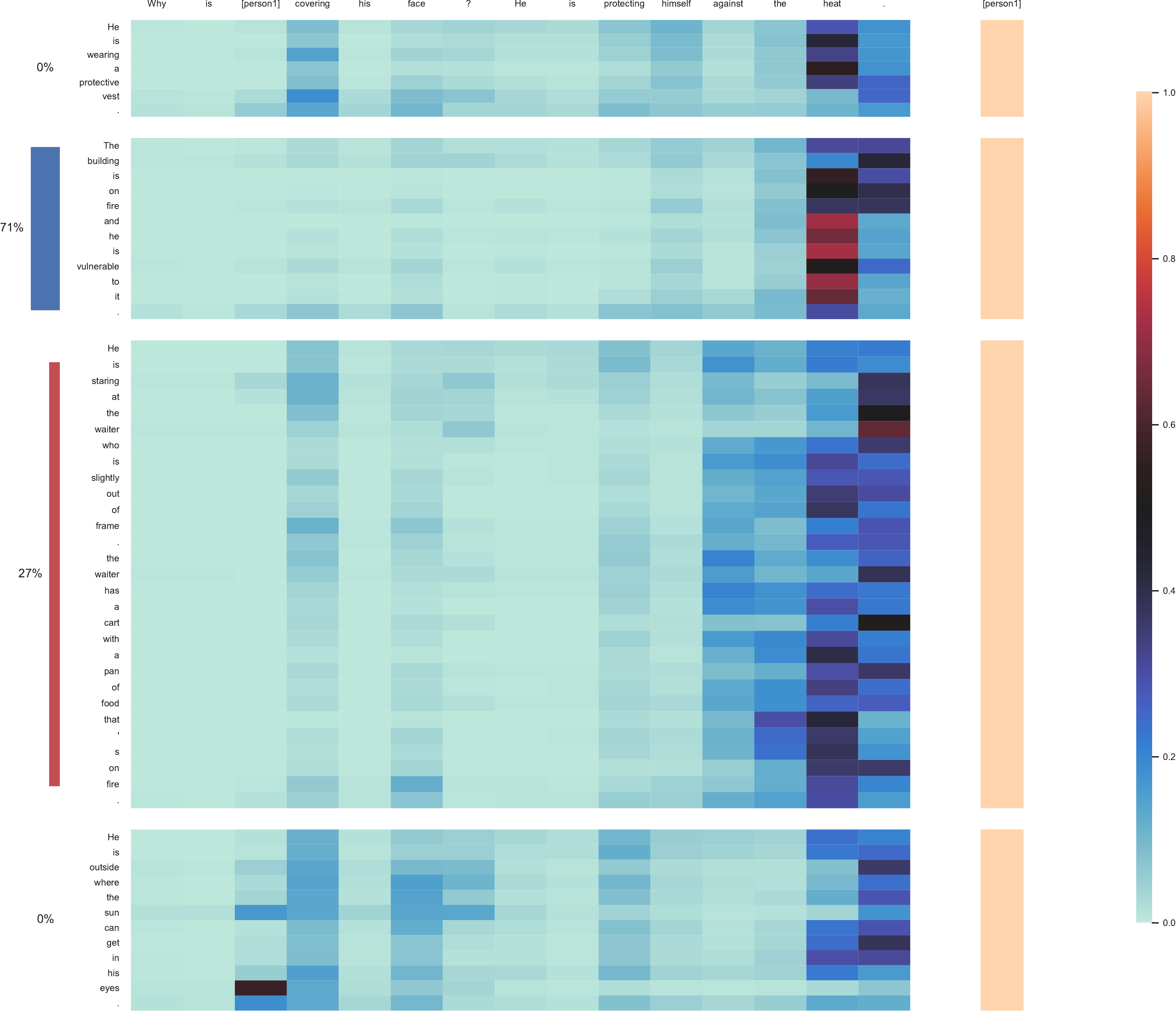}
    \caption{An example from the $QA \rightarrow R$ task. Each super-row is a response choice (four in total). The first super-column is the query, and the second super-column holds the relevant objects (here just a single person, as no other objects were mentioned by the query or responses). Each block is a heatmap of the attention between each response choice and the query, as well as the attention between each response choice and the objects. The final prediction is given by the bar graph on the left: The model is 71\% confident that the right rationale is \textbf{b.}, which is correct.}
    \label{fig:heatmap26qar}
\end{figure*}
% \newpage
% }

% \afterpage{%
\begin{figure*}[ht]
    \vspace{-3mm}
    \centering
    \includegraphics[width=\linewidth]{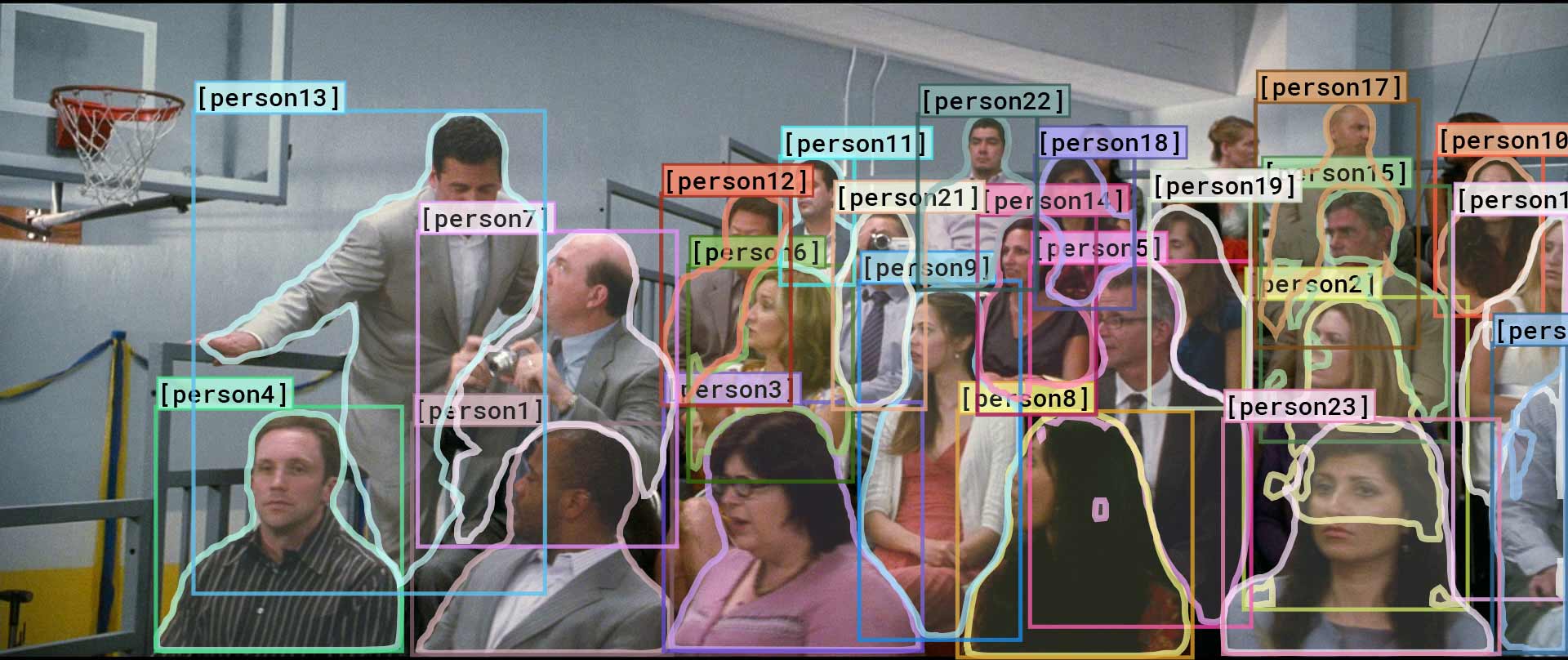}
    
    \vspace*{0.2cm}
    \includegraphics[width=\linewidth]{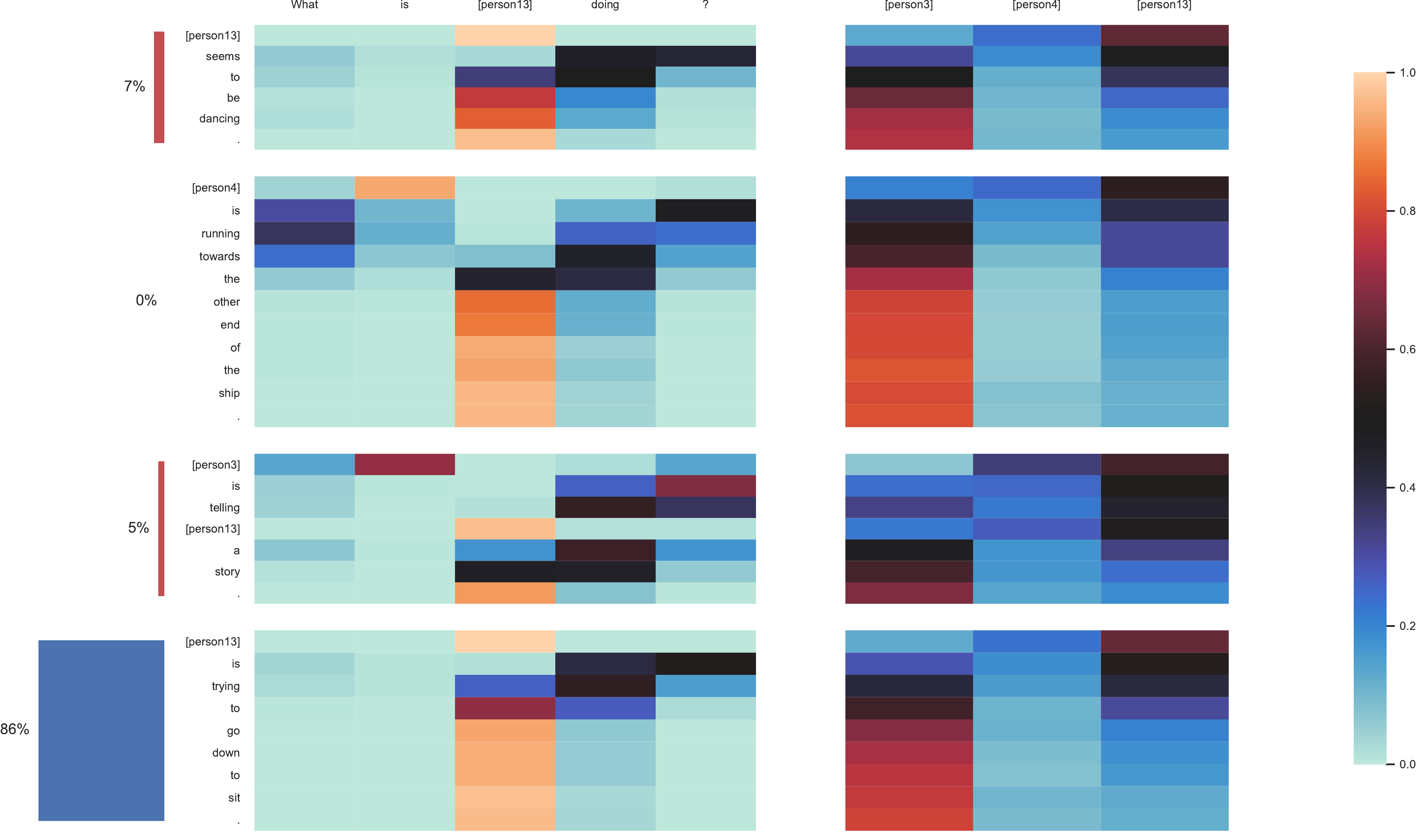}
    \caption{An example from the $Q \rightarrow A$ task. Each super-row is a response choice (four in total). The first super-column is the question: Here, `What is \texttt{[person13] doing?}' and the second super-column represents the relevant objects in the image that \modelname~attends to. Accordingly, each block is a heatmap of the attention between each response choice and the query, as well as each response choice and the objects. The final prediction is given by the bar graph on the left: The model is 86\% confident that the right answer is \textbf{d.}, which is correct.}
    \label{fig:heatmap53qa}
\end{figure*}
% \newpage
% }

% \afterpage{%
\begin{figure*}[ht]
    \vspace{-4mm}
    \centering
    \includegraphics[width=3cm]{figures/ex53.jpg}
    
    \vspace*{0.01cm}
    \includegraphics[width=\linewidth]{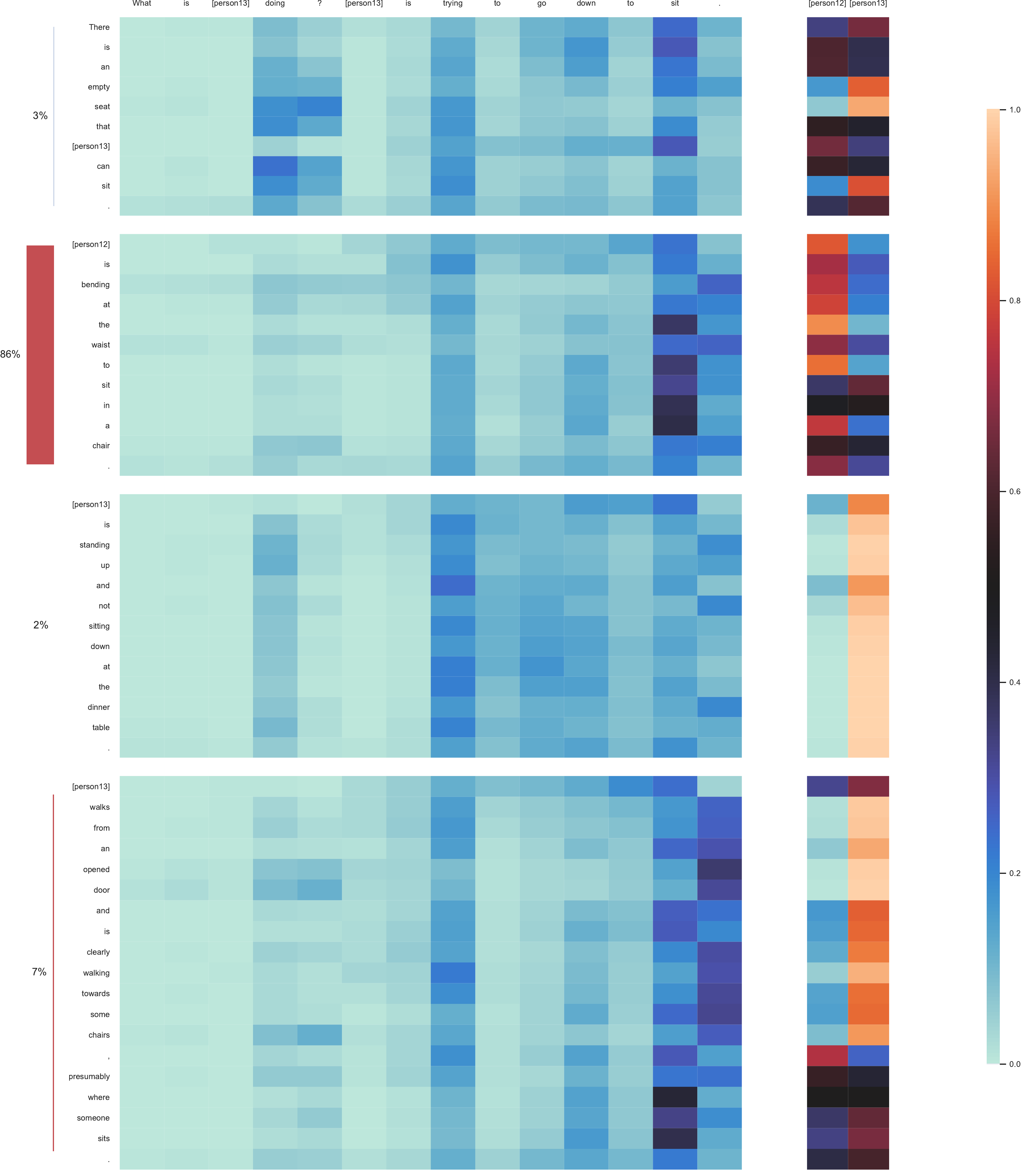}
    
    \vspace*{-0.2cm}
    \caption{An example from the $QA \rightarrow R$ task. Each super-row is a response choice (four in total). The first super-column is the query, and the second super-column holds the relevant objects. Each block is a heatmap of the attention between each response choice and the query, as well as the attention between each response choice and the objects. The final prediction is given by the bar graph on the left: The model is 86\% confident that the right rationale is b., which is incorrect - the correct choice is \textbf{a.}}
    \label{fig:heatmap53qar}
\end{figure*}
% \newpage
% }

% \afterpage{%
\begin{figure*}[ht]
    \vspace{-3mm}
    \centering
    \includegraphics[width=\linewidth]{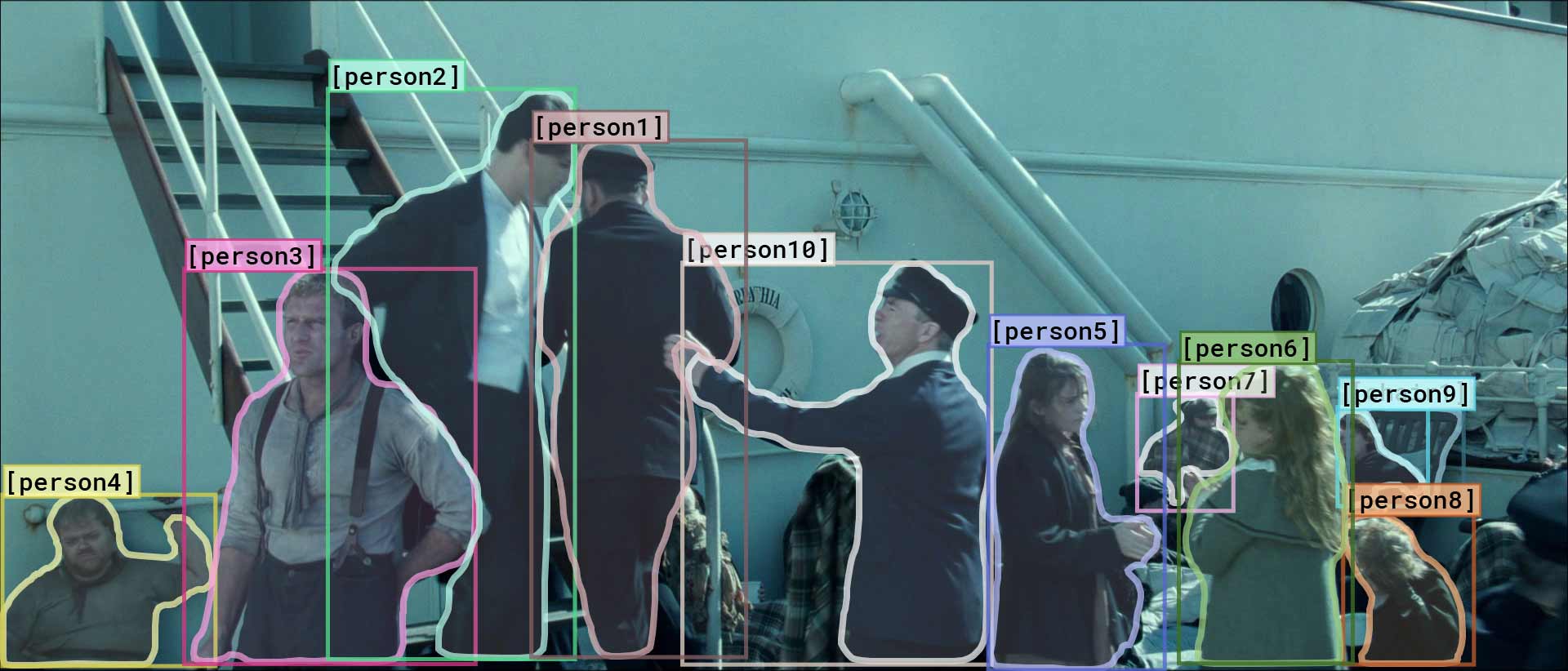}
    
    \vspace*{0.2cm}
    \includegraphics[width=\linewidth]{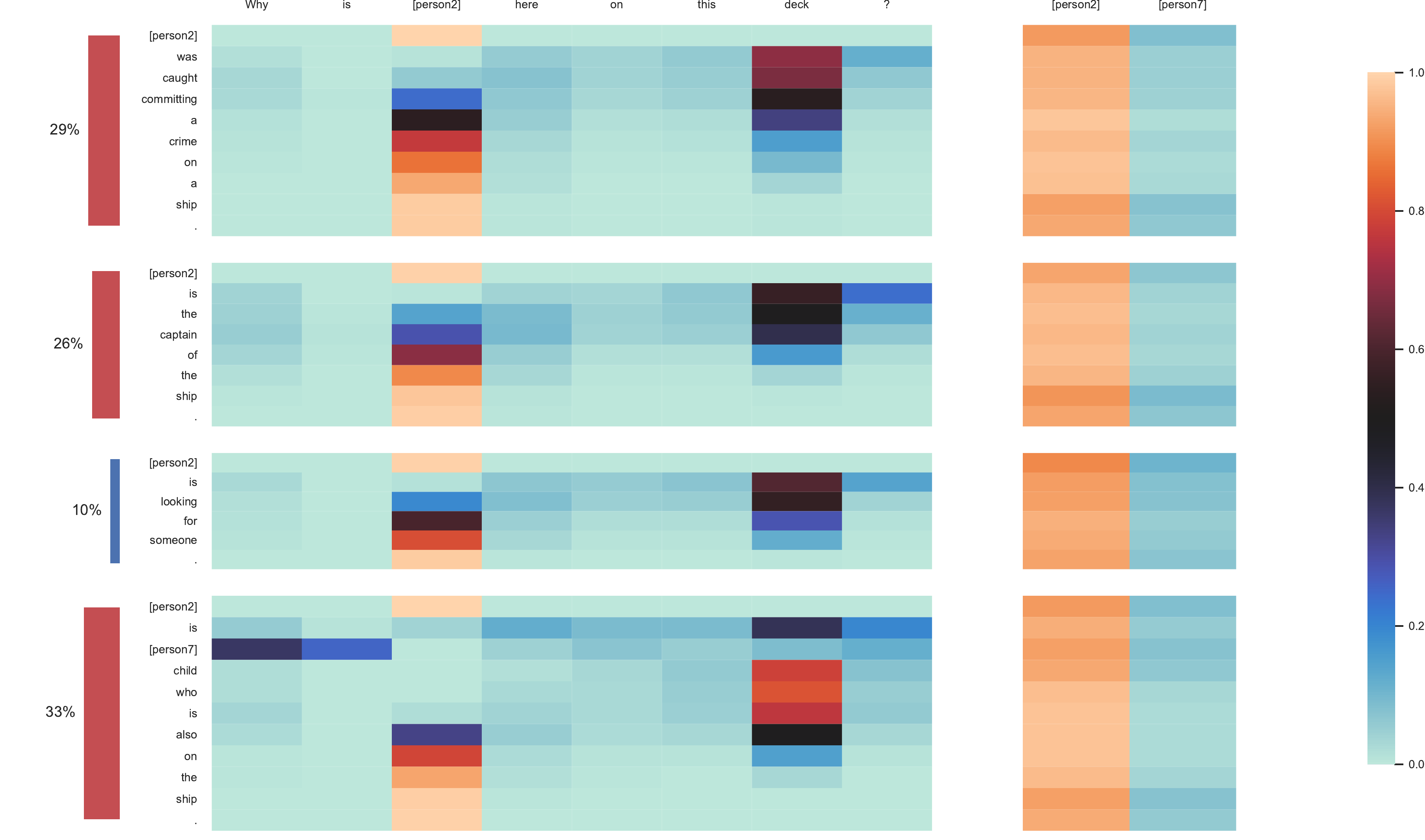}
    \caption{An example from the $Q \rightarrow A$ task. Each super-row is a response choice (four in total). The first super-column is the question: Here, `Why is \texttt{[person2] here on this deck?}' and the second super-column represents the relevant objects in the image that \modelname~attends to. Accordingly, each block is a heatmap of the attention between each response choice and the query, as well as each response choice and the objects. The final prediction is given by the bar graph on the left: The model is 33\% confident that the right answer is d., which is incorrect - the correct answer is correct answer is \textbf{c.}}
    \label{fig:heatmap132qa}
\end{figure*}
% \newpage
% }

% \afterpage{%
\begin{figure*}[ht]
    \vspace{-3mm}
    \centering
    \includegraphics[width=3cm]{figures/ex132.jpg}
    
    \vspace*{0.2cm}
    \includegraphics[width=\linewidth]{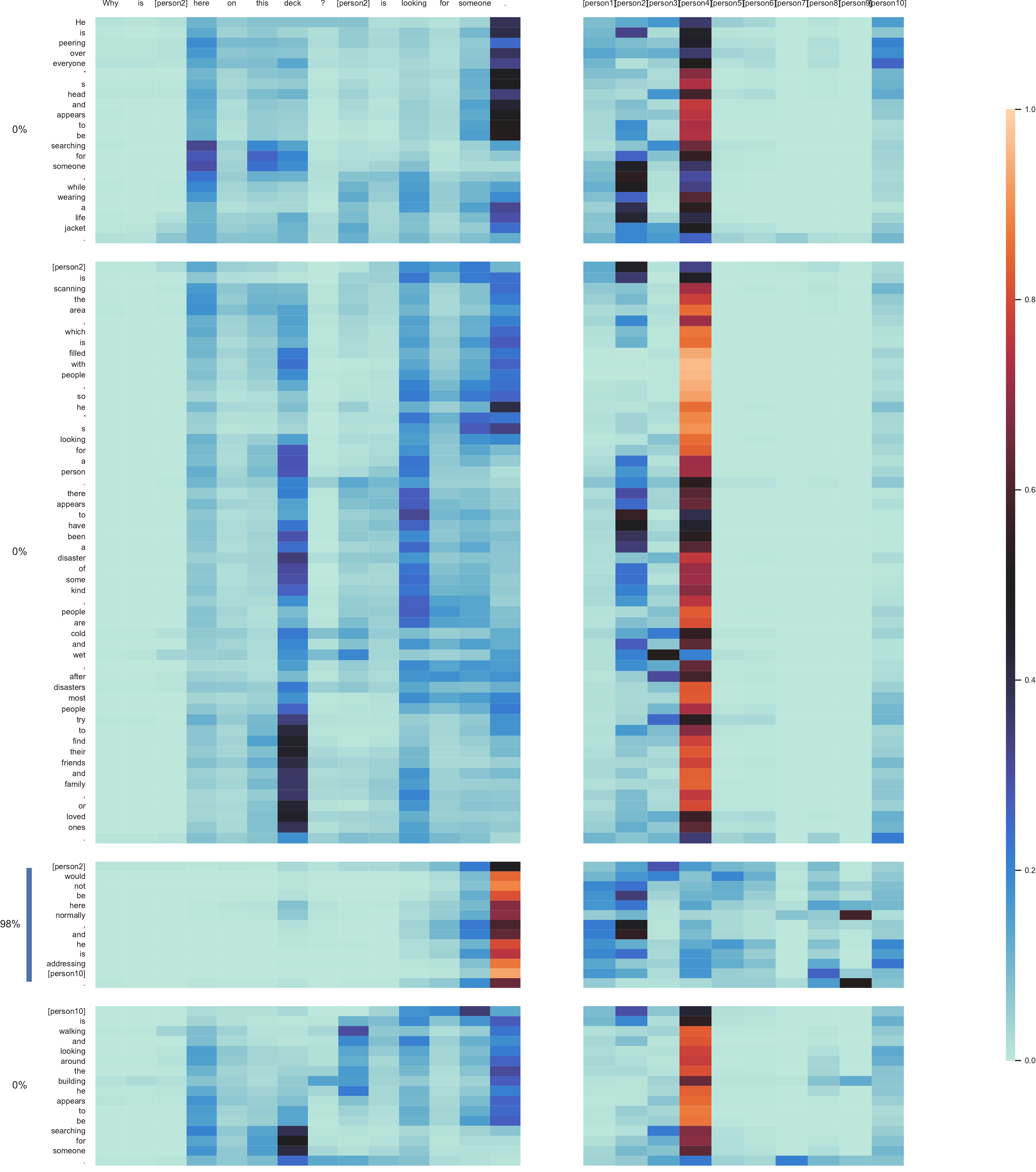}
    \caption{An example from the $QA \rightarrow R$ task. Each super-row is a response choice (four in total). The first super-column is the query, and the second super-column holds the relevant objects. Each block is a heatmap of the attention between each response choice and the query, as well as the attention between each response choice and the objects. The final prediction is given by the bar graph on the left: The model is 98\% confident that the right rationale is \textbf{c.}, which is correct.}    \label{fig:heatmap132qar}
\end{figure*}
% \newpage
% }

% \section{Stuff to go in the appendix}
\clearpage

{\small
\bibliographystyle{ieee}
\bibliography{cvpr2019}

\begin{thebibliography}{10}\itemsep=-1pt

\bibitem{agrawal2018don}
Aishwarya Agrawal, Dhruv Batra, Devi Parikh, and Aniruddha Kembhavi.
\newblock Don’t just assume; look and answer: Overcoming priors for visual
  question answering.
\newblock In {\em Proceedings of the IEEE Conference on Computer Vision and
  Pattern Recognition}, pages 4971--4980, 2018.

\bibitem{alahi2016social}
Alexandre Alahi, Kratarth Goel, Vignesh Ramanathan, Alexandre Robicquet, Li
  Fei-Fei, and Silvio Savarese.
\newblock Social lstm: Human trajectory prediction in crowded spaces.
\newblock In {\em Proceedings of the IEEE Conference on Computer Vision and
  Pattern Recognition}, pages 961--971, 2016.

\bibitem{alayrac2016unsupervised}
Jean-Baptiste Alayrac, Piotr Bojanowski, Nishant Agrawal, Josef Sivic, Ivan
  Laptev, and Simon Lacoste-Julien.
\newblock Unsupervised learning from narrated instruction videos.
\newblock In {\em Proceedings of the IEEE Conference on Computer Vision and
  Pattern Recognition}, pages 4575--4583, 2016.

\bibitem{Anderson2017updown}
Peter Anderson, Xiaodong He, Chris Buehler, Damien Teney, Mark Johnson, Stephen
  Gould, and Lei Zhang.
\newblock Bottom-up and top-down attention for image captioning and visual
  question answering.
\newblock In {\em CVPR}, 2018.

\bibitem{antol2015vqa}
Stanislaw Antol, Aishwarya Agrawal, Jiasen Lu, Margaret Mitchell, Dhruv Batra,
  C Lawrence~Zitnick, and Devi Parikh.
\newblock Vqa: Visual question answering.
\newblock In {\em Proceedings of the IEEE international conference on computer
  vision}, pages 2425--2433, 2015.

\bibitem{Ben-younes_2017_ICCV}
Hedi Ben-younes, Remi Cadene, Matthieu Cord, and Nicolas Thome.
\newblock {MUTAN}: {M}ultimodal {T}ucker {F}usion for {V}isual {Q}uestion
  {A}nswering.
\newblock In {\em The IEEE International Conference on Computer Vision (ICCV)},
  Oct 2017.

\bibitem{biran2017explanation}
Or Biran and Courtenay Cotton.
\newblock Explanation and justification in machine learning: A survey.
\newblock In {\em IJCAI-17 Workshop on Explainable AI (XAI)}, page~8, 2017.

\bibitem{bowman2015snli}
Samuel~R. Bowman, Gabor Angeli, Christopher Potts, and Christopher~D. Manning.
\newblock A large annotated corpus for learning natural language inference.
\newblock In {\em Proceedings of the 2015 Conference on Empirical Methods in
  Natural Language Processing, {EMNLP} 2015, Lisbon, Portugal, September 17-21,
  2015}, pages 632--642, 2015.

\bibitem{chandrasekaran2018explanations}
Arjun Chandrasekaran, Viraj Prabhu, Deshraj Yadav, Prithvijit Chattopadhyay,
  and Devi Parikh.
\newblock Do explanations make vqa models more predictable to a human?
\newblock In {\em Proceedings of the 2018 Conference on Empirical Methods in
  Natural Language Processing}, pages 1036--1042, 2018.

\bibitem{chen2017enhanced}
Qian Chen, Xiaodan Zhu, Zhen-Hua Ling, Si Wei, Hui Jiang, and Diana Inkpen.
\newblock Enhanced lstm for natural language inference.
\newblock In {\em Proceedings of the 55th Annual Meeting of the Association for
  Computational Linguistics (Volume 1: Long Papers)}, volume~1, pages
  1657--1668, 2017.

\bibitem{cho2014learning}
Kyunghyun Cho, Bart van Merrienboer, Caglar Gulcehre, Dzmitry Bahdanau, Fethi
  Bougares, Holger Schwenk, and Yoshua Bengio.
\newblock Learning phrase representations using rnn encoder--decoder for
  statistical machine translation.
\newblock In {\em Proceedings of the 2014 Conference on Empirical Methods in
  Natural Language Processing (EMNLP)}, pages 1724--1734, 2014.

\bibitem{ActProperly2018}
Ching-Yao Chuang, Jiaman Li, Antonio Torralba, and Sanja Fidler.
\newblock Learning to act properly: Predicting and explaining affordances from
  images.
\newblock In {\em CVPR}, 2018.

\bibitem{Davis2015CommonsenseRA}
Ernest Davis and Gary Marcus.
\newblock Commonsense reasoning and commonsense knowledge in artificial
  intelligence.
\newblock {\em Commun. ACM}, 58:92--103, 2015.

\bibitem{deng2009imagenet}
Jia Deng, Wei Dong, Richard Socher, Li-Jia Li, Kai Li, and Li Fei-Fei.
\newblock Imagenet: A large-scale hierarchical image database.
\newblock In {\em Computer Vision and Pattern Recognition, 2009. CVPR 2009.
  IEEE Conference on}, pages 248--255. Ieee, 2009.

\bibitem{devlin2018bert}
Jacob Devlin, Ming-Wei Chang, Kenton Lee, and Kristina Toutanova.
\newblock Bert: Pre-training of deep bidirectional transformers for language
  understanding.
\newblock {\em arXiv preprint arXiv:1810.04805}, 2018.

\bibitem{Devlin2015ExploringNN}
Jacob Devlin, Saurabh Gupta, Ross~B. Girshick, Margaret Mitchell, and
  C.~Lawrence Zitnick.
\newblock Exploring nearest neighbor approaches for image captioning.
\newblock {\em CoRR}, abs/1505.04467, 2015.

\bibitem{Ehsani_2018_CVPR}
Kiana Ehsani, Hessam Bagherinezhad, Joseph Redmon, Roozbeh Mottaghi, and Ali
  Farhadi.
\newblock Who let the dogs out? modeling dog behavior from visual data.
\newblock In {\em The IEEE Conference on Computer Vision and Pattern
  Recognition (CVPR)}, June 2018.

\bibitem{felsen2017will}
Panna Felsen, Pulkit Agrawal, and Jitendra Malik.
\newblock What will happen next? forecasting player moves in sports videos.
\newblock In {\em Proceedings of the IEEE Conference on Computer Vision and
  Pattern Recognition}, pages 3342--3351, 2017.

\bibitem{andrew_flowers_most_2015}
Andrew Flowers.
\newblock The {Most} {Common} {Unisex} {Names} {In} {America}: {Is} {Yours}
  {One} {Of} {Them}?, June 2015.

\bibitem{Fouhey18}
David~F. Fouhey, Weicheng Kuo, Alexei~A. Efros, and Jitendra Malik.
\newblock From lifestyle vlogs to everyday interactions.
\newblock In {\em CVPR}, 2018.

\bibitem{gal2016theoretically}
Yarin Gal and Zoubin Ghahramani.
\newblock A theoretically grounded application of dropout in recurrent neural
  networks.
\newblock In {\em Advances in neural information processing systems}, pages
  1019--1027, 2016.

\bibitem{Gardner2017AllenNLP}
Matt Gardner, Joel Grus, Mark Neumann, Oyvind Tafjord, Pradeep Dasigi,
  Nelson~F. Liu, Matthew Peters, Michael Schmitz, and Luke~S. Zettlemoyer.
\newblock Allennlp: A deep semantic natural language processing platform.
\newblock 2017.

\bibitem{gebru2018datasheets}
Timnit Gebru, Jamie Morgenstern, Briana Vecchione, Jennifer~Wortman Vaughan,
  Hanna Wallach, Hal Daume{\'e}~III, and Kate Crawford.
\newblock Datasheets for datasets.
\newblock {\em arXiv preprint arXiv:1803.09010}, 2018.

\bibitem{Detectron2018}
Ross Girshick, Ilija Radosavovic, Georgia Gkioxari, Piotr Doll\'{a}r, and
  Kaiming He.
\newblock Detectron.
\newblock \url{https://github.com/facebookresearch/detectron}, 2018.

\bibitem{goyal2017making}
Yash Goyal, Tejas Khot, Douglas Summers-Stay, Dhruv Batra, and Devi Parikh.
\newblock Making the v in vqa matter: Elevating the role of image understanding
  in visual question answering.
\newblock In {\em CVPR}, volume~1, page~9, 2017.

\bibitem{balanced_vqa_v2}
Yash Goyal, Tejas Khot, Douglas Summers{-}Stay, Dhruv Batra, and Devi Parikh.
\newblock Making the {V} in {VQA} matter: Elevating the role of image
  understanding in {V}isual {Q}uestion {A}nswering.
\newblock In {\em Conference on Computer Vision and Pattern Recognition
  (CVPR)}, 2017.

\bibitem{Gupta_2018_CVPR}
Agrim Gupta, Justin Johnson, Li Fei-Fei, Silvio Savarese, and Alexandre Alahi.
\newblock Social gan: Socially acceptable trajectories with generative
  adversarial networks.
\newblock In {\em The IEEE Conference on Computer Vision and Pattern
  Recognition (CVPR)}, June 2018.

\bibitem{gururangan2018annotation}
Suchin Gururangan, Swabha Swayamdipta, Omer Levy, Roy Schwartz, Samuel~R.
  Bowman, and Noah~A. Smith.
\newblock Annotation artifacts in natural language inference data.
\newblock In {\em Proc. of NAACL}, 2018.

\bibitem{He2017MaskR}
Kaiming He, Georgia Gkioxari, Piotr Doll{\'a}r, and Ross~B. Girshick.
\newblock Mask r-cnn.
\newblock {\em 2017 IEEE International Conference on Computer Vision (ICCV)},
  pages 2980--2988, 2017.

\bibitem{he2016deep}
Kaiming He, Xiangyu Zhang, Shaoqing Ren, and Jian Sun.
\newblock Deep residual learning for image recognition.
\newblock In {\em Proceedings of the IEEE conference on computer vision and
  pattern recognition}, pages 770--778, 2016.

\bibitem{hendricks2016generating}
Lisa~Anne Hendricks, Zeynep Akata, Marcus Rohrbach, Jeff Donahue, Bernt
  Schiele, and Trevor Darrell.
\newblock Generating visual explanations.
\newblock In {\em European Conference on Computer Vision}, pages 3--19.
  Springer, 2016.

\bibitem{hendricks2018grounding}
Lisa~Anne Hendricks, Ronghang Hu, Trevor Darrell, and Zeynep Akata.
\newblock Grounding visual explanations.
\newblock {\em European Conference on Computer Vision (ECCV)}, 2018.

\bibitem{hendricks17iccv}
Lisa~Anne Hendricks, Oliver Wang, Eli Shechtman, Josef Sivic, Trevor Darrell,
  and Bryan Russell.
\newblock Localizing moments in video with natural language.
\newblock In {\em Proceedings of the IEEE International Conference on Computer
  Vision (ICCV)}, 2017.

\bibitem{Hochreiter:1997:LSM:1246443.1246450}
Sepp Hochreiter and J\"{u}rgen Schmidhuber.
\newblock Long short-term memory.
\newblock {\em Neural Comput.}, 9(8):1735--1780, Nov. 1997.

\bibitem{hu2018explainable}
Ronghang Hu, Jacob Andreas, Trevor Darrell, and Kate Saenko.
\newblock Explainable neural computation via stack neural module networks.
\newblock In {\em Proceedings of the European Conference on Computer Vision
  (ECCV)}, pages 53--69, 2018.

\bibitem{hu2017modeling}
Ronghang Hu, Marcus Rohrbach, Jacob Andreas, Trevor Darrell, and Kate Saenko.
\newblock Modeling relationships in referential expressions with compositional
  modular networks.
\newblock In {\em Computer Vision and Pattern Recognition (CVPR), 2017 IEEE
  Conference on}, pages 4418--4427. IEEE, 2017.

\bibitem{Park_2018_CVPR}
Dong Huk~Park, Lisa Anne~Hendricks, Zeynep Akata, Anna Rohrbach, Bernt Schiele,
  Trevor Darrell, and Marcus Rohrbach.
\newblock Multimodal explanations: Justifying decisions and pointing to the
  evidence.
\newblock In {\em The IEEE Conference on Computer Vision and Pattern
  Recognition (CVPR)}, June 2018.

\bibitem{jabri2016revisiting}
Allan Jabri, Armand Joulin, and Laurens van~der Maaten.
\newblock Revisiting visual question answering baselines.
\newblock In {\em European conference on computer vision}, pages 727--739.
  Springer, 2016.

\bibitem{jang2017tgif}
Yunseok Jang, Yale Song, Youngjae Yu, Youngjin Kim, and Gunhee Kim.
\newblock Tgif-qa: Toward spatio-temporal reasoning in visual question
  answering.
\newblock In {\em IEEE Conference on Computer Vision and Pattern Recognition
  (CVPR 2017). Honolulu, Hawaii}, pages 2680--8, 2017.

\bibitem{jonker1987shortest}
Roy Jonker and Anton Volgenant.
\newblock A shortest augmenting path algorithm for dense and sparse linear
  assignment problems.
\newblock {\em Computing}, 38(4):325--340, 1987.

\bibitem{kim2018textual}
Jinkyu Kim, Anna Rohrbach, Trevor Darrell, John Canny, and Zeynep Akata.
\newblock Textual explanations for self-driving vehicles.
\newblock In {\em 15th European Conference on Computer Vision}, pages 577--593.
  Springer, 2018.

\bibitem{Kim2017}
Jin-Hwa Kim, Kyoung~Woon On, Woosang Lim, Jeonghee Kim, Jung-Woo Ha, and
  Byoung-Tak Zhang.
\newblock {Hadamard Product for Low-rank Bilinear Pooling}.
\newblock In {\em The 5th International Conference on Learning
  Representations}, 2017.

\bibitem{kim2016pororoqa}
K Kim, C Nan, MO Heo, SH Choi, and BT Zhang.
\newblock Pororoqa: Cartoon video series dataset for story understanding.
\newblock In {\em Proceedings of NIPS 2016 Workshop on Large Scale Computer
  Vision System}, 2016.

\bibitem{Kingma2014AdamAM}
Diederik~P. Kingma and Jimmy Ba.
\newblock Adam: A method for stochastic optimization.
\newblock {\em CoRR}, abs/1412.6980, 2014.

\bibitem{visualgenome}
Ranjay Krishna, Yuke Zhu, Oliver Groth, Justin Johnson, Kenji Hata, Joshua
  Kravitz, Stephanie Chen, Yannis Kalantidis, Li-Jia Li, David~A Shamma, et~al.
\newblock Visual genome: Connecting language and vision using crowdsourced
  dense image annotations.
\newblock {\em International Journal of Computer Vision}, 123(1):32--73, 2017.

\bibitem{lei2018tvqa}
Jie Lei, Licheng Yu, Mohit Bansal, and Tamara~L Berg.
\newblock Tvqa: Localized, compositional video question answering.
\newblock In {\em EMNLP}, 2018.

\bibitem{li2016tgif}
Yuncheng Li, Yale Song, Liangliang Cao, Joel Tetreault, Larry Goldberg,
  Alejandro Jaimes, and Jiebo Luo.
\newblock Tgif: A new dataset and benchmark on animated gif description.
\newblock In {\em Proceedings of the IEEE Conference on Computer Vision and
  Pattern Recognition}, pages 4641--4650, 2016.

\bibitem{li2017learning}
Zhizhong Li and Derek Hoiem.
\newblock Learning without forgetting.
\newblock {\em IEEE Transactions on Pattern Analysis and Machine Intelligence},
  2017.

\bibitem{lin2014microsoft}
Tsung-Yi Lin, Michael Maire, Serge Belongie, James Hays, Pietro Perona, Deva
  Ramanan, Piotr Doll{\'a}r, and C~Lawrence Zitnick.
\newblock Microsoft coco: Common objects in context.
\newblock In {\em European conference on computer vision}, pages 740--755.
  Springer, 2014.

\bibitem{lin2016leveraging}
Xiao Lin and Devi Parikh.
\newblock Leveraging visual question answering for image-caption ranking.
\newblock In {\em European Conference on Computer Vision}, pages 261--277.
  Springer, 2016.

\bibitem{maharaj2017dataset}
Tegan Maharaj, Nicolas Ballas, Anna Rohrbach, Aaron~C Courville, and
  Christopher~Joseph Pal.
\newblock A dataset and exploration of models for understanding video data
  through fill-in-the-blank question-answering.
\newblock In {\em Computer Vision and Pattern Recognition (CVPR)}, 2017.

\bibitem{mao2016generation}
Junhua Mao, Jonathan Huang, Alexander Toshev, Oana Camburu, Alan~L Yuille, and
  Kevin Murphy.
\newblock Generation and comprehension of unambiguous object descriptions.
\newblock In {\em Proceedings of the IEEE conference on computer vision and
  pattern recognition}, pages 11--20, 2016.

\bibitem{mostafazadeh_corpus_2016}
Nasrin Mostafazadeh, Nathanael Chambers, Xiaodong He, Devi Parikh, Dhruv Batra,
  Lucy Vanderwende, Pushmeet Kohli, and James Allen.
\newblock A corpus and evaluation framework for deeper understanding of
  commonsense stories.
\newblock {\em arXiv preprint arXiv:1604.01696}, 2016.

\bibitem{mottaghi2016happens}
Roozbeh Mottaghi, Mohammad Rastegari, Abhinav Gupta, and Ali Farhadi.
\newblock “what happens if...” learning to predict the effect of forces in
  images.
\newblock In {\em European Conference on Computer Vision}, pages 269--285.
  Springer, 2016.

\bibitem{munkres1957algorithms}
James Munkres.
\newblock Algorithms for the assignment and transportation problems.
\newblock {\em Journal of the society for industrial and applied mathematics},
  5(1):32--38, 1957.

\bibitem{pennington2014glove}
Jeffrey Pennington, Richard Socher, and Christopher Manning.
\newblock Glove: Global vectors for word representation.
\newblock In {\em Proceedings of the 2014 conference on empirical methods in
  natural language processing (EMNLP)}, pages 1532--1543, 2014.

\bibitem{peters2018deep}
Matthew Peters, Mark Neumann, Mohit Iyyer, Matt Gardner, Christopher Clark,
  Kenton Lee, and Luke Zettlemoyer.
\newblock Deep contextualized word representations.
\newblock In {\em Proceedings of the 2018 Conference of the North American
  Chapter of the Association for Computational Linguistics: Human Language
  Technologies, Volume 1 (Long Papers)}, volume~1, pages 2227--2237, 2018.

\bibitem{pirsiavash2014inferring}
Hamed Pirsiavash, Carl Vondrick, and Antonio Torralba.
\newblock Inferring the why in images.
\newblock {\em arXiv preprint arXiv:1406.5472}, 2014.

\bibitem{plummer2017phrase}
Bryan~A Plummer, Arun Mallya, Christopher~M Cervantes, Julia Hockenmaier, and
  Svetlana Lazebnik.
\newblock Phrase localization and visual relationship detection with
  comprehensive image-language cues.
\newblock In {\em Proc. ICCV}, 2017.

\bibitem{plummer2015flickr30k}
Bryan~A Plummer, Liwei Wang, Chris~M Cervantes, Juan~C Caicedo, Julia
  Hockenmaier, and Svetlana Lazebnik.
\newblock Flickr30k entities: Collecting region-to-phrase correspondences for
  richer image-to-sentence models.
\newblock In {\em Proceedings of the IEEE international conference on computer
  vision}, pages 2641--2649, 2015.

\bibitem{poliak_hypothesis_2018}
Adam Poliak, Jason Naradowsky, Aparajita Haldar, Rachel Rudinger, and Benjamin
  Van~Durme.
\newblock Hypothesis {Only} {Baselines} in {Natural} {Language} {Inference}.
\newblock {\em arXiv:1805.01042 [cs]}, May 2018.
\newblock arXiv: 1805.01042.

\bibitem{ramakrishnan2018overcoming}
Sainandan Ramakrishnan, Aishwarya Agrawal, and Stefan Lee.
\newblock Overcoming language priors in visual question answering with
  adversarial regularization.
\newblock In {\em Advances in Neural Information Processing Systems}, 2018.

\bibitem{ren2015faster}
Shaoqing Ren, Kaiming He, Ross Girshick, and Jian Sun.
\newblock Faster r-cnn: Towards real-time object detection with region proposal
  networks.
\newblock In {\em Advances in neural information processing systems}, pages
  91--99, 2015.

\bibitem{rhinehart2017first}
Nicholas Rhinehart and Kris~M Kitani.
\newblock First-person activity forecasting with online inverse reinforcement
  learning.
\newblock In {\em Proceedings of the IEEE International Conference on Computer
  Vision}, pages 3696--3705, 2017.

\bibitem{rohrbach2016grounding}
Anna Rohrbach, Marcus Rohrbach, Ronghang Hu, Trevor Darrell, and Bernt Schiele.
\newblock Grounding of textual phrases in images by reconstruction.
\newblock In {\em European Conference on Computer Vision}, pages 817--834.
  Springer, 2016.

\bibitem{RohrbachCVPR2017a}
Anna Rohrbach, Marcus Rohrbach, Siyu Tang, Seong~Joon Oh, and Bernt Schiele.
\newblock Generating descriptions with grounded and co-referenced people.
\newblock In {\em Proceedings IEEE Conference on Computer Vision and Pattern
  Recognition (CVPR) 2017}, Piscataway, NJ, USA, July 2017. IEEE.

\bibitem{rohrbach_movie_2017}
Anna Rohrbach, Atousa Torabi, Marcus Rohrbach, Niket Tandon, Christopher Pal,
  Hugo Larochelle, Aaron Courville, and Bernt Schiele.
\newblock Movie {Description}.
\newblock {\em International Journal of Computer Vision}, 123(1):94--120, May
  2017.

\bibitem{rudinger2017social}
Rachel Rudinger, Chandler May, and Benjamin Van~Durme.
\newblock Social bias in elicited natural language inferences.
\newblock In {\em Proceedings of the First ACL Workshop on Ethics in Natural
  Language Processing}, pages 74--79, 2017.

\bibitem{sap2017connotation}
Maarten Sap, Marcella~Cindy Prasettio, Ari Holtzman, Hannah Rashkin, and Yejin
  Choi.
\newblock Connotation frames of power and agency in modern films.
\newblock In {\em Proceedings of the 2017 Conference on Empirical Methods in
  Natural Language Processing}, pages 2329--2334, 2017.

\bibitem{saxe2013exact}
Andrew~M Saxe, James~L McClelland, and Surya Ganguli.
\newblock Exact solutions to the nonlinear dynamics of learning in deep linear
  neural networks.
\newblock {\em arXiv preprint arXiv:1312.6120}, 2013.

\bibitem{schofield2016gender}
Alexandra Schofield and Leo Mehr.
\newblock Gender-distinguishing features in film dialogue.
\newblock In {\em Proceedings of the Fifth Workshop on Computational
  Linguistics for Literature}, pages 32--39, 2016.

\bibitem{Schwartz:2017}
Roy Schwartz, Maarten Sap, Ioannis Konstas, Li Zilles, Yejin Choi, and Noah~A.
  Smith.
\newblock The effect of different writing tasks on linguistic style: A case
  study of the {ROC} story cloze task.
\newblock In {\em Proc. of CoNLL}, 2017.

\bibitem{sennrich-EtAl:2017:EACLDemo}
Rico Sennrich, Orhan Firat, Kyunghyun Cho, Alexandra Birch, Barry Haddow,
  Julian Hitschler, Marcin Junczys-Dowmunt, Samuel L\"{a}ubli, Antonio~Valerio
  Miceli~Barone, Jozef Mokry, and Maria Nadejde.
\newblock Nematus: a toolkit for neural machine translation.
\newblock In {\em Proceedings of the Software Demonstrations of the 15th
  Conference of the European Chapter of the Association for Computational
  Linguistics}, pages 65--68, Valencia, Spain, April 2017. Association for
  Computational Linguistics.

\bibitem{singh2016krishnacam}
Krishna~Kumar Singh, Kayvon Fatahalian, and Alexei~A Efros.
\newblock Krishnacam: Using a longitudinal, single-person, egocentric dataset
  for scene understanding tasks.
\newblock In {\em Applications of Computer Vision (WACV), 2016 IEEE Winter
  Conference on}, pages 1--9. IEEE, 2016.

\bibitem{tapaswi2016movieqa}
Makarand Tapaswi, Yukun Zhu, Rainer Stiefelhagen, Antonio Torralba, Raquel
  Urtasun, and Sanja Fidler.
\newblock Movieqa: Understanding stories in movies through question-answering.
\newblock In {\em Proceedings of the IEEE conference on computer vision and
  pattern recognition}, pages 4631--4640, 2016.

\bibitem{torralba2011unbiased}
Antonio Torralba and Alexei~A Efros.
\newblock Unbiased look at dataset bias.
\newblock In {\em Computer Vision and Pattern Recognition (CVPR), 2011 IEEE
  Conference on}, pages 1521--1528. IEEE, 2011.

\bibitem{moviegraphs}
Paul Vicol, Makarand Tapaswi, Lluis Castrejon, and Sanja Fidler.
\newblock Moviegraphs: Towards understanding human-centric situations from
  videos.
\newblock In {\em {IEEE Conference on Computer Vision and Pattern Recognition
  (CVPR)}}, 2018.

\bibitem{vondrick2016anticipating}
Carl Vondrick, Hamed Pirsiavash, and Antonio Torralba.
\newblock Anticipating visual representations from unlabeled video.
\newblock In {\em Proceedings of the IEEE Conference on Computer Vision and
  Pattern Recognition}, pages 98--106, 2016.

\bibitem{wagner18eccvw}
Misha Wagner, Hector Basevi, Rakshith Shetty, Wenbin Li, Mateusz Malinowski,
  Mario Fritz, and Ales Leonardis.
\newblock Answering visual what-if questions: From actions to predicted scene
  descriptions.
\newblock In {\em Visual Learning and Embodied Agents in Simulation
  Environments Workshop at European Conference on Computer Vision}, 2018.

\bibitem{Wang2017FVQAFV}
Peng Wang, Qi Wu, Chunhua Shen, Anton van~den Hengel, and Anthony~R. Dick.
\newblock Fvqa: Fact-based visual question answering.
\newblock {\em IEEE transactions on pattern analysis and machine intelligence},
  2017.

\bibitem{wieting2017learning}
John Wieting, Jonathan Mallinson, and Kevin Gimpel.
\newblock Learning paraphrastic sentence embeddings from back-translated
  bitext.
\newblock In {\em Proceedings of the 2017 Conference on Empirical Methods in
  Natural Language Processing}, pages 274--285, 2017.

\bibitem{williams17multisnli}
Adina Williams, Nikita Nangia, and Samuel Bowman.
\newblock A broad-coverage challenge corpus for sentence understanding through
  inference.
\newblock In {\em Proceedings of the 2018 Conference of the North American
  Chapter of the Association for Computational Linguistics: Human Language
  Technologies, Volume 1 (Long Papers)}, pages 1112--1122. Association for
  Computational Linguistics, 2018.

\bibitem{Wu2016AskMA}
Qi Wu, Peng Wang, Chunhua Shen, Anthony~R. Dick, and Anton van~den Hengel.
\newblock Ask me anything: Free-form visual question answering based on
  knowledge from external sources.
\newblock {\em 2016 IEEE Conference on Computer Vision and Pattern Recognition
  (CVPR)}, pages 4622--4630, 2016.

\bibitem{ye2018interpretable}
Tian Ye, Xiaolong Wang, James Davidson, and Abhinav Gupta.
\newblock Interpretable intuitive physics model.
\newblock In {\em European Conference on Computer Vision}, pages 89--105.
  Springer, 2018.

\bibitem{yoshikawa2018stair}
Yuya Yoshikawa, Jiaqing Lin, and Akikazu Takeuchi.
\newblock Stair actions: A video dataset of everyday home actions.
\newblock {\em arXiv preprint arXiv:1804.04326}, 2018.

\bibitem{yu_visual_2015}
Licheng Yu, Eunbyung Park, Alexander~C. Berg, and Tamara~L. Berg.
\newblock Visual {Madlibs}: {Fill} in the blank {Image} {Generation} and
  {Question} {Answering}.
\newblock {\em arXiv:1506.00278 [cs]}, May 2015.
\newblock arXiv: 1506.00278.

\bibitem{yu2016modeling}
Licheng Yu, Patrick Poirson, Shan Yang, Alexander~C Berg, and Tamara~L Berg.
\newblock Modeling context in referring expressions.
\newblock In {\em European Conference on Computer Vision}, pages 69--85.
  Springer, 2016.

\bibitem{yu2017joint}
Licheng Yu, Hao Tan, Mohit Bansal, and Tamara~L Berg.
\newblock A joint speakerlistener-reinforcer model for referring expressions.
\newblock In {\em Computer Vision and Pattern Recognition (CVPR)}, volume~2,
  2017.

\bibitem{zellers2018swagaf}
Rowan Zellers, Yonatan Bisk, Roy Schwartz, and Yejin Choi.
\newblock Swag: A large-scale adversarial dataset for grounded commonsense
  inference.
\newblock In {\em Proceedings of the 2018 Conference on Empirical Methods in
  Natural Language Processing (EMNLP)}, 2018.

\bibitem{zhao2017men}
Jieyu Zhao, Tianlu Wang, Mark Yatskar, Vicente Ordonez, and Kai-Wei Chang.
\newblock Men also like shopping: Reducing gender bias amplification using
  corpus-level constraints.
\newblock In {\em Proceedings of the 2017 Conference on Empirical Methods in
  Natural Language Processing}, pages 2979--2989, 2017.

\bibitem{Zhou2018TowardsAL}
Luowei Zhou, Chenliang Xu, and Jason~J. Corso.
\newblock Towards automatic learning of procedures from web instructional
  videos.
\newblock In {\em AAAI}, 2018.

\bibitem{zhou2015temporal}
Yipin Zhou and Tamara~L Berg.
\newblock Temporal perception and prediction in ego-centric video.
\newblock In {\em Proceedings of the IEEE International Conference on Computer
  Vision}, pages 4498--4506, 2015.

\bibitem{zhu2016cvpr}
Yuke Zhu, Oliver Groth, Michael Bernstein, and Li Fei-Fei.
\newblock {Visual7W: Grounded Question Answering in Images}.
\newblock In {\em {IEEE Conference on Computer Vision and Pattern
  Recognition}}, 2016.

\bibitem{moviebook}
Yukun Zhu, Ryan Kiros, Richard Zemel, Ruslan Salakhutdinov, Raquel Urtasun,
  Antonio Torralba, and Sanja Fidler.
\newblock Aligning books and movies: Towards story-like visual explanations by
  watching movies and reading books.
\newblock In {\em arXiv preprint arXiv:1506.06724}, 2015.

\end{thebibliography}
}

\end{document}